\documentclass[nohyperref]{article}

\usepackage{microtype}
\usepackage{graphicx}
\usepackage{subfigure}
\usepackage{booktabs} 

\usepackage{hyperref}

\usepackage[accepted]{icml2023}

\usepackage{amsmath}
\usepackage{amssymb}
\usepackage{mathtools}
\usepackage{amsthm}

\usepackage[capitalise,noabbrev,nameinlink]{cleveref}
\creflabelformat{equation}{#2\textup{#1}#3}
\creflabelformat{section}{#2\textup{#1}#3}
\creflabelformat{appendix}{#2\textup{#1}#3}
\creflabelformat{figure}{#2\textup{#1}#3}
\crefname{algocf}{Algorithm}{Algorithms}
\Crefname{algocf}{Algorithm}{Algorithms}

\definecolor{DarkBlue}{rgb}{0,0.08,0.45}

\hypersetup{%
colorlinks=true,
linkcolor=DarkBlue,
citecolor=DarkBlue,
filecolor=DarkBlue,
urlcolor=DarkBlue}

\usepackage{amsmath,amssymb,graphicx,color,booktabs,bm,relsize,enumitem,multirow,amsthm,epsfig,mathtools,xcolor}
\usepackage{dsfont}
\usepackage{xspace}
\usepackage{wrapfig}

\newcommand{\blap}[1]{\vbox to 0pt{\hbox{#1}\vss}}

\newcommand{\degree}{^{\circ}\xspace}
\theoremstyle{plain}

\theoremstyle{definition}

\theoremstyle{remark}

\usepackage{tikz}
\usetikzlibrary{fit,calc}
\newcommand*{\tikzmk}[1]{\tikz[remember picture,overlay,] \node (#1) {};\ignorespaces}
\newcommand{\boxit}[1]{\tikz[remember picture,overlay]{\node[yshift=4pt,fill=#1,opacity=.125,fit={(A)($(B)+(1.0\linewidth,.8\baselineskip)$)}] {};}\ignorespaces}
\newcommand{\boxitt}[1]{\tikz[remember picture,overlay]{\node[yshift=1.5pt,fill=#1,opacity=.125,fit={(A)($(B)+(1.0\linewidth,.8\baselineskip)$)}] {};}\ignorespaces}

\colorlet{pink}{red!40}
\colorlet{blue}{cyan!60}

\usepackage[textsize=tiny]{todonotes}

\icmltitlerunning{Multi-View Masked World Models for Visual Robotic Manipulation}

\begin{document}

\twocolumn[
\icmltitle{Multi-View Masked World Models for Visual Robotic Manipulation}



\icmlsetsymbol{equal}{*}

\begin{icmlauthorlist}
\icmlauthor{Younggyo Seo}{equal,kaist}
\icmlauthor{Junsu Kim}{equal,kaist}
\icmlauthor{Stephen James}{dyson}
\icmlauthor{Kimin Lee}{google}
\icmlauthor{Jinwoo Shin}{kaist}
\icmlauthor{Pieter Abbeel}{berkeley}
\end{icmlauthorlist}

\icmlaffiliation{kaist}{KAIST}
\icmlaffiliation{dyson}{Dyson Robot Learning Lab}
\icmlaffiliation{google}{Google Research}
\icmlaffiliation{berkeley}{UC Berkeley}

\icmlcorrespondingauthor{Younggyo Seo}{younggyo.seo@kaist.ac.kr}

\icmlkeywords{Machine Learning, ICML}

\vskip 0.3in
]

\printAffiliationsAndNotice{\icmlEqualContribution}

\begin{abstract}
Visual robotic manipulation research and applications often use multiple cameras, or views, to better perceive the world. How else can we utilize the richness of multi-view data?
In this paper, we investigate how to learn good representations with multi-view data and utilize them for visual robotic manipulation.
Specifically, we train a multi-view masked autoencoder which reconstructs pixels of randomly masked viewpoints and then learn a world model operating on the representations from the autoencoder.
We demonstrate the effectiveness of our method in a range of scenarios, including multi-view control and single-view control with auxiliary cameras for representation learning.
We also show that the multi-view masked autoencoder trained with multiple randomized viewpoints enables training a policy with strong viewpoint randomization and transferring the policy to solve real-robot tasks without camera calibration and an adaptation procedure. 
Video demonstrations are available at: \url{https://sites.google.com/view/mv-mwm}.
\end{abstract}

\section{Introduction}
The camera is a ubiquitous instrument for robot vision that provides rich information about a workspace from various viewpoints.
Thus it has been a widely-used technique for roboticists to utilize multiple cameras for solving complex manipulation tasks \citep{akkaya2019solving,akinola2020learning,james2021coarse}. 
However, prior work utilize multi-view data naively as inputs and has yet to investigate how to learn effective multi-view representations.
Considering recent studies have shown the benefit of single-view representation learning for control \citep{nair2022r3m,radosavovic2022real},
it is desirable to explore the potential of multi-view representation learning for visual robotic manipulation.

A notable exception is the work of \citet{sermanet2018time}, which learns view-invariant representations via contrastive learning.
However, enforcing viewpoint invariance assumes that all viewpoints share similar information and thus requires careful curation of positive and negative pairs, similar to other contrastive approaches that often depend on complex design choices about sampling such pairs \citep{arora2019theoretical}.
This can make it challenging to learn representations from diverse multi-view data and limit its applicability to a narrow distribution of visual robotic manipulation setups.
Instead, we aim to develop a simple multi-view representation learning method effective for more diverse setups.

In this paper, we present Multi-View Masked World Model (MV-MWM), a reinforcement learning framework that trains a multi-view masked autoencoder for representation learning and a world model to solve visual manipulation tasks.
Our autoencoder consists of a synergistic combination of \textit{view-masking}: which masks viewpoints at random, and \textit{video autoencoding}: which reconstructs video frames of both masked and unmasked viewpoints.
We find our autoencoder effectively learns representations that capture useful information of the current viewpoint but also the cross-view information from different viewpoints.
For behavior learning, we learn a world model on frozen representations from either single-view or multi-view data, which is particularly feasible as the autoencoder consists of vision transformer \citep{dosovitskiy2020image} layers that take inputs of variable sizes.
We then train actor and critic with imaginary trajectories from the world model \citep{hafner2020mastering}.

\begin{figure*} [t!] \centering
\includegraphics[width=0.99\textwidth]{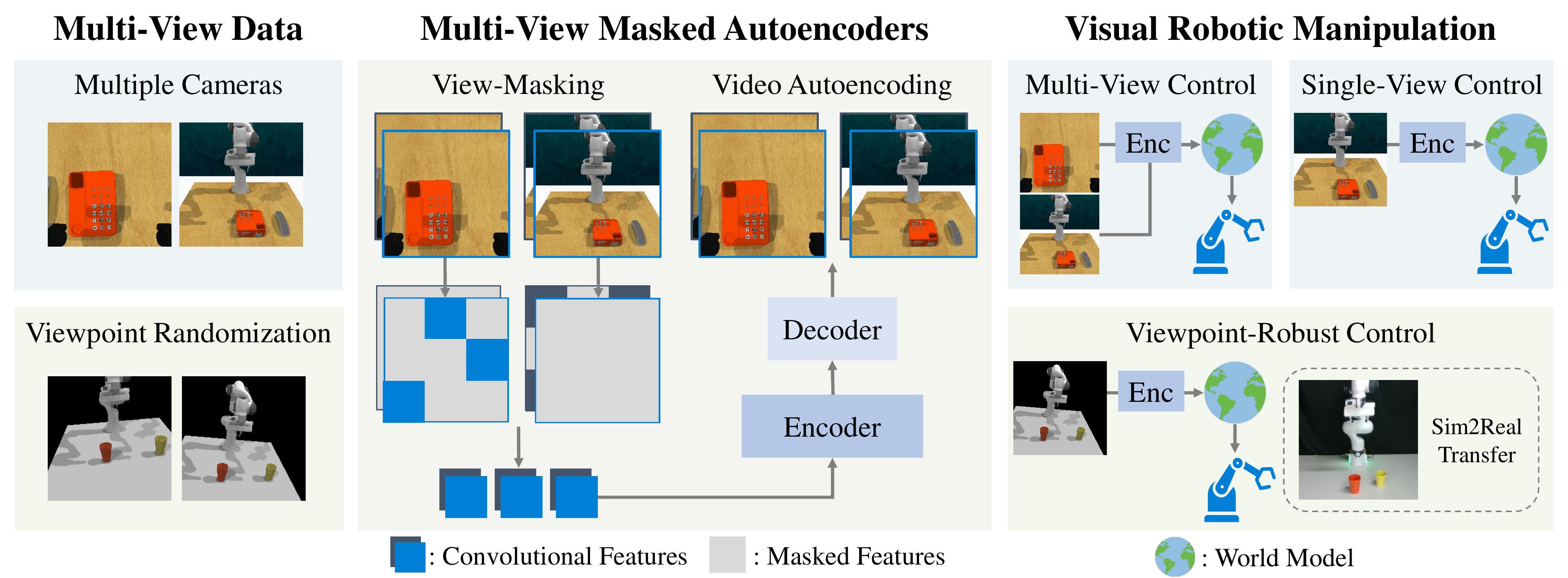}
    \vspace{-0.05in}
    \caption{Illustration of our framework. Given multi-view data from multiple cameras or multiple randomized viewpoints, we mask viewpoints from video frames at random and train a multi-view masked autoencoder to reconstruct pixels of both masked and unmasked viewpoints.
    We then learn a world model upon frozen autoencoder representations to solve tasks from various robotic manipulation setups, including a multi-view control, a single-view control, and a viewpoint-robust control in both simulation and real-world.
    } \label{fig:mvmwm_overview}
\end{figure*}

We highlight the main contributions of this paper below:
\begin{itemize}[topsep=1.0pt,itemsep=0.85pt]
    \item [$\bullet$] We present Multi-View Masked World Model, a reinforcement learning framework that trains a multi-view masked autoencoder with a view-masking and learns a world model upon autoencoder representations.
    \item [$\bullet$] We demonstrate the effectiveness of MV-MWM
    in various visual robotic manipulation setups.
    These setups include (i) a \textit{multi-view} control where agents operate on multi-view data, (ii) a \textit{single-view} control where agents operate on single-view data but use auxiliary cameras for representation learning, and (iii) a \textit{viewpoint-robust} control where agents operate on single randomized viewpoint but use multiple randomized viewpoints for representation learning, as illustrated in \cref{fig:mvmwm_overview}.
    Our experiments on RLBench \citep{james2020rlbench} show that our method outperforms single-view representation learning baselines \citep{radford2021learning,he2021masked,seo2022masked} and a multi-view representation learning baseline \citep{sermanet2018time}.
    \item [$\bullet$] We show that MV-MWM can solve real-world robotic manipulation tasks by transferring a policy trained in simulation to a real-robot without camera calibration.
    We further show that MV-MWM works on a range of viewpoints and even with a hand-held camera subject to rotation or shaking while solving the tasks; showcasing impressive visual servoing robustness.
\end{itemize}

\section{Related Work}
\paragraph{Visual control with multiple cameras}
Leveraging multiple cameras has long been considered a practical and feasible technique in robotics, as the camera is usually an affordable and ubiquitous device~\citep{sola2008fusing, carrera2011slam, yang2021asynchronous}.
Based on recent advances in computer vision and robot learning, there have been several approaches that utilize multi-view data from multiple cameras for visual control~\citep{sermanet2018time,akinola2020learning,zhan2020framework,chen2021unsupervised,hsu2022vision,jangir2022look,shridhar2022perceiver,guhur2022instruction}.
While most approaches utilize multi-view data directly as inputs for robots, recent works have demonstrated that self-supervised learning that learns view-invariant representations~\citep{sermanet2018time} or 3D keypoints~\citep{chen2021unsupervised} can be useful for downstream tasks.
Yet these approaches assume viewpoints have similar characteristics or require multiple cameras for both representation learning and behavior learning phases, limiting their applicability to a narrow set of setups.
Instead this work aims to develop a framework that can learn representations from diverse viewpoints and leverage them for various setups.

\paragraph{Unsupervised representation learning for visual control}
Most prior researches on representation learning for visual control have focused on solving control tasks using the representations learned with single-view data \citep{watter2015embed,oord2018representation,gelada2019deepmdp,hafner2019learning,yarats2021mastering,seo2022reinforcement}.
We instead demonstrate that multi-view representations can be also useful for single-view control.
Another line of works related to our work have demonstrated that pre-training with self-supervised learning enables agents to solve control tasks with frozen representations \citep{stooke2021decoupling,schwarzer2021pretraining,nair2022r3m,parisi2022unsurprising,xiao2022masked,radosavovic2022real}.
Our framework also learns to solve tasks with frozen representations but it differs in that representations are continually updated using the online samples throughout training.
Incorporating pre-training into our framework would be an interesting future direction.

\begin{figure*} [t!] \centering
\includegraphics[width=0.99\textwidth]{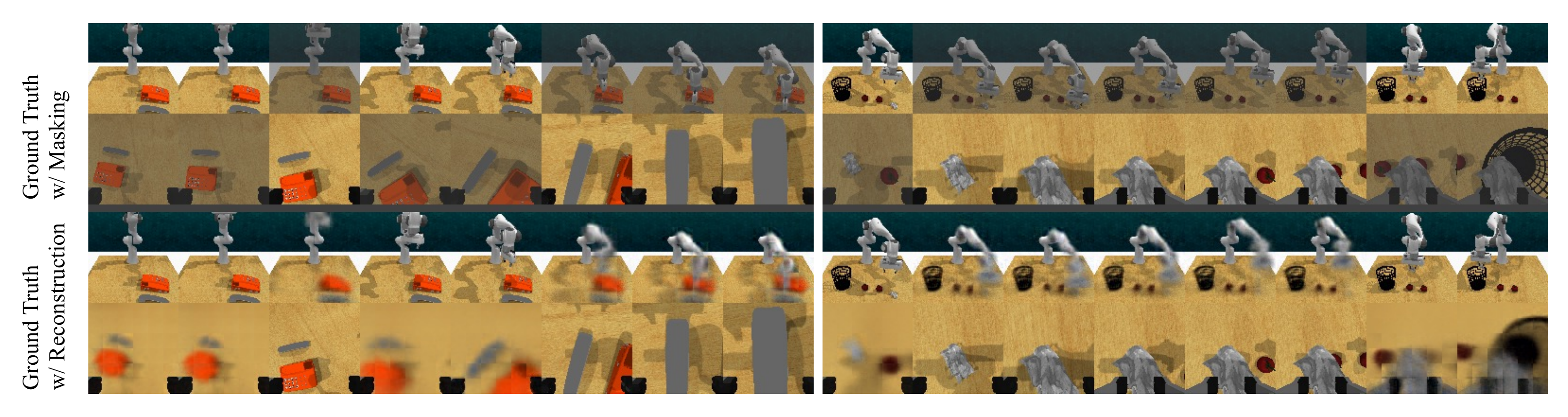}
\vspace{-0.2in}
\caption{Masked view reconstruction on Phone On Base (left) and Put Rubbish in Bin (right) tasks from RLBench~\citep{james2020rlbench}. We visualize ground-truth frames with masked viewpoints (upper two rows) and ground-truth frames with reconstructed frames (lower two rows). We find that the model can reconstruct masked viewpoints, successfully capturing the location of objects in the scene.}  \label{fig:mvmae_reconstruction}
\end{figure*}

\section{Multi-View Masked World Models for Visual Robotic Manipulation}
We present Multi-View Masked World Models (MV-MWM), a reinforcement learning framework
that learns multi-view representations and utilize them for visual robotic manipulation.
Our method builds on top of the Masked World Models (MWM; \citealt{seo2022masked}) framework, which learns a world model on frozen masked autoencoder features.
We first introduce how to learn multi-view representations in \cref{sec:method_multi_view}.
We then describe in \cref{sec:method_world_model_and_behavior} how to utilize them for learning world models and behaviors to solve visual manipulation tasks.
We provide the overview of our framework in \cref{fig:mvmwm_overview}.
The key difference to MWM is detailed in \cref{appendix:baselines} and \cref{alg:mvmwm}.

\subsection{Multi-View Representation Learning}
\label{sec:method_multi_view}
For representation learning with multi-view data, we introduce a new model called Multi-View Masked Autoencoder (MV-MAE). 
Our main idea is to train a \textit{video masked autoencoder}~\citep{feichtenhofer2022masked,tong2022videomae} with \textit{view-masking} to reconstruct missing pixels of randomly masked viewpoints.
We also incorporate the idea of a prior work~\citep{seo2022masked} that masks convolutional features instead of pixel patches~\citep{he2021masked} and predicts rewards to learn representations capturing fine-grained details required for visual control.
We first describe each component in detail and provide the formal objective.

\paragraph{Convolutional feature embedding} Unlike prior work that masks random pixel patches~\citep{he2021masked}, we embed camera observations into convolutional feature maps and mask these features following the design of~\citet{seo2022masked}.
This is based on the observation where masked image modeling with pixel patch masking can make it difficult for the model to learn fine-grained details within patches.
Specifically, we downsample $96\times96\times3$ inputs images to convolutional feature maps with the spatial size of $6\times6$ by introducing 4 convolutional layers. We separately process observations from each viewpoint with convolutional layers that share parameters. For each viewpoint, we add fixed 2D sin-cos position embeddings~\citep{chen2021empirical} to the features.
We also add learnable 1D parameters representing each viewpoint and timestep to features of each video frame from different viewpoints, following \citet{geng2022multimodal} that introduces parameters for vision and language inputs.

\paragraph{View masking} To learn cross-view information from multiple viewpoints,
we introduce a novel view-masking strategy that masks all the features from a randomly selected viewpoint.
Specifically, we mask randomly selected viewpoints from video frames by randomly sampling one viewpoint for each frame.
We also mask randomly selected features from remaining viewpoints (see~\cref{fig:mvmwm_overview}) because we want the autoencoder to learn not only cross-view information but also the information within each viewpoint by reconstructing raw visual observations with masked features.
Then we flatten the unmasked features and concatenate them into a single sequence.
We empirically find that the proposed view-masking can be more effective than uniform-masking scheme by explicitly encouraging multi-view representation learning (see \cref{fig:analysis_view_masking} for supporting experiments).

\paragraph{Video autoencoding} 
Despite its potential, the proposed masked view reconstruction objective might be too challenging for the autoencoder without any access to information from missing viewpoints.
To address this issue, we consider a combination of video masked autoencoding~\citep{feichtenhofer2022masked,tong2022videomae} and the proposed view-masking strategy.
Because the autoencoder attends to unmasked neighbor frames from the same view, the model can focus on modeling important information such as the movement of robot arms, while ignoring redundant information such as background for reconstructing masked viewpoints (see~\cref{fig:mvmae_reconstruction} for examples of masked view reconstruction).
Specifically, our encoder processes a sequence of unmasked features from all viewpoints and video frames through a series of vision transformer (ViT; \citealt{dosovitskiy2020image}) layers.
Then we concatenate a set of mask tokens with encoded features, and add learnable parameters for each viewpoint and frame to corresponding features and mask tokens.
The decoder processes them through ViT layers and linearly projects them into pixel patch predictions.
We also follow the idea of \citet{seo2022masked} that predicts a reward to encode task-relevant information.

\paragraph{Objective} Let $o^{v}_{t,T}= \{o^{v}_{t}, ..., o^{v}_{t+T-1}\}$ be a video from a viewpoint $v \in \mathcal{V}$ where $t$ is current timestep, $T$ is window size of the video autoencoder, and
$\mathcal{V}$ is a set of available viewpoints. Given videos $\{o^{v}_{t,T}\}_{v \in \mathcal{V}}$ from multiple viewpoints, rewards $r_{t, T} = \{r_{1}, ..., r_{t+T-1}\}$, and a mask ratio of $m$, MV-MAE consists of following components:
\begin{align}
&\text{Convolution stem:} &&h^{v}_{t,T}= f^{\tt{conv}}_{\phi}(o^{v}_{t,T})\nonumber \\
&\text{View-masking:} &&h_{t,T}^{m}\sim p^{\tt{mask}}(h_{t,T}^{m}\,|\,\{h^{v}_{t,T}\}_{v \in \mathcal{V}}, m)
\nonumber \\
&\text{ViT encoder:} &&z_{t,T}^{m}\sim p_{\phi}(z_{t,T}^{m} \,|\,h_{t,T}^{m}) \label{eq:mvmae}\\
&\text{ViT decoder:} 
&&\begin{aligned}\raisebox{1.9ex}{\llap{\blap{\ensuremath{ \hspace{0.1ex} \begin{cases} \hphantom{R} \\ \hphantom{T} \end{cases} \hspace*{-4ex}
}}}}
&\{\hat{o}^{v}_{t,T}\}_{v \in \mathcal{V}}\sim p_\phi(\{\hat{o}^{v}_{t,T}\}_{v \in \mathcal{V}}\,|\,z_{t,T}^{m})\\
&\hat{r}_{t,T}\sim p_{\phi}(\hat{r}_{t,T} \,|\,z_{t,T}^{m})
\end{aligned} \nonumber
\end{align}
We train the model to reconstruct pixels and predict rewards, which corresponds to optimizing model parameters $\phi$ by minimizing the negative log-likelihood as below:
\begin{align}
    &\mathcal{L}^{\tt{mvmae}}(\phi) = -\ln p_{\phi}(\{o^{v}_{t,T}\}_{v \in \mathcal{V}}\,|\,z^{m}_{t,T}) -\ln p_{\phi}(r_{t,T}\,|\,z^{m}_{t,T}) \nonumber
\end{align}

\subsection{World Model and Behavior Learning}
\label{sec:method_world_model_and_behavior}
\paragraph{Representations} 
Once we learn multi-view representations, we leverage them for learning a world model and utilize the world model for visual control.
A favorable property of MV-MAE is that the ViT encoder can extract representations from single-view images even if the encoder is trained with multi-view video data.
Based on this property, we learn a world model for solving visual robotic manipulation tasks from two setups: \textit{multi-view} and \textit{single-view} control. These setups vary in the availability of viewpoints during the control phase, yet both adopt a multi-view data for representation learning.
To provide input to the world model, we extract representations by encoding multi-view data $\{o^{v}_{t}\}_{v \in \mathcal{V}}$ or single-view data $o^{\tilde{v}}_{t}$ by using the MV-MAE encoder in \cref{eq:mvmae} with $m=0$ and $T=1$.\footnote{We extract image representations as our world model operates upon observations from each timestep with a recurrent architecture.}
We exploit the notation by denoting both representations as $z_{t}$ because the objective for learning the world model is same for both single-view and multi-view world models.

\paragraph{World model learning}
Following \citet{seo2022masked}, we implement the world model as a variant of recurrent state-space model (RSSM; \citealt{hafner2019learning}) that takes frozen autoencoder representations as inputs and reconstruction targets.
The world model consists of following components:
\begin{gather}
\begin{aligned}
&\text{Encoder:} &&s_t\sim q_\theta(s_{t} \,|\,s_{t-1},a_{t-1}, z_{t}) \\
&\text{Decoder:} &&\begin{aligned}\raisebox{1.9ex}{\llap{\blap{\ensuremath{ \hspace{0.1ex} \begin{cases} \hphantom{R} \\ \hphantom{T} \end{cases} \hspace*{-4ex}
}}}}
&\hat{z}_t\sim p_\theta(\hat{z}_{t} \,|\,s_{t})\\
&\hat{r}_t\sim p_\theta(\hat{r}_{t} \,|\,s_{t})
\end{aligned} \\
&\text{Dynamics model:} &&\hat{s}_t\sim p_\theta(\hat{s}_{t} \,|\,s_{t-1}, a_{t-1})
\label{eq:mvmwm}
\end{aligned}
\end{gather}
The encoder extracts state $s_{t}$ from previous state $s_{t-1}$, previous action $a_{t-1}$, and current autoencoder representations $z_{t}$.
The dynamics model learns to predict $s_{t}$ without an access to $z_{t}$, which enables the model to predict forward into the future.
Following \citet{hafner2020mastering}, we utilize discrete latents for $s_{t}$.
The decoder reconstructs $z_{t}$ to provide learning signal for model states and predicts $r_{t}$ to allow for computing rewards from future states without decoding future autoencoder representations.
All model parameters $\theta$ are jointly optimized by minimizing the negative variational lower bound \citep{kingma2013auto}:
\begin{gather}
\begin{aligned}
    &\mathcal{L}^{\tt{wm}}(\theta) = -\ln p_\theta(z_{t} \,|\,s_{t})-\ln p_\theta(r_{t} \,|\,s_{t}) \\
    &\quad+  \beta\,\text{KL}\big[ q_{\theta}(s_{t}|s_{t-1},a_{t-1},z_{t}) \,\Vert\,  p_{\theta}(\hat{s}_{t}|s_{t-1},a_{t-1}) \big], \nonumber
\end{aligned}
\end{gather}
where $\beta$ is a scale hyperparameter.

\begin{algorithm}[t]
\caption{Multi-View Masked World Models. \\Key differences to MWM \citep{seo2022masked} in gray.} \label{alg:mvmwm}
\begin{algorithmic}[1]
\STATE Initialize parameters $\phi$, $\theta$, $\psi$, $\xi$ 
\STATE Initialize a buffer $\mathcal{B}$ and a fixed expert buffer $\mathcal{B}^{\tt{e}}$
\FOR{each timestep $t$}
\STATE {\textsc{// Collect transitions}}
\STATE Update state $s_{t} \sim q_{\theta}(s_{t}|s_{t-1},a_{t-1},z_{t})$
\STATE Sample action $a_{t} \sim p_{\psi}(a_{t}|s_{t})$ 
\STATE Add transition to replay buffer $\mathcal{B}$\vspace{0.05in}
\STATE \tikzmk{A}{\textsc{// Multi-view representation learning}}
\STATE Sample $(\{o^{v}_{j,T}\}_{v \in \mathcal{V}}, r_{j,T}) \sim \mathcal{B}$
\STATE Update $\phi$ by minimizing $\mathcal{L}^{\tt{mvmae}}(\phi)$\vspace{0.05in}
\STATE \tikzmk{B}\boxit{gray}{{\textsc{// World model learning}}}
\STATE Sample $(\{o^{v}_{j}\}_{v \in \mathcal{V}}, a_{j-1}, r_{j}) \sim \mathcal{B}$
\STATE \tikzmk{A}Obtain $z_{j}$ from either $\{o^{v}_{j}\}_{v \in \mathcal{V}}$ (multi-view control) or $o_{j}^{\tilde{v}}$ (single-view control with $\tilde{v}$)
\STATE \tikzmk{B}\boxitt{gray}Update $\theta$ by minimizing $\mathcal{L}^{\tt{wm}}(\theta)$\vspace{0.05in}
\STATE {\textsc{// Behavior learning}}
\STATE Obtain initial state $\hat{s}_{0}$ and imagine $\{(\hat{s_{i}}, \hat{a_{i}}, \hat{r_{i}})\}_{i=1}^{H}$
\STATE Sample expert demonstration $(\{o^{v}_{j}\}_{v \in \mathcal{V}}, a^{\tt{e}}_{j}) \sim \mathcal{B}^{\tt{e}}$
\STATE Update $\xi$ by minimizing $\mathcal{L}^{\tt{critic}}(\xi)$
\STATE Update $\psi$ by minimizing $\mathcal{L}^{\tt{actor}}(\psi)$
\ENDFOR
\end{algorithmic}
\end{algorithm}
\vspace{-.1in}

\paragraph{Behavior learning}
For behavior learning, we employ the actor-critic learning scheme of DreamerV2~\citep{hafner2020mastering} where the objective is to train a policy that maximizes the predicted future values by gradients propagated through the world model.
Specifically, a stochastic actor and a deterministic critic is defined as below:
\begin{gather}
\begin{aligned}
&\text{Actor:} &&\hat{a}_{t} \sim p_{\psi}(\hat{a}_{t}\,|\,\hat{s}_{t}) \\
&\text{Critic:} &&v_{\xi}(\hat{s}_{t}) \approx \mathbb{E}_{p_{\theta}}\left[\textstyle\sum_{i \leq t}\gamma^{i - t} \hat{r}_{i}\right]
\label{eq:actor_critic}
\end{aligned}
\end{gather}
where $\{(\hat{s}_{t}, \hat{a}_{t}, \hat{r}_{t})\}_{t=1}^{H}$ is predicted from initial state $\hat{s}_{0}$ using the stochastic actor and dynamics model in \cref{eq:mvmwm}. Given a $\lambda$-return \citep{schulman2015high} defined as below:
\begin{align}
    V_{t}^{\lambda}\doteq \hat{r}_{t} + \gamma
    \begin{cases}
      (1 - \lambda)v_{\xi}(\hat{s}_{t+1})+\lambda V_{t+1}^{\lambda} & \text{if}\ t<H \\
      v_{\xi}(\hat{s}_{H}) & \text{if}\ t=H
    \end{cases}
    \label{eq:lambda_return}
\end{align}
the critic is trained to regress the $\lambda$-return and the actor is trained to maximize the $\lambda$-return with gradients backpropagated through the world model. Moreover, to better utilize expert demonstrations which we assume the access to by following the setup of \citet{james2022q}, we introduce a behavior cloning loss that encourages the actor to imitate expert actions. Objectives are summarized as below:
\begin{align}
    &\mathcal{L}^{\tt{critic}}(\xi)\doteq\mathbb{E}_{p_{\theta},p_{\psi}}\left[\sum^{H-1}_{t=1} \frac{1}{2} \left(v_{\xi}(\hat{s}_{t}) - \text{sg}(V_{t}^{\lambda})\right)^{2}\right] \nonumber \\
    &\mathcal{L}^{\tt{actor}}(\psi)\doteq \mathbb{E}_{p_{\theta},p_{\psi}} \left[-V_{t}^{\lambda} - \eta\,\text{H}\left[a_{t}|\hat{s}_{t}\right] \right] -\ln p_{\psi}(a_{t}^{\tt{e}}|s_{t}) \nonumber
\end{align}
where sg is a stop gradient operation, $\eta$ is a scale hyperparameter for an entropy $\text{H}\left[a_{t}|\hat{s}_{t}\right]$, and $a^{\tt{e}}_{t}$ is an expert action.

To summarize, we iterate the processes of (i) training the \textit{video autoencoder} with \textit{view-masking} for multi-view representation learning, (ii) learning the world model and behaviors for \textit{multi-view} or \textit{single-view} control, (iii) collecting samples with environment interaction.
We highlight the key differences between MWM and MV-MWM in \cref{alg:mvmwm}.

\vspace{-0.1in}
\section{Experiments}
\label{sec:experiments}
We evaluate MV-MWM on challenging visual robotic manipulation tasks from RLBench \citep{james2020rlbench} -- a standard benchmark for vision-based robotics which has been shown to serve as a proxy for real-robot experiments \citep{james2022q}.
Furthermore, we evaluate the zero-shot performance of our method on real-world by transferring the trained agent to control real-robots.
We designed our experiments to explore the benefit of multi-view representation learning on practical and important robotics scenarios.
Specifically, we aim to investigate the following questions:
\vspace{-0.05in}
\begin{itemize}[leftmargin=5.5mm]
    \item Can MV-MWM learn multi-view representations useful for various visual robotic manipulation setups?
    \item Can MV-MWM trained in simulation be transferred to solve real-world visual robotic manipulation tasks?
    \item How does MV-MWM compare to baselines in terms of sample-efficiency and asymptotic performance?
    \item What is the effect of each component in MV-MWM?
\end{itemize}
\vspace{-0.1in}
\paragraph{Implementation}
We build our implementation upon the official implementation of MWM \citep{seo2022masked} and implementation details are same unless otherwise specified.
We run 8 parallel simulators to accelerate training by avoiding the bottleneck from a slow simulator.
Our autoencoder consists of the 8-layer ViT encoder and 6-layer ViT decoder, where the embedding dimension is set to 256.
We use the same set of hyperparameters for all experiments.
We provide more implementation details in \cref{appendix:implementation}.

\begin{figure} [t!] \centering
    \hfill
    \subfigure[Phone On Base]
    {
    \includegraphics[width=0.225\textwidth]{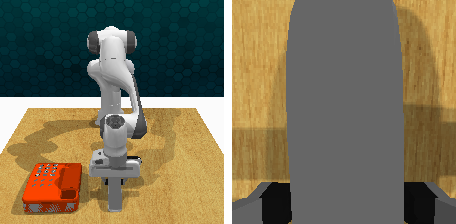}
    \label{fig:example_phone_on_base}} 
    \hfill
    \subfigure[Take Umbrella Out of Stand]
    {
    \includegraphics[width=0.225\textwidth]{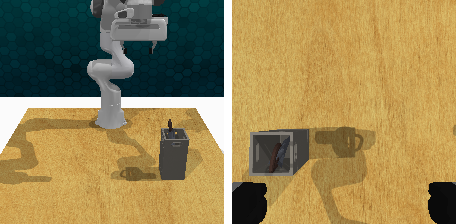}
    \label{fig:example_take_umbrella_out_of_stand}}
    \hfill\vspace{-0.075in}\\
    \hfill
    \subfigure[Put Rubbish in Bin]
    {
    \includegraphics[width=0.225\textwidth]{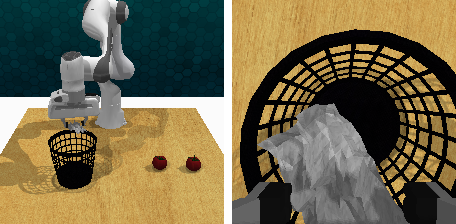}
    \label{fig:example_put_rubbish_in_bin}} 
    \hfill
    \subfigure[Stack Wine]
    {
    \includegraphics[width=0.225\textwidth]{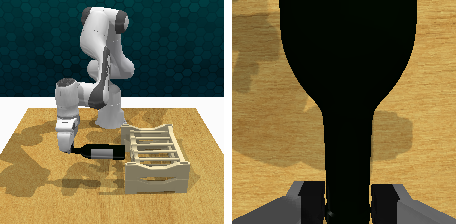}
    \label{fig:example_stack_wine}} 
    \hfill
    \vspace{-0.1in}
    \caption{
    Examples of multi-view data consisting of front and wrist camera observations used in our experiments. Front camera observations provide a broad look at a robot workspace and wrist camera observations provide a closer look at target objects.}
    \label{fig:example}
    \vspace{-0.1in}
\end{figure}

\paragraph{Environment}
For all experiments, we use only $96\times96$ RGB observations from each camera; proprioceptive state and depth are not used.
While RLBench is originally designed to evaluate the performance in a sparse reward setup, we design dense rewards for our experiments.
Moreover, to ease the difficulty of exploration in large action space, we enforce a robot gripper to be in an upright position except for a case where rotation is required for solving the task.
Following \citet{james2022q}, we fill a replay buffer with expert demonstrations.
Unlike prior approaches that utilize path planner with the policy to output next best gripper pose~\citep{james2022q,james2021coarse,shridhar2022perceiver,james2022lpr,james2022tree}, our RL agent outputs relative change in gripper position.
We provide further details in \cref{appendix:implementation}.

\begin{figure*} [t!]
    \centering
    \includegraphics[width=0.9\textwidth]{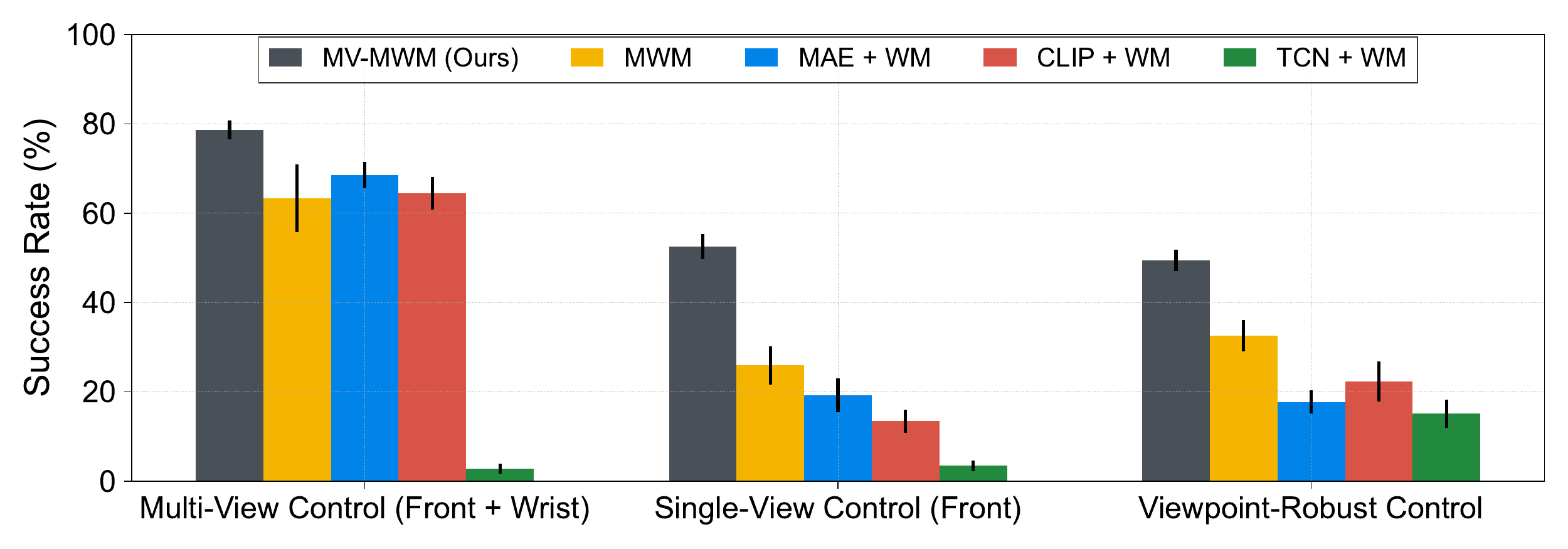}\vspace{-0.15in}
    \caption{
    Aggregate success rate on five multi-view and single-view control tasks and two viewpoint-robust control tasks.
    We report the average success rate evaluated using the last five model checkpoints.
    The result shows the mean and stratified bootstrap confidence interval across 32 runs for viewpoint-robust control and 20 runs otherwise.
    We provide the learning curve for all tasks in \cref{appendix:full_results}.}
    \label{fig:main_no_rand}
\end{figure*}

\paragraph{Baselines}
We first compare MV-MWM with MWM to evaluate the benefit of multi-view representation learning.
We note that both use the same amount of training data.
We also consider baselines that utilize frozen pre-trained representations. Specifically, we consider CLIP \citep{radford2021learning} and MAE \citep{he2021masked} representations as they have recently shown to be effective for robotic manipulation \citep{shridhar2021cliport,radosavovic2022real}.
Moreover, to compare our method with other multi-view representation learning method designed for visual control, we consider time contrastive network (TCN; \citet{sermanet2018time}) that enforces view-invariance through contrastive learning as a baseline.
We call these baselines as CLIP+WM, MAE+WM, and TCN+WM to denote that we learn world models upon these representations.
All methods use frozen representations for world model and behavior learning.
We note that MV-MWM, MWM, and TCN+WM learn representations from scratch throughout training, but we do not fine-tune the representations of CLIP+WM and MAE+WM for a comparison with frozen pre-trained representations.
We provide more details on baselines in \cref{appendix:baselines}.

\subsection{Multi-View and Single-View Control}
\label{sec:experiments_multi_view_single_view}
\paragraph{Multi-view control setup}
We evaluate MV-MWM on a multi-view control setup where the agent operates on both front and wrist cameras, which is a widely-used setup for visual manipulation with multiple cameras \citep{zhan2020framework,james2022q,jangir2022look}.
For all baselines, we extract representations from each viewpoint and use concatenated features as inputs to the agent.
For our method, we use multi-view representations from our autoencoder as we previously mentioned in \cref{sec:method_world_model_and_behavior}.

\paragraph{Single-view control setup}
We also consider a single-view control setup to investigate whether multi-view representation learning with auxiliary cameras can be helpful for training a single-view agent.
This can be useful for a practical scenario where we can utilize additional cameras during training, but the robot should operate on a single camera at deployment time.
We note that this setup has also been investigated in \citet{sermanet2018time}, but we use more different types of cameras for representation learning.
Specifically, we learn visual representations using multi-view data consisting of front and wrist camera observations and train the RL agent that operates on the front camera.

\paragraph{Results}
In \cref{fig:main_no_rand}, we observe that MV-MWM outperforms MWM, which shows that multi-view representation learning can be helpful for both multi-view and single-view control.
We find that TCN significantly fails to solve most of the tasks in both setups.
This shows the critical drawback of TCN, which suffers from mode collapse when negative samples are too similar (\textit{e.g.,} wrist camera observations look similar after grasping the objects in our case).
Interestingly, we also find that MAE+WM and CLIP+WM achieve non-zero success rates, which shows that large pre-trained models can extract useful representations.
However, MV-MWM largely outperforms both baselines, demonstrating that multi-view representation learning with in-domain data can be crucial for visual robotic manipulation.
Given the results, it would be an interesting direction to integrate pre-training with our multi-view representation learning by employing recently developed efficient fine-tuning techniques \citep{gao2021clip,zhang2021tip,sharma2023lossless}.

\begin{figure*}[t!]
\vspace{-0.1in}
\centering
\subfigure{
    \includegraphics[width=0.46\textwidth]{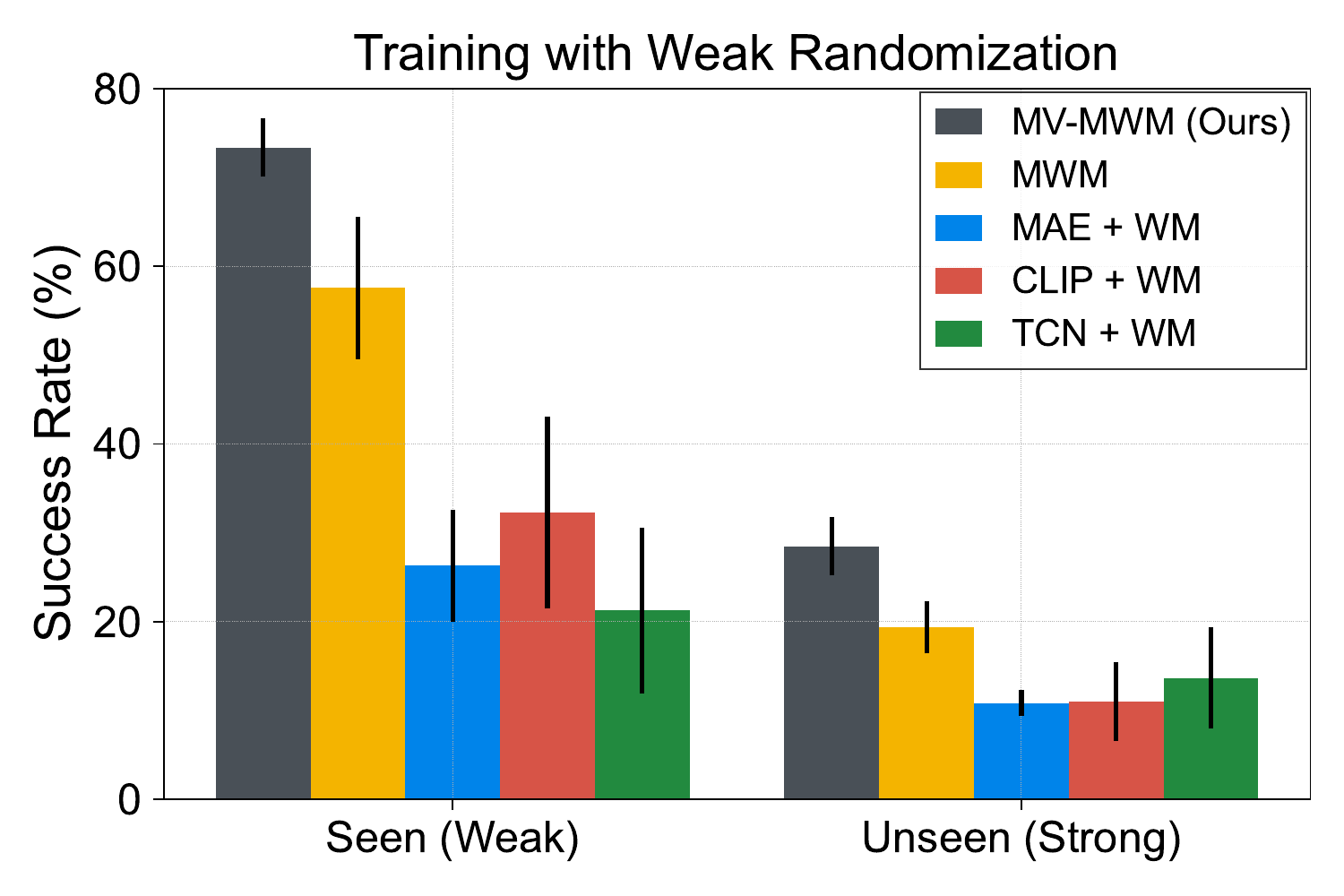}
    \label{fig:weak_aggr}
}
\subfigure{
    \includegraphics[width=0.46\textwidth]{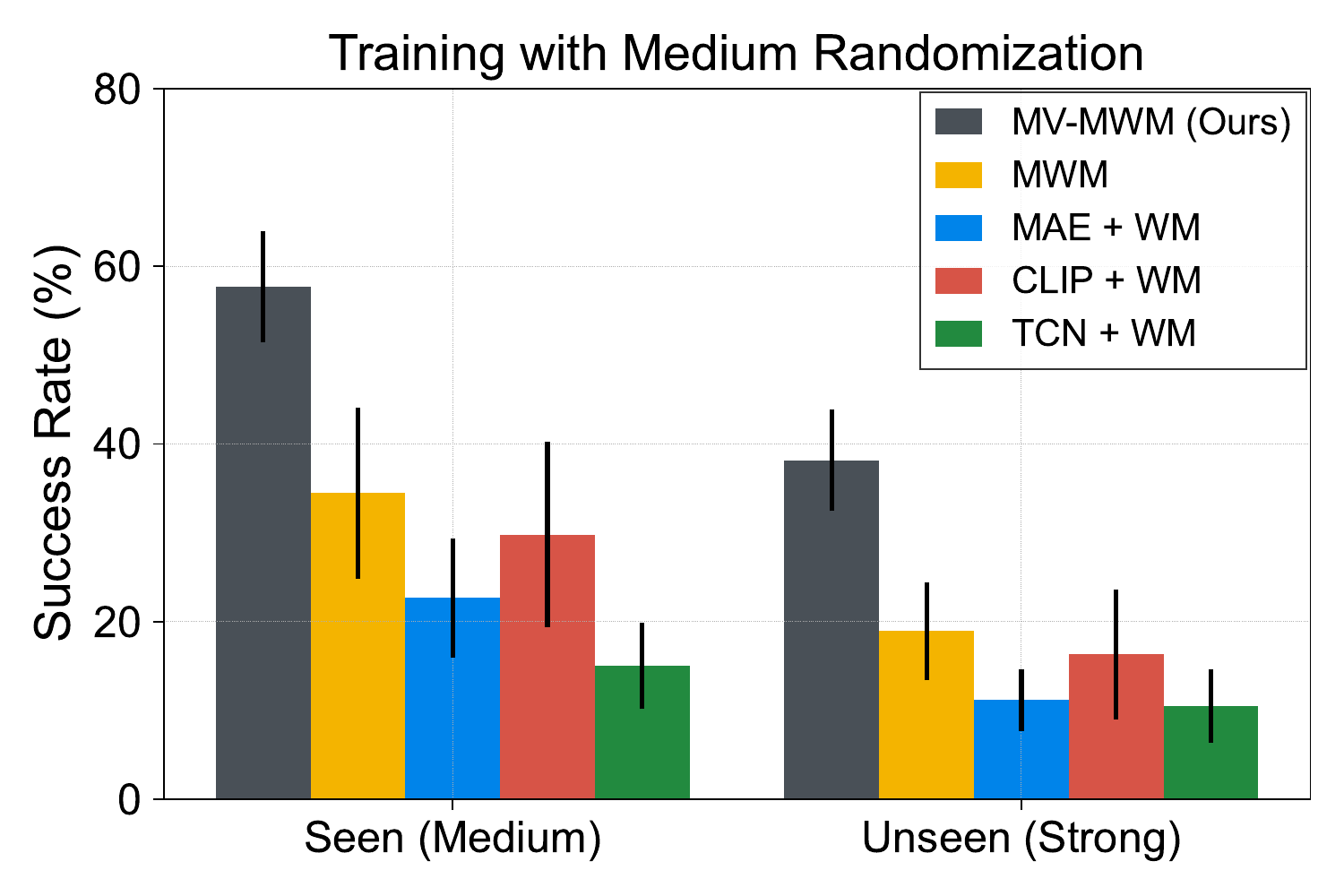}
    \label{fig:medium_aggr}
}\vspace{-0.15in}\\
\centering
\subfigure{
\includegraphics[width=0.98\textwidth]{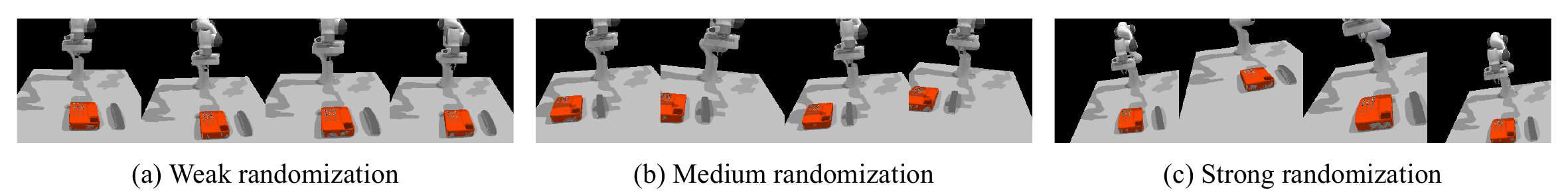}
\label{fig:rand_examples}
}\vspace{-0.25in}
\caption{Aggregate success rate on two viewpoint-robust control tasks under both seen and unseen viewpoint randomization setups exemplified in the second row.
We report the average success rate evaluated using the last five model checkpoints.
The result shows the mean and stratified bootstrap confidence interval across 8 runs.
We provide the learning curve for all tasks in \cref{appendix:full_results}.
}\vspace{-0.1in}
\label{fig:main_viewpoint_randomization}
\end{figure*}

\subsection{Viewpoint-Robust Control}
\label{sec:experiments_viewpoint_robust}
\paragraph{Problem setup}
We consider a viewpoint-robust control setup where we learn a policy robust to camera perturbations, which can enable us to deploy robots to real-world without a tedious camera calibration (see \cref{sec:experiments_sim_to_real} for our real-robot experiments).
However, we observe that learning to solve tasks under hard viewpoint randomization is a very challenging problem.
To address this, we propose to utilize multi-view representation learning by generating multiple randomized viewpoints and learn representations with them (see \cref{fig:mvmwm_overview} for illustration).
Then we train a single-view agent on randomized viewpoints upon these representations to solve tasks with randomized cameras.

\paragraph{Experimental setup}
We randomize the position and orientation of the front camera at every episode.
We construct three randomization type, which represents how strongly viewpoints are randomized: \textit{weak}, \textit{medium}, and \textit{strong} as exemplified in \cref{fig:main_viewpoint_randomization}.
For faster experimentation, we modify the tasks to be more easier by making a target object be located upright (\textit{i.e.,} not rotated related to the table).
Because we aim to evaluate viewpoint-robust control agents to solve real-world robotic tasks, we also modify the colors of a simulated workspace to be similar to colors of a real-world setup and apply brightness and contrast augmentation to videos throughout training.
For evaluation, we train all agents on weak and medium randomization setups and report the performance under both seen (\textit{i.e.,} weak and medium) and unseen (\textit{i.e.,} strong) randomization setups.

\paragraph{Results} 
\cref{fig:main_viewpoint_randomization} shows the performance of visual control agents on both seen and unseen randomization setups.
We first observe that our single-view baseline, MWM, can achieve competitive training performance on weak randomization setup.
This aligns with the observation of \citet{sadeghi2018sim2real} that shows a recurrent policy can be robust to randomized viewpoints.
However, we observe that the performance of MWM significantly degrades as randomization gets stronger, implying that the recurrent architecture alone is not enough for viewpoint-robust control.
We find that other baselines also struggle on medium randomization setup, failing to improve their generalization performance on unseen randomization setup.
On the other hand, MV-MWM learns to solve the tasks under medium randomization and outperforms all baselines on both seen and unseen setups (see \cref{fig:main_no_rand} for aggregate performance).
We also find that TCN + WM trained on randomized viewpoints achieves non-zero success rates, in contrast to results with the front and wrist cameras in \cref{sec:experiments_multi_view_single_view}.
We hypothesize this is because contrastive learning becomes easier when using similar viewpoints.
Nonetheless, MV-MWM largely outperforms TCN + WM, which highlights the benefit of our simple yet effective masked view reconstruction objective.

\subsection{Real-Robot Manipulation via Sim-to-Real Transfer}
\label{sec:experiments_sim_to_real}
\paragraph{Setup} We also evaluate the zero-shot performance of 
agents trained in simulation for solving real-world manipulation tasks.
We deploy the agents trained under medium randomization to solve the Pick Up Cup task without camera calibration and adaptation procedures.
We conduct experiments with the Franka Research 3 and use MoveIt2 library based on ROS 2 framework for controlling the arm.
We use RGB observations from RealSense D435f camera.

\begin{table}[t!] \centering
    \vspace{-0.075in}
    \caption{Zero-shot sim-to-real transfer performance of viewpoint-robust control agents trained in simulation for solving a real-world visual robotic manipulation task.}
    \vspace{0.1in}
    \begin{minipage}{0.48\textwidth}
        \centering
        \includegraphics[width=0.44\textwidth]{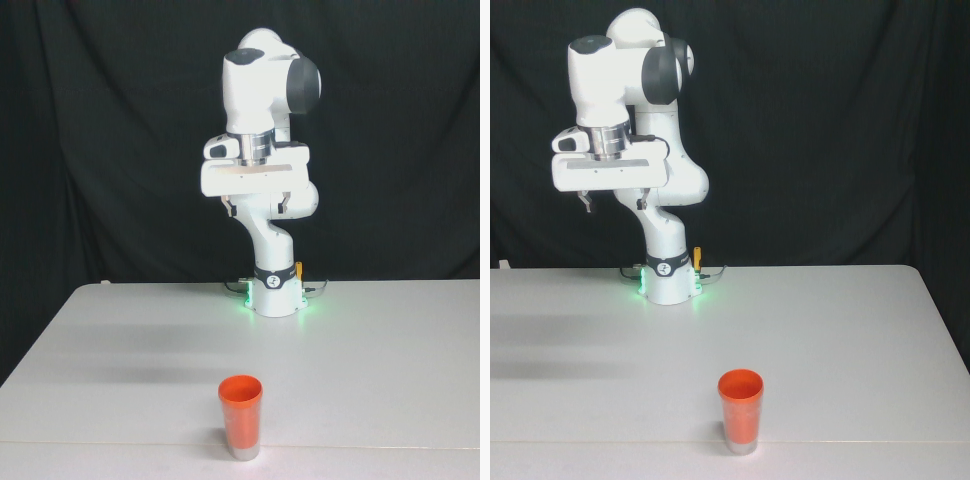}
        \centering
        \small
        \begin{tabular}[b]{lc}
        \toprule
         Method & \multicolumn{1}{c}{Success rate} \\ 
         \midrule
         MAE + WM & \multicolumn{1}{l}{7.1\%} \\
         MWM & \multicolumn{1}{l}{11.3\%} \\
         MV-MWM & \multicolumn{1}{l}{\textbf{74.7\%}}\\
         \bottomrule
        \end{tabular}
    \end{minipage}
\label{tbl:main_sim2real_fixed}
\vspace{-0.1in}
\end{table}

\begin{figure*} [t!] 
    \centering
    \includegraphics[width=0.98\textwidth]{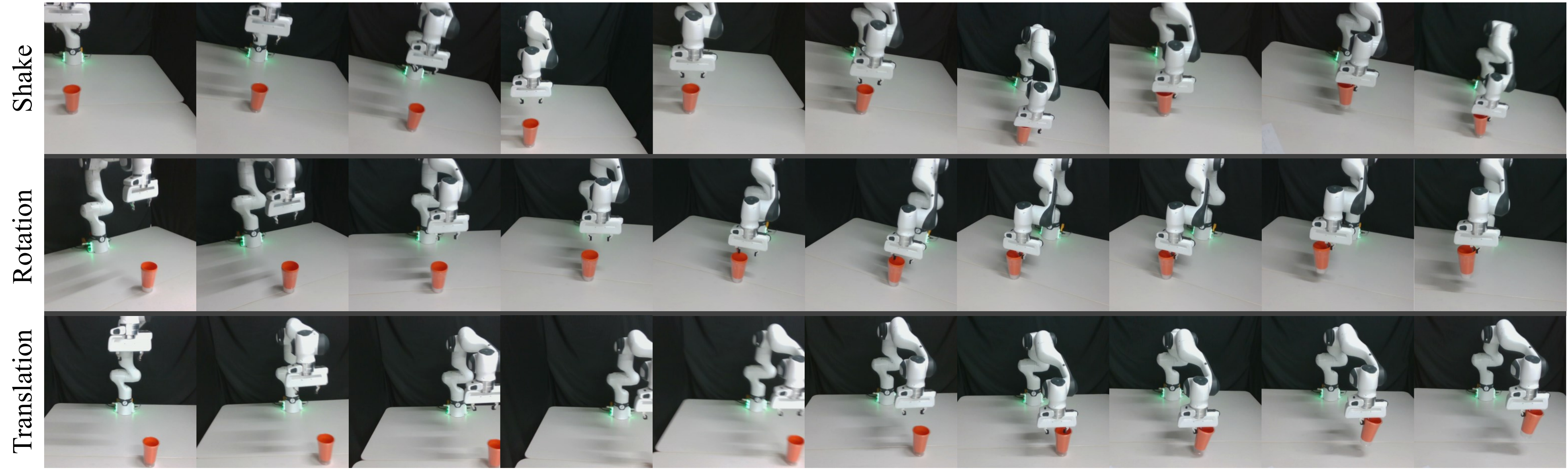}
    \vspace{-0.1in}
    \caption{We study the robustness of MV-MWM to camera perturbations in a real-world by considering an extreme setup where the agent operates on a hand-held camera subject to shake, rotation, and translation.
    We observe that MV-MWM can solve the task using only RGB observations without proprioceptive states and any adaptation procedure.
    Best viewed as videos provided in the webpage: \url{https://sites.google.com/view/mv-mwm}. 
    }
    \label{fig:main_sim2real_handheld}
    \vspace{-0.15in}
\end{figure*}

\begin{figure*} [t!] \centering
    \subfigure[View-masking]
    {
    \includegraphics[width=0.31\textwidth]{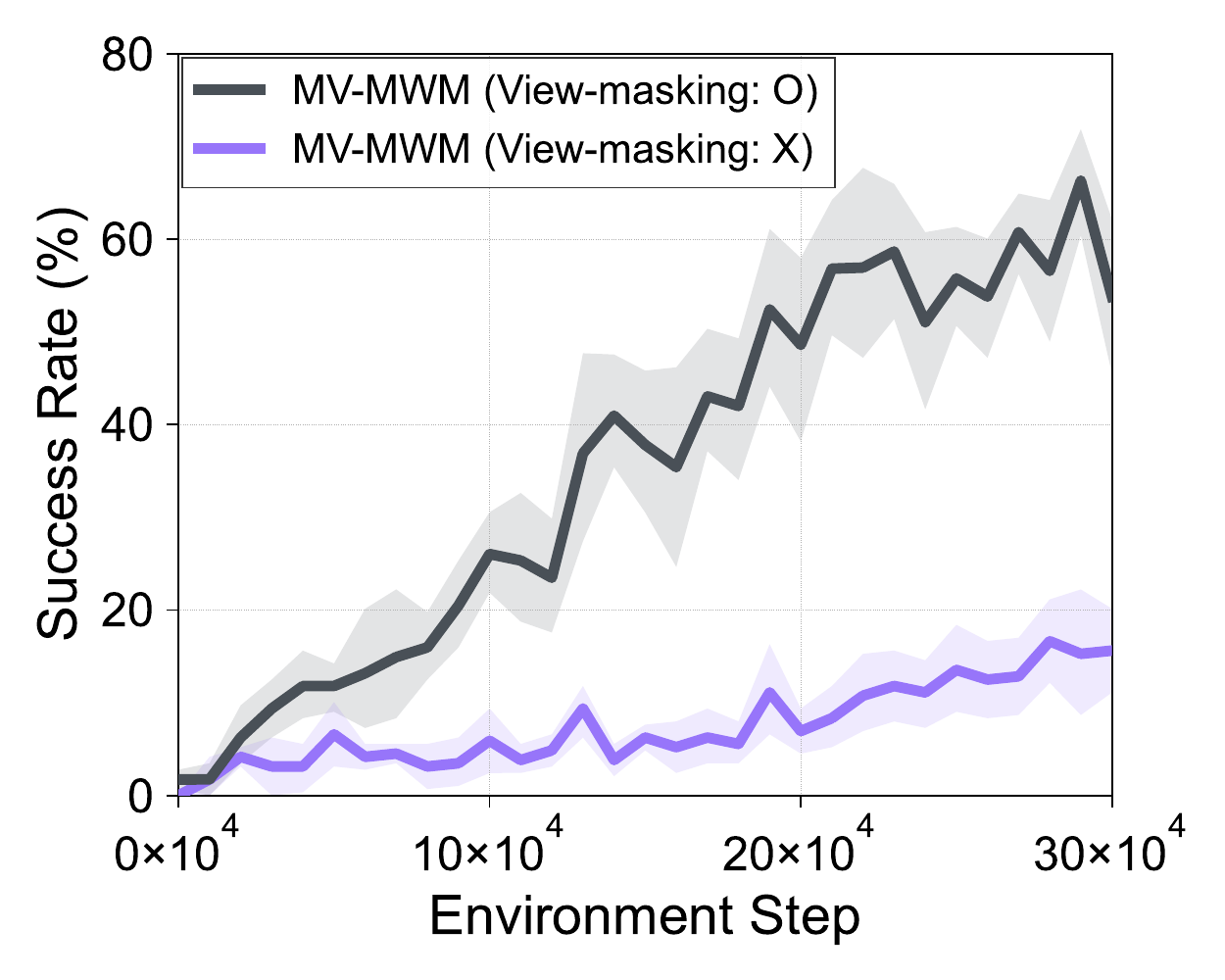}
    \label{fig:analysis_view_masking}}
    \subfigure[Video autoencoding]
    {
    \includegraphics[width=0.31\textwidth]{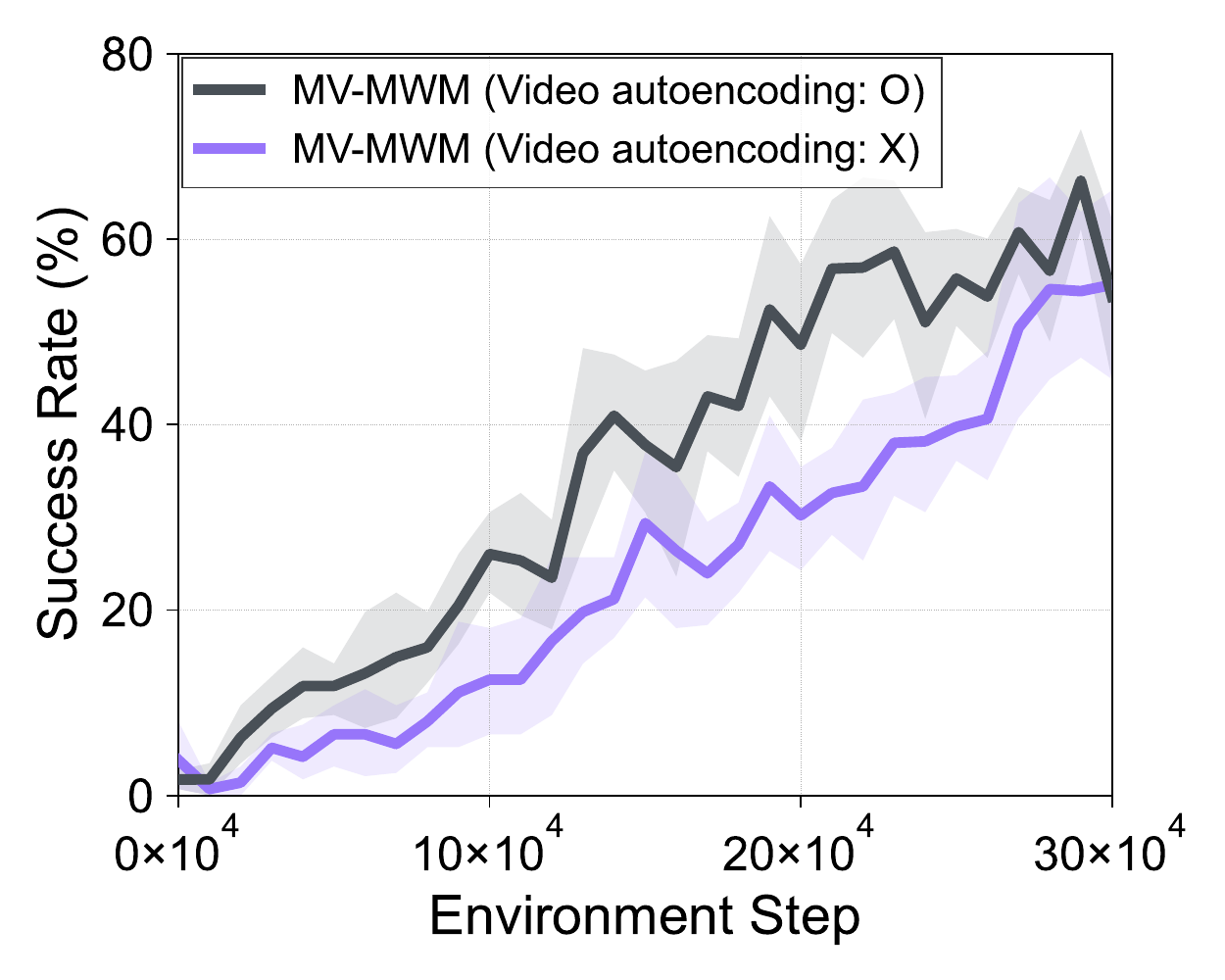}
    \label{fig:analysis_video_autoencoding}}
    \subfigure[Masking ratio]
    {
    \includegraphics[width=0.31\textwidth]{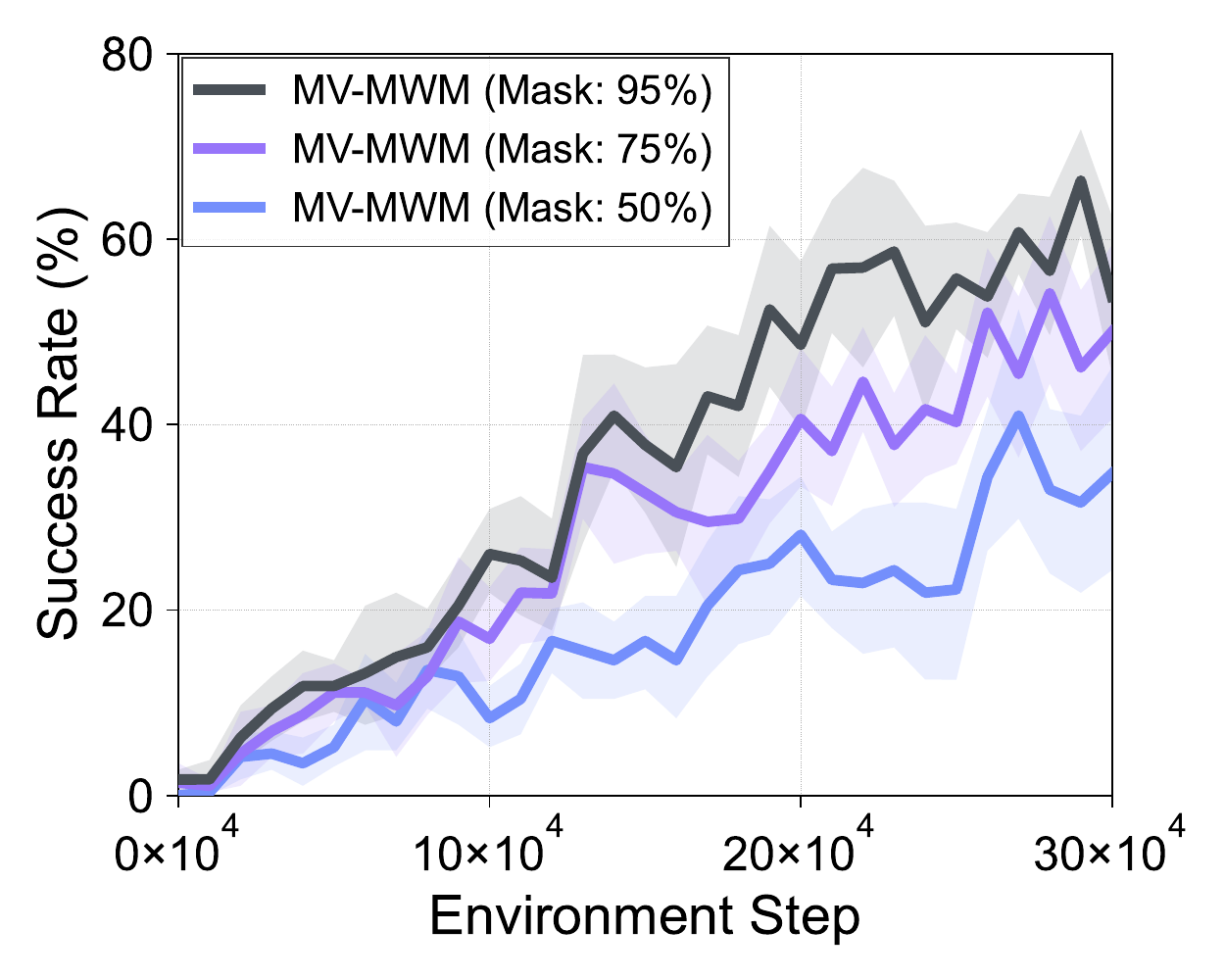}
    \label{fig:analysis_masking_ratio}}
    \vspace{-0.125in}
    \caption{Learning curves of single-view visual control agents operating on the front camera for solving three manipulation tasks from RLBench~\citep{james2020rlbench}, investigating the effect of (a) view masking, (b) video autoencoding and (c) masking ratio.
    The solid line and shaded regions represent the mean and stratified bootstrap confidence interval across 12 runs.}
    \vspace{-0.025in}
    \label{fig:analysis2}
\end{figure*}

\paragraph{Results}
For evaluation, we measure the success rate across 3 viewpoints and 20 randomized cup positions for each viewpoint.
\cref{tbl:main_sim2real_fixed} shows the real-world performance of MV-MWM and two baselines we selected for their overall similarity to our method.
We find that MV-MWM can solve the task without camera calibration and largely outperforms all baselines. 
We further evaluate MV-MWM on an extreme setup where we use a hand-held camera which is subject to shake, rotation, and translation while solving the task.
We note that viewpoint is not randomized within the episode for training our world model, which makes the setup more challenging.
Surprisingly, \cref{fig:main_sim2real_handheld} shows that MV-MWM can solve the task on this challenging setup only using RGB observations, which showcases the effectiveness of our method for real-world visual robotic manipulation.
Video demonstrations are available at our webpage: \url{https://sites.google.com/view/mv-mwm}.

\subsection{Ablation Study and Analysis}

\paragraph{Effect of view-masking}
We evaluate the performance of MV-MWM with and without the proposed view-masking scheme in \cref{fig:analysis_view_masking}.
Specifically, we consider a baseline trained with the uniform-masking scheme that mask random features from multi-view inputs, \textit{i.e.,} MV-MWM (View-masking: X).
We observe that performance largely degrades without view-masking, which shows that view-masking is crucial for multi-view representation learning.

\paragraph{Effect of video autoencoding}
\cref{fig:analysis_video_autoencoding} shows that sample-efficiency of our method significantly improves with video autoencoding.
This shows that enabling the model to have access to unmasked frames of the same view ease the difficulty of masked view reconstruction,
making the combination of view-masking and video autoencoding synergistic.

\paragraph{Masking ratio}
\cref{fig:analysis_masking_ratio} shows that MV-MWM performance keeps increasing with a higher masking ratio.
We hypothesize this is because spatial information redundancy~\citep{he2021masked} is more significant in visual observations from manipulation tasks than natural images.
This also aligns with the observation of prior work~\citep{tong2022videomae} where 90\% masking ratio has shown to be effective for videos.

\paragraph{Scaling property}
In~\cref{fig:analysis_training_ratio} and \cref{fig:analysis_model_size} available at \cref{appendix:additional_expriments}, we also investigate whether MV-MWM can be further scaled up for better performance by improving the number of gradient steps and increasing the model size. We find that training with more gradient steps and larger models can further improve the sample-efficiency.

\paragraph{Data augmentation for viewpoint-robust control}
To investigate the importance of using visual observations from randomized cameras, we consider a baseline that uses images perturbed with data augmentation (\textit{i.e.,} rotation, translation, brightness, and contrast) for multi-view representation learning.
As shown in \cref{fig:data_aug} available at \cref{appendix:additional_expriments}, we observe that using the images from physically perturbed images largely outperforms the baseline based on data augmentation.
This is because such a randomized camera can provide images from different perspectives that contain additional information which is not available from the single, fixed camera viewpoint.
On the other hand, data augmentation fails to give such information useful for implicitly capturing 3D information of a robot workspace.
Given this result, investigation into how to set up a real-world robot learning setup with randomized cameras would be an interesting and important future research direction.

We provide learning curves for ablation study in~\cref{appendix:full_results} and additional analysis in~\cref{appendix:additional_expriments}.

\begin{figure}
\centering
\subfigure{
\includegraphics[width=0.46\textwidth]{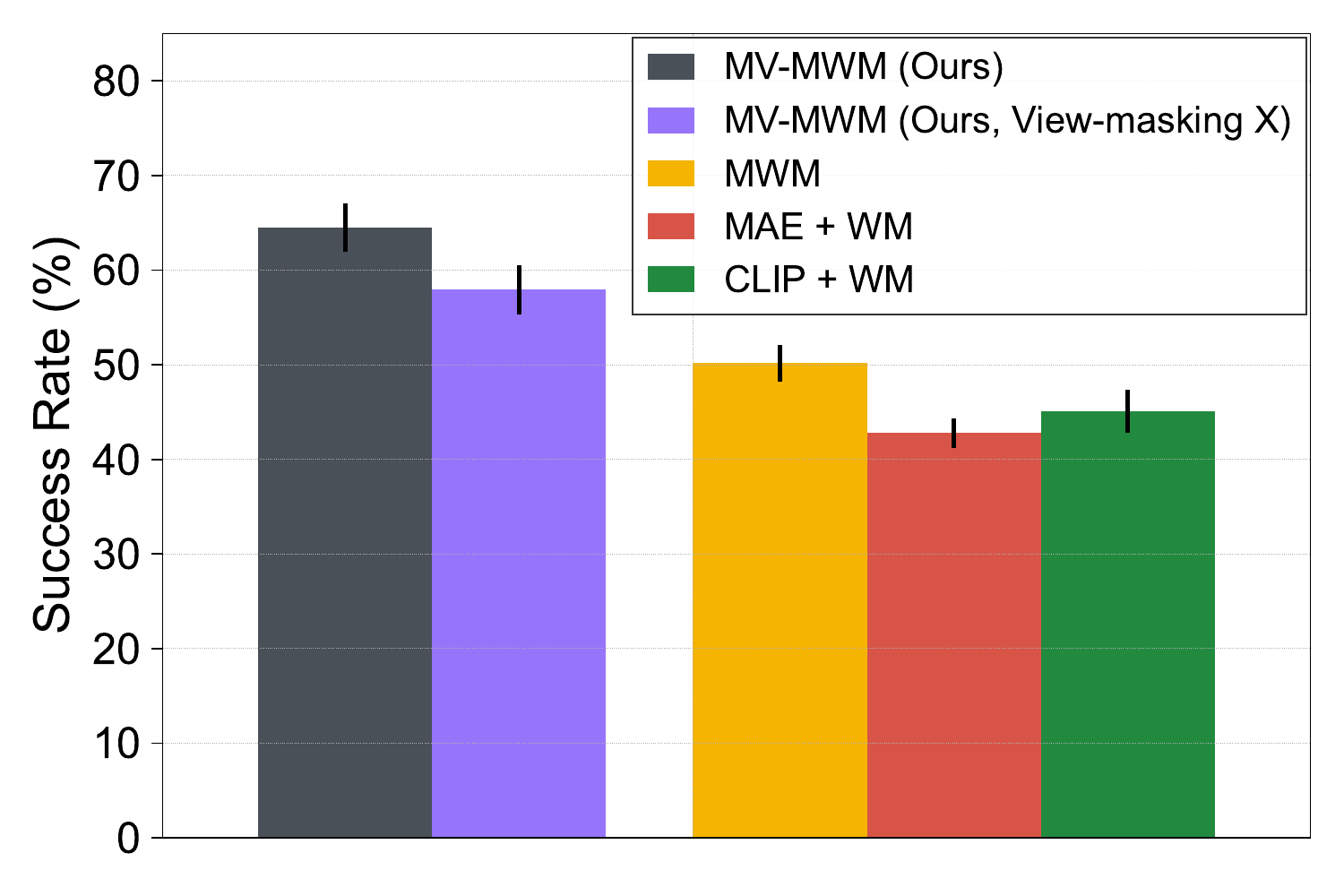}
}
\vspace{-0.225in}
\caption{Aggregate success rate of imitation learning agents on five single-view control tasks.
The result shows the mean and stratified bootstrap confidence interval across 20 runs.}
\label{fig:imitation}
\vspace{-0.075in}
\end{figure}

\subsection{Imitation Learning Experiments}
\paragraph{Setup} 
Finally, we investigate the effectiveness of our multi-view representation learning method on imitation learning (IL) setup, which is a widely-considered setup in the field of robotics.
We consider a setup where we train visual control agents only with behavior cloning (BC) instead of RL to solve the tasks.
Specifically, we consider the single-view control setup as in \cref{sec:experiments_multi_view_single_view}, using the same set of five manipulation tasks.
For training, we use 100 expert demonstrations collected with the scripted policies available from the RLBench simulator.
Unlike in the RL setup, we follow a setup of prior work that utilizes the output next best gripper pose.
We use the same architecture as in RL experiments but find that using more stronger L2 weight decay largely improves the performance by preventing the overfitting.
We also disable video autoencoding in IL experiments because the model is trained until convergence in this setup so that there is no problem from training difficulty from the view-masking as in RL setup.
For baselines, we consider the same set of baselines as previous experiments but exclude TCN due to its overall low performance.
For evaluation, we measure the average success rate over 500 episodes where the object position is randomized every episode.

\paragraph{Results}
In~\cref{fig:imitation}, we observe that the trend in IL experiments is the same as in prior experiments, where our method, MV-MWM, outperforms all the baselines.
In particular, MV-MWM outperforms MWM, which uses the same amount of training data, by a large margin (14.35\%p).
We also find that the proposed view-masking scheme significantly improves the performance,
\textit{e.g.,} the view-masking scheme improves the performance from 57.92\% to 64.48\%.
This experimental result shows that the benefit of multi-view representation learning along with the proposed view-masking scheme is consistent across both RL and IL setups, highlighting the effectiveness of our method for diverse, practical robotic manipulation setups.

\section{Discussion}
\paragraph{Limitation and future directions}
One limitation of our work is that considered tasks are simple in that they do not require a long-horizon planning and involve a single object.
Scaling up our framework to solve more challenging tasks is a direction we hope to investigate in future works. 
For instance, incorporating a more scalable architecture \citep{jaegle2021perceiver} along with large-scale pre-training on large datasets \citep{deng2009imagenet,dasari2019robonet} would be an interesting direction that can improve the generalization capability of visual manipulation system while having the benefit of multi-view representation learning.
Another interesting direction would be to design a viewpoint randomization setup for real-world robot learning, where it is non-trivial to aggressively randomize viewpoints as we have done in sim-to-real transfer experiments.

\vspace{-0.1in}
\paragraph{Conclusion}
We present Multi-View Masked World Models, a reinforcement learning framework that learns multi-view representations and utilize them for diverse visual robotic manipulation setups.
We conduct extensive experiments and find that our method consistently outperforms various baselines across a range of tasks in both simulation and real-world environments.
We hope this work encourages future research to further explore the potential of multi-view representation learning for visual robotic manipulation.

\paragraph{Acknowledgements} We would like to thank Danijar Hafner, Sangwoo Mo, Youngwoon Lee, Xingyu Lin, Hao Liu, Jongjin Park, Carlo Sferrazza, Sihyun Yu, and anonymous reviewers for helpful comments.
This work was partially supported by Institute of Information \& Communications Technology Planning \& Evaluation (IITP) grant funded by the Korea government (MSIT) (No.2022-0-00953, Self-directed AI Agents with Problem-solving Capability; No.2019-0-00075, Artificial Intelligence Graduate School Program (KAIST)) and KAIST-NAVER Hypercreative AI Center.
This material is based upon work supported by the Google Cloud Research Credits program with the award (N6U8-0LLR-JDTW-JNWP). We also appreciate NVIDIA Corporation (\url{https://www.nvidia.com/}) and Cirrascale Cloud Services (\url{https://cirrascale.com/}) for providing compute resources.

\clearpage

\bibliography{main}
\bibliographystyle{icml2023}

\clearpage

\appendix

\onecolumn
\section{Implementation Details}
\label{appendix:implementation}

\paragraph{RLBench details}
For RLBench experiments, we designed dense rewards for five manipulation tasks used in our experiments. We first construct waypoints where the robot should reach (\textit{i.e.,} phone and base positions in Phone On Base). Then we define the reward to be the distance between the gripper and the next checkpoint.
In tasks that does not need rotation in solving (\textit{i.e.,} Phone On Base, Pick Up Cup, Take Umbrella Out of Umbrella Stand, Put Rubbish in Bin), we disable rotation to reduce redundancy in exploring rotating actions. For disabling rotation, we use a path planner with identity quaternion to force the robot to be in an upright position. In these tasks, we train the RL agent to output relative change in (x, y, z) position.
For tasks that requires rotation (\textit{i.e.,} Stack Wine), RL agent is trained to output relative quaternion changes as well as (x,y,z) position changes.
For viewpoint-robust control, 
we ease the difficulty of tasks. For Phone On Base, we make phone and base be located upright (\textit{i.e.,} not rotated related to the table). For Pick Up Cup, we make the distractor cup colors be fixed as yellow instead of changing the color for each episode. For such tasks, we append asterisk (*) after the (shortened) task name; \textit{i.e.,} Phone* and Cup*.
For more details, we refer readers to the source code we have attached.

\paragraph{Viewpoint randomization}
We randomize viewpoint by adjusting position and orientation of front camera. Specifically, we randomize five parameters of camera position and orientation: $\theta, \phi, d, h, \psi$, which represents as follows:

\begin{itemize}[leftmargin=5.5mm]
    \item $\theta$: angle that determines how camera is moved clockwise (from the front camera) with respect to the origin.
    \item $\phi$: angle that determines how camera is tilted downward.
    \item $\psi$: angle that determined how camera is rolled clockwise. 
    \item $d$: distance from the origin of the simulator. 
    \item $h$: height of camera from the floor.
\end{itemize}
\begin{figure*} [h] \centering
    \subfigure[Images with varying $\theta$]
    {
    \includegraphics[width=0.31\textwidth]{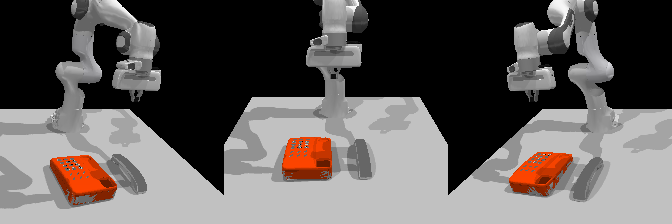}
    \label{fig:theta}}
    \subfigure[Images with varying $\phi$]
    {
    \includegraphics[width=0.31\textwidth]{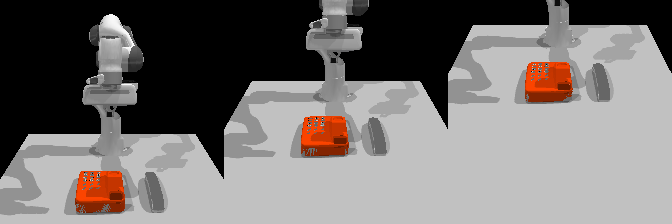}
    \label{fig:phi}}
    \subfigure[Images with varying $\psi$]
    {
    \includegraphics[width=0.31\textwidth ]{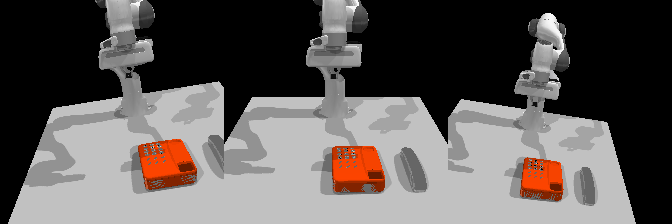}
    \label{fig:psi}}
    \subfigure[Images with varying $d$]
    {
    \includegraphics[width=0.31\textwidth]{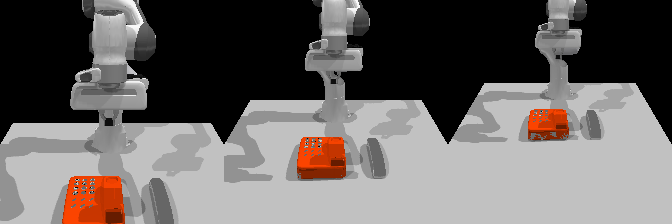}
    \label{fig:radius}}
    \subfigure[Images with varying $h$]
    {
    \includegraphics[width=0.31\textwidth]{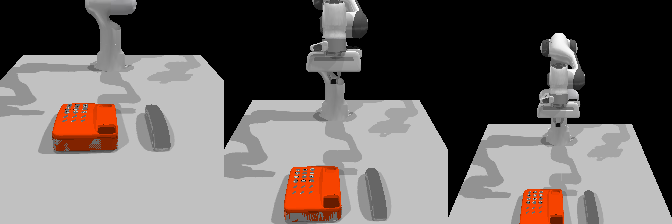}
    \label{fig:height}}
    
    \caption{Rendered images with varying randomization parameters.}
    \label{fig:viewpoint_randomization_examples}
\end{figure*}

We exemplify how each parameter affects viewpoint in~\cref{fig:viewpoint_randomization_examples}. By randomizing aforementioned five parameters, we design three randomization types (weak, medium, and strong) as follows.

\begin{itemize}[leftmargin=5.5mm]
    \item weak:
    \begin{gather*}
    \quad \theta \sim [-5 \degree, 5 \degree], \quad \phi \sim [26 \degree, 28 \degree],\quad \psi \sim [-5 \degree, 5 \degree], \quad d \sim [1.25 \degree, 1.45 \degree], \quad h \sim [1.5, 1.7]
    \end{gather*}
    \item medium: 
    \begin{gather*}
    \quad \theta \sim [-7.5 \degree, 7.5 \degree], \quad \phi \sim [25.5 \degree, 28.5 \degree],\quad \psi \sim [-7.5 \degree, 7.5 \degree], \quad d \sim [1.2, 1.5], \quad h \sim [1.45, 1.75]
    \end{gather*}
    \item strong: 
    \begin{gather*}
    \quad \theta \sim [-10 \degree, -7.5 \degree] \cup [7.5 \degree, 10 \degree] , \quad \phi \sim [25 \degree, 25.5 \degree]  \cup [28.5 \degree, 29 \degree],\quad \psi \sim [-10 \degree, -7.5 \degree] \cup [7.5 \degree, 10 \degree], \\ \quad d \sim [1.15, 1.2] \cup [1.5, 1.55], \quad h \sim [1.4, 1.45] \cup [1.75, 1.8]
    \end{gather*}
\end{itemize}

Note that we design the strong randomization, whose distribution does not overlap with that of weak or medium randomization, in order to evaluate the performance under unseen viewpoint conditions. When we start episodes to collect transitions for training, we sample two random viewpoints under the randomization type. In evaluation phase, we sample one random viewpoint under the randomization type. The sampled viewpoints are maintained for an episode, and we re-sample viewpoints when new episode begins.

\paragraph{Architecture and optimization details} Our architecture is based on the publicly available source code of \citet{seo2022masked}, which is implemented with tfimm\footnote{\url{https://github.com/martinsbruveris/tensorflow-image-models}} library. We use 8-layer ViT encoder and 6-layer ViT decoder.
For each view and time step, we introduce additional 1D learnable parameters that have the same embedding size as transformer blocks. We add these parameters to 2D fixed sin-cos embeddings and add them to features.
We note that these parameters are shared across the same times and the same views.
We do not introduce separate parameters for randomized viewpoints in our viewpoint-robust control experiments.
For optimization, we use Adam optimizer~\citep{kingma2014adam} with the learning rate of $3e-4$, the weight decay of $1e-6$, and the batch size of 1024. For training MV-MAE, we apply warm-up learning rate scheduling over initial 2500 gradient steps from learning rate of 0. We take 1 gradient step per every 16 environment steps.
We follow the training schemes and details of \citet{seo2022masked} regarding the architecture, unless otherwise specified.

\paragraph{Computation} 
We use 24 CPU cores (Intel Xeon CPU @ 2.2GHz) and 1 GPU (NVIDIA A100 40GB GPU) for our experiments.
We find that there is no significant difference between all methods with regard to wall time because rendering speed of the RLBench simulator is a bottleneck rather than algorithmic difference. Running experiments over 300k environment steps for MV-MWM takes approximately 12 hours. 

\paragraph{Hyperparameters} We report the hyperparameters used in our experiments in~\cref{tbl:hyperparameters}.

\begin{table}[h]
\caption{Hyperparameters used in our experiments. Unless otherwise specified, we use the same hyperparameters used in MWM~\citep{seo2022masked}.}
\vskip 0.15in
\begin{center}
\begin{tabular}{ll}
\toprule
\textbf{Hyperparameter} & \textbf{Value}  \\
\midrule
\textit{Representation learning} \\
\midrule
Image observation  & $96 \times 96 \times 3$ \\
Image normalization  & Mean: $(0.485, 0.456, 0.406)$, Std: $(0.229, 0.224, 0.225)$ \\
Autoencoder batch size    & $1024$  \\ 
Autoencoder initialization steps    & $10000$  \\
Autoencoder warm-up steps    & $2500$  \\ 
Autoencoder learning rate    & $3\cdot10^{-4}$  \\ 
Autoencoder masking ratio    & $0.95$ \\ 
Autoencoder ViT encoder size    & $8$ layers, 4 heads, 256 units  \\
Autoencoder ViT decoder size    & $6$ layers, 4 heads, 256 units  \\ 
\midrule
\textit{Behavior learning} \\
\midrule
Action repeat    & $1$ \\
Max episode length    & $150$ \\
Early episode termination   & True (when path planner fails) \\
Reward normalization   & True \\
Number of expert demonstrations   & 50 (single-view and multi-view control), 100 (viewpoint-robust control) \\
World model batch size   & $36$ \\
World model expert batch size   & $12$ \\
World model sequence length  & $50$ \\
World model tradeoff ($\beta$)  & $1.0$ \\
World model ViT encoder size  & $2$ layers, $4$ heads, 128 units \\
World model ViT decoder size  & $2$ layers, $4$ heads, 128 units \\
\bottomrule
\end{tabular}
\label{tbl:hyperparameters}
\end{center}
\vskip -0.1in
\end{table} 

\newpage

\section{Baselines}
\label{appendix:baselines}
\subsection{Masked world models}
Masked world models (MWM; \citealt{seo2022masked}) is a visual model-based reinforcement learning framework which decouples visual representation learning and dynamicd model learning.
Specifically, MWM trains a self-supervised vision transformer (ViT; \citealt{dosovitskiy2020image}) to reconstruct pixels given masked convolutional features for representation model.
Then, a world model is trained on top of the frozen visual representations.
The major methodological difference between MV-MWM and MWM is that MV-MWM learns cross-view information from multiple viewpoints by training a multi-view masked autoencoder but MWM only considers visual information within each viewpoint. 
Specifically, MWM does not employ pair information between multiple viewpoints in representation learning, but considers images from different viewpoints as independent instances in training.
Whereas, MV-MWM considers cross-view information along with a synergistic combination of view-masking and video-autoencoding in contrast to MWM that trains an image autoencoder with uniform masking.
With this methodological difference, MV-MWM allows for learning world models that can be useful for a range of important and practical visual robotic manipulation setups.

\subsection{Time contrastive network}
 Time contrastive network (TCN; \citealt{sermanet2018time}) is a contrastive approach that learns view-invariant representations by attracting the representations of simultaneous viewpoints but making the representations from the same viewpoints be far located.
 For a given (anchor) frame in one viewpoint, we sample a positive and negative frame as follows. For the positive frame, we use the frame that has the same timestep as the given anchor frame but from another viewpoint. For the negative frame, we choose a frame that is a temporally faraway frame from the same viewpoint. Specifically, we sample a random frame among frames that are at least 30 timesteps away from the anchor frame. After building a triplet of anchor, positive, and negative frame, we train the encoder model with triplet loss \citep{schroff2015facenet}, which is formulated as follows:
 \begin{align}
 \begin{split}
     \mathcal{L}_{\mathtt{TCN}} = \mathtt{max} (\Vert f(o^{a}) - f(o^{p}) \Vert_{2}^{2} - \Vert f(o^{a}) - f(o^{n}) \Vert_{2}^{2} + \alpha, 0), 
 \end{split}
 \end{align}
 where $\alpha$ is margin, $f(\cdot)$ refers embedding of a frame, and $o^{a}, o^{p}$, and $o^{n}$ are anchor, positive, and negative frame, respectively.
 In our implementation of TCN, we use same encoder architecture with MV-MWM for a fair comparison; 8-layer ViT encoder. For embedding $f(\cdot)$, we use class embedding from the last layer of ViT encoder. In the control phase, we freeze the encoder and use average pooled token embeddings as an input for the RL agent. For RL agent, we employ the same world model and policy architecture with MV-MWM for a fair comparison. 
 
\subsection{Pretrained MAE and CLIP with world model (MAE+WM, CLIP+WM)}

Masked autoencoder (MAE; \citealt{he2021masked}) learns visual representation in a self-supervised manner by training a vision Transformer (ViT; \citealt{dosovitskiy2020image}) to reconstruct masked patches. 
Contrastive language-image pre-training (CLIP; \citealt{radford2021learning}) learns visual representation by aligning embedding of text and image with contrastive learning.
Recently, it has been shown that pre-training MAE with (in-the wild) large-scale dataset, and then training a control module on top of frozen representation can solve real-world robotic manipulation tasks \citep{radosavovic2022real, shridhar2021cliport}. To compare such training scheme with ours, we design MAE+WM and CLIP+WM, which learn world model upon frozen representation of pretrained MAE or CLIP. For a fair comparison, we match the world model size with that used in MV-MWM: \textit{i.e.,} 2 layers and 4 heads for the world model ViT encoder and decoder. For the pretrained MAE and CLIP, we employ open-sourced pretrained model from huggingface transformers library.\footnote{\url{https://huggingface.co/facebook/vit-mae-base}}\footnote{\url{https://huggingface.co/openai/clip-vit-base-patch32}}
For the input of pretrained MAE and CLIP, we use 224 $\times$ 224 RGB observation from each camera. 
Then, we feed 7 $\times$ 7 patches into the world model. 

\clearpage

\section{Full Experimental Results}
\label{appendix:full_results}

\subsection{Multi-view Control with Front and Wrist Camera}
\vspace{-.1in}
\begin{figure*}[h]
\centering
\subfigure{
\includegraphics[width=0.31\textwidth]{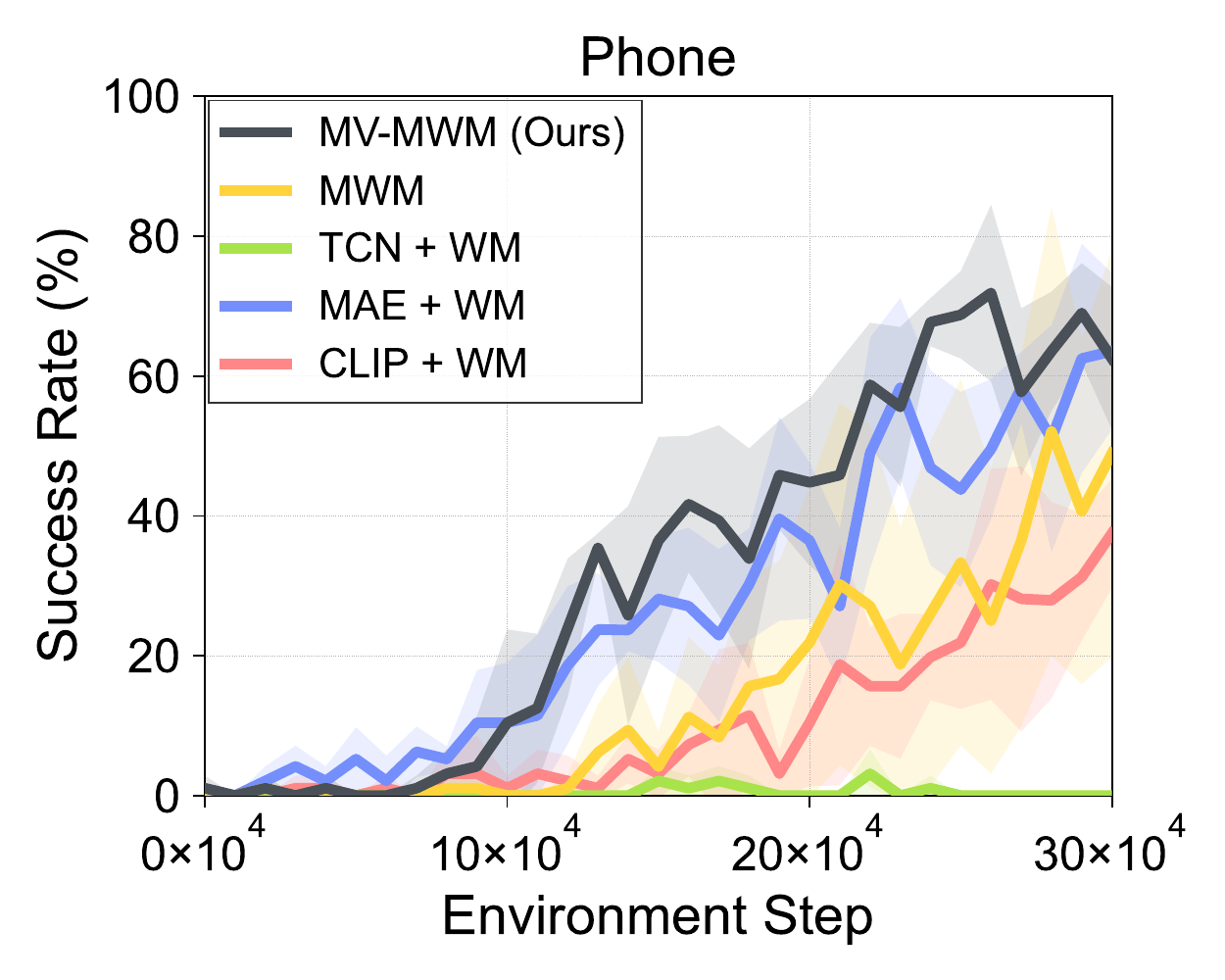}
\label{fig:multiview_phone}
}
\subfigure{
\includegraphics[width=0.31\textwidth]{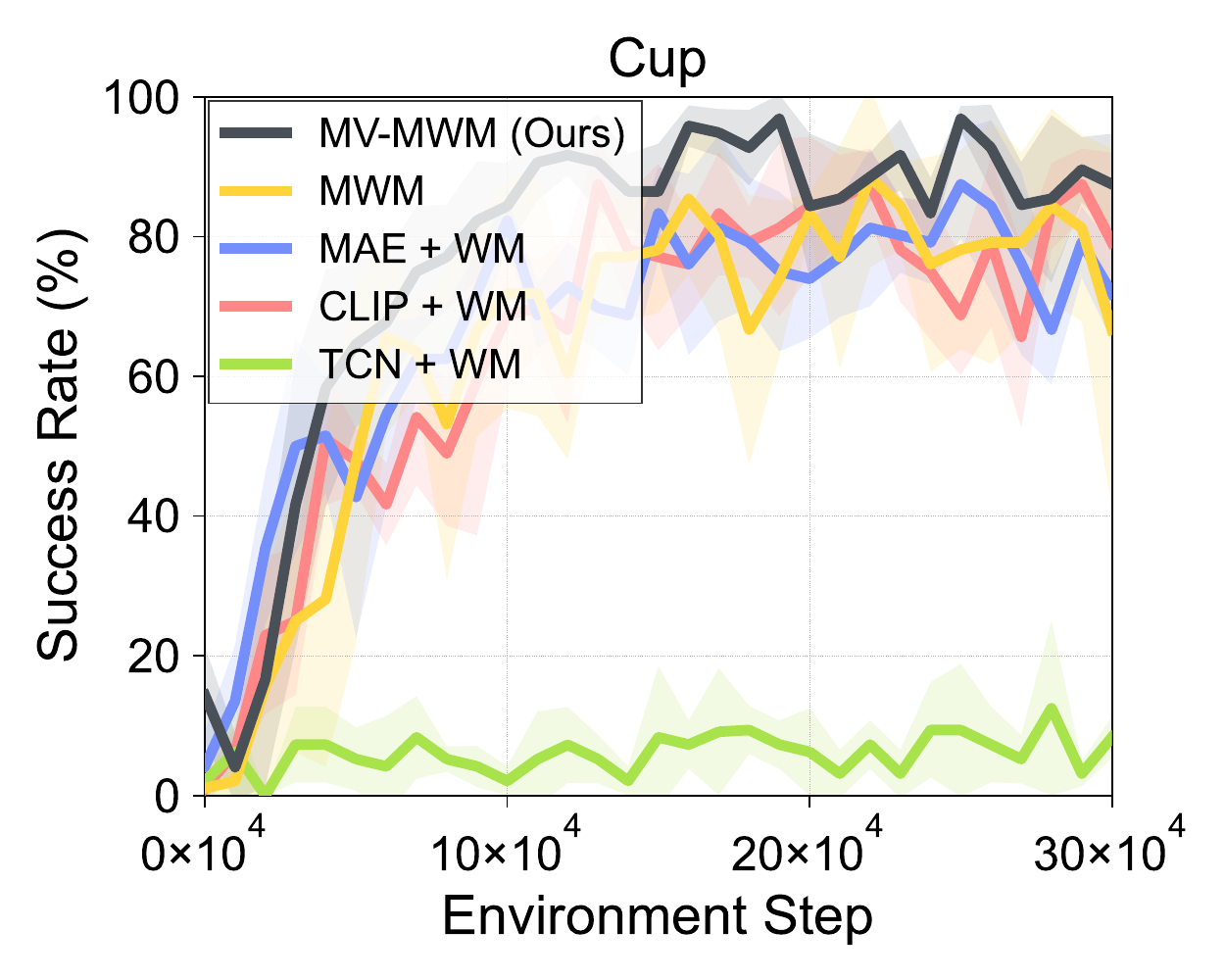}
\label{fig:multiview_cup}
}
\subfigure{
\includegraphics[width=0.31\textwidth]{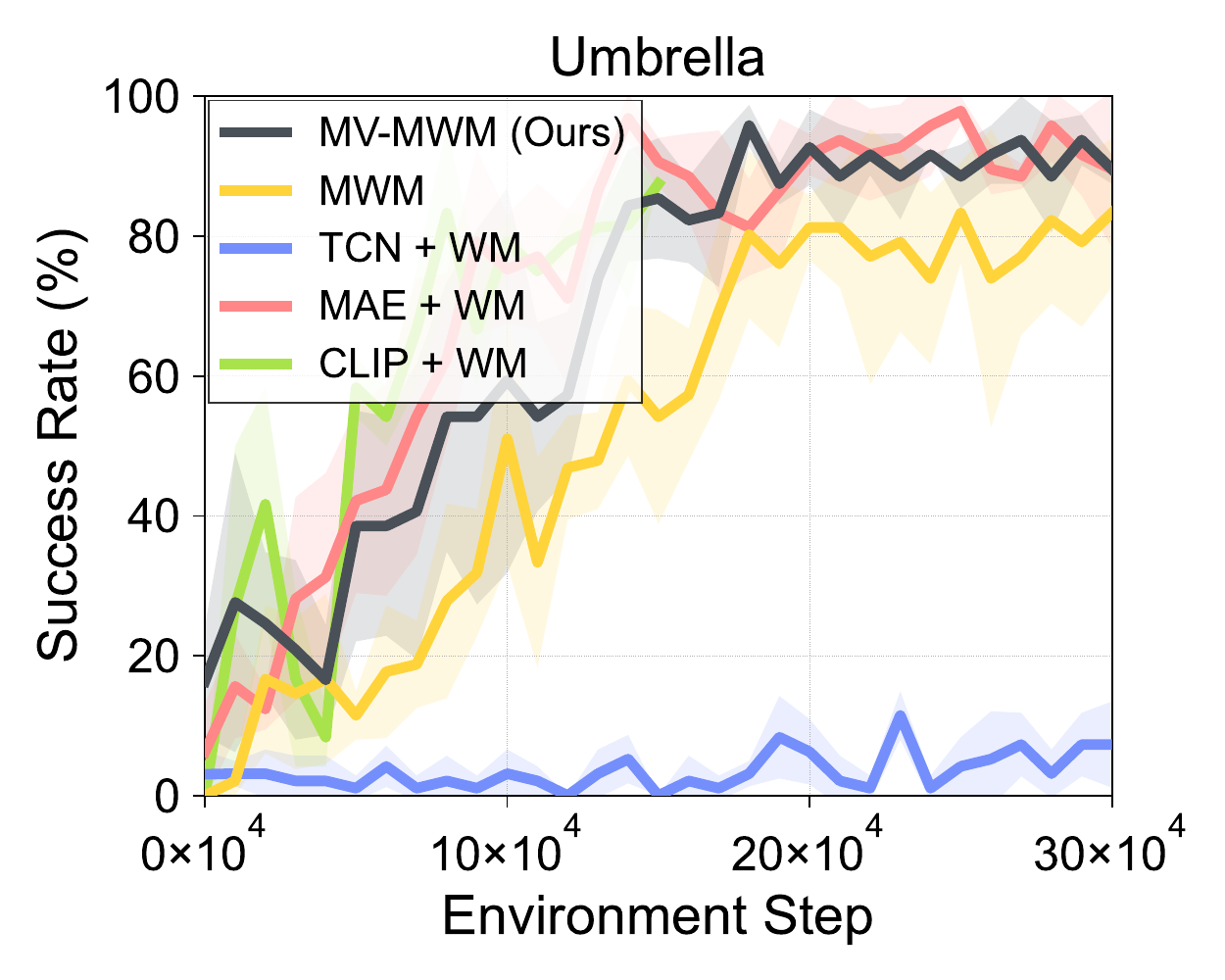}
\label{fig:multiview_umbrella}
}\\
\subfigure{
\includegraphics[width=0.31\textwidth]{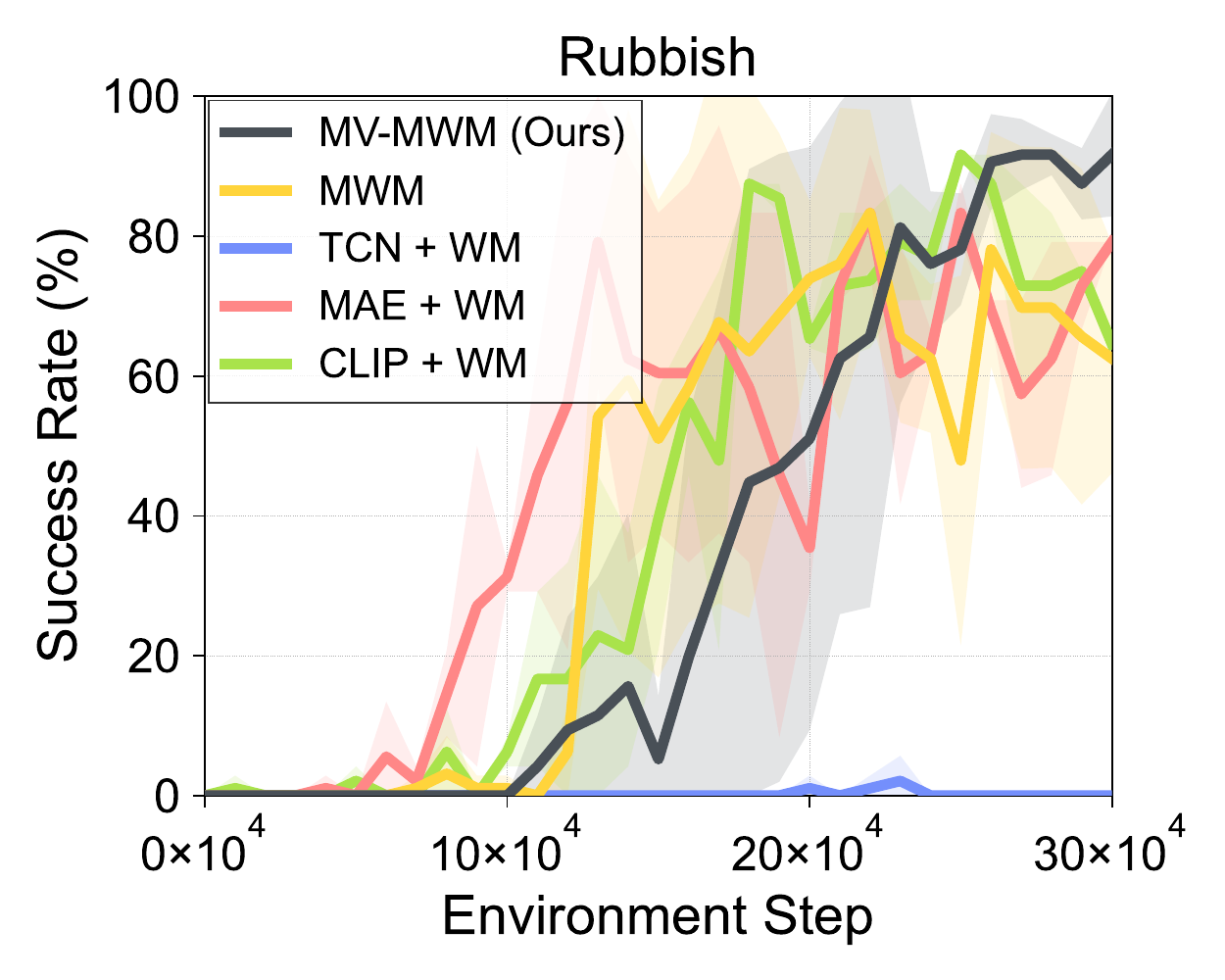}
\label{fig:multiview_rubbish}
}
\subfigure{
\includegraphics[width=0.31\textwidth]{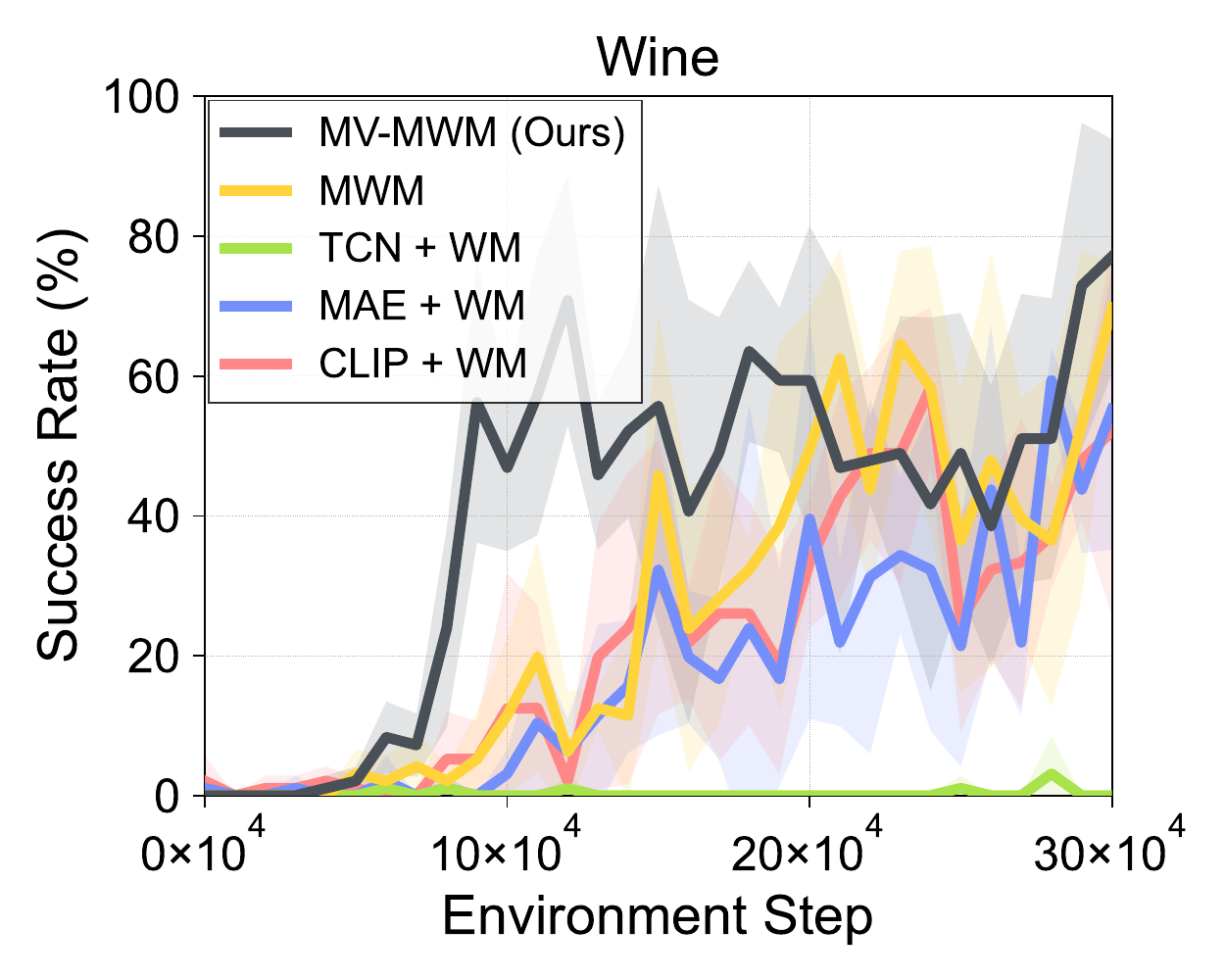}
\label{fig:multiview_wine}
}
\caption{Learning curves of RL agents that operate on front and wrist camera observation for solving five tasks from RLBench as measured on the success rate. The solid line and shaded regions represent the mean and standard deviations, respectively, across 4 runs.}
\end{figure*}

\subsection{Single-view Control with Front Camera}
\vspace{-0.2in}
\begin{figure*}[b!]
\centering
\subfigure{
\includegraphics[width=0.31\textwidth]{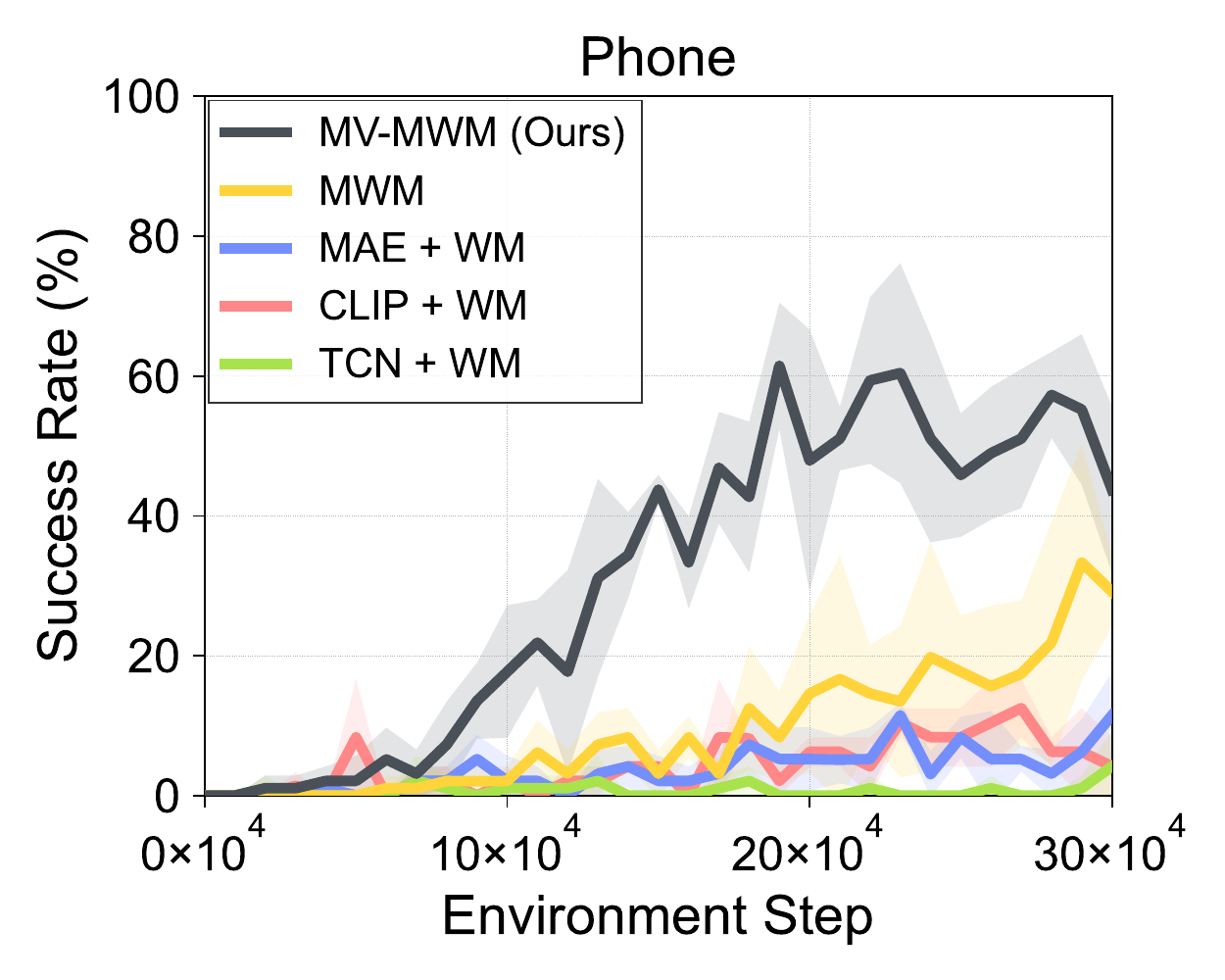}
\label{fig:singleview_phone}
}
\subfigure{
\includegraphics[width=0.31\textwidth]{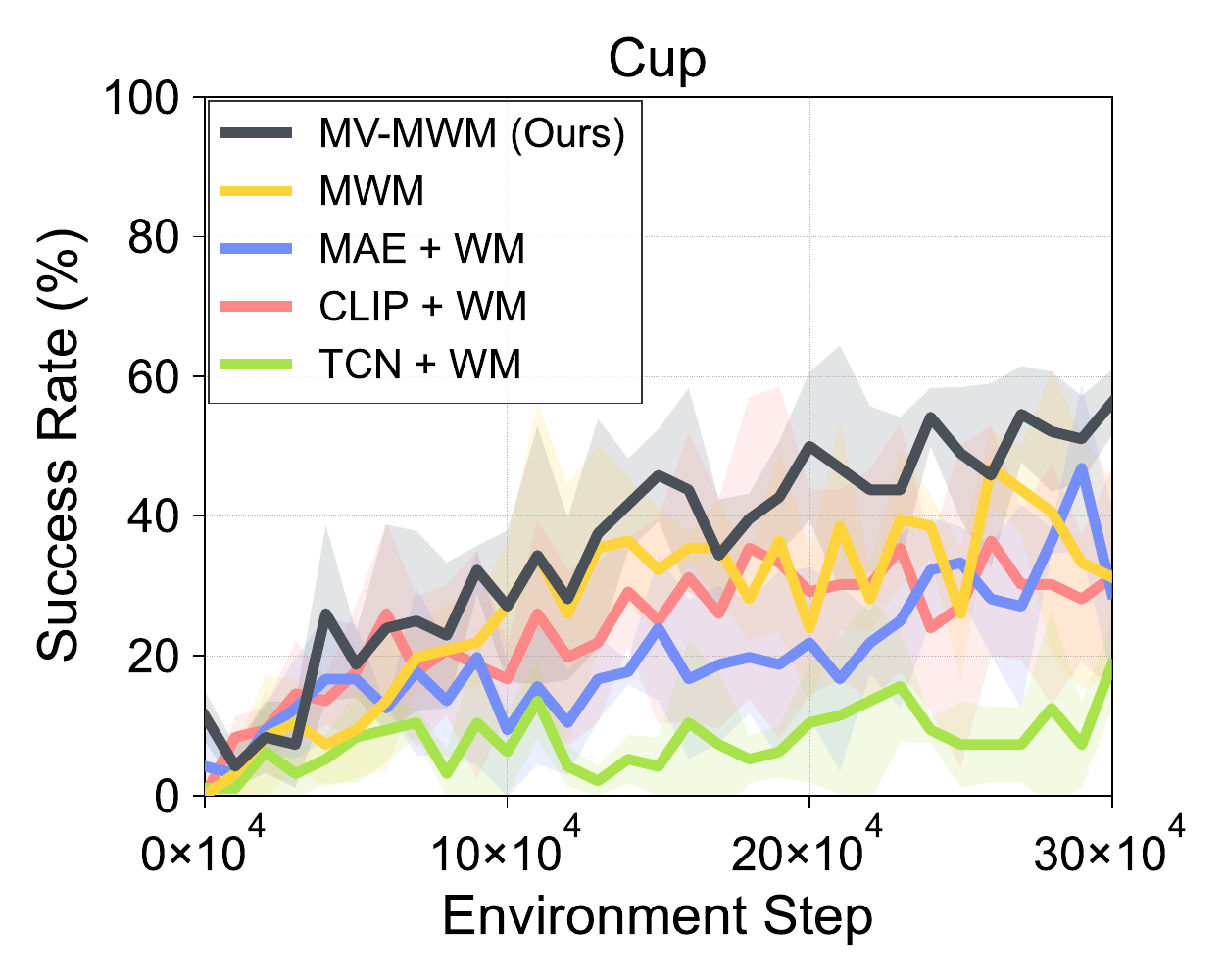}
\label{fig:singleview_cup}
}
\subfigure{
\includegraphics[width=0.31\textwidth]{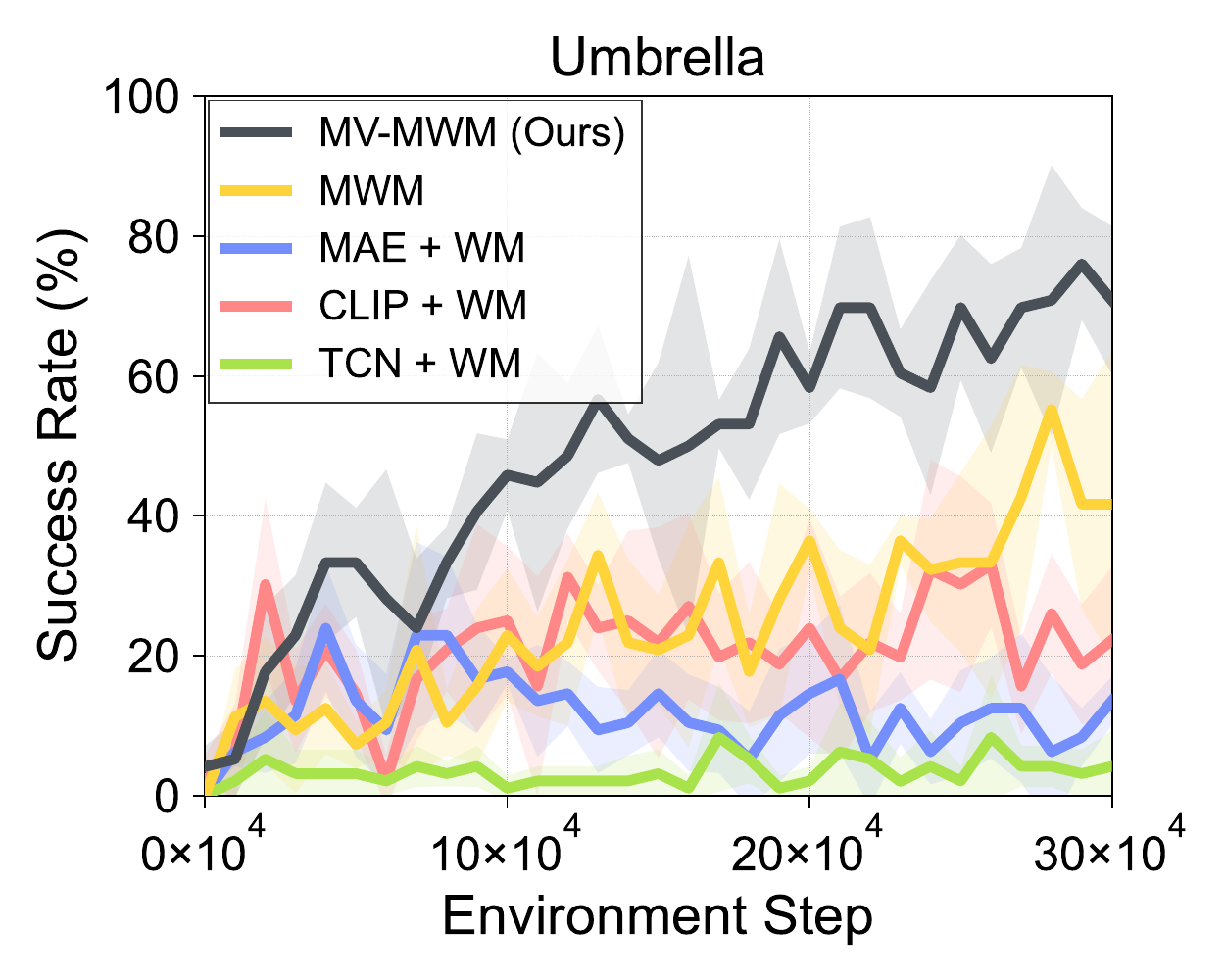}
\label{fig:singleview_umbrella}
}
\subfigure{
\includegraphics[width=0.31\textwidth]{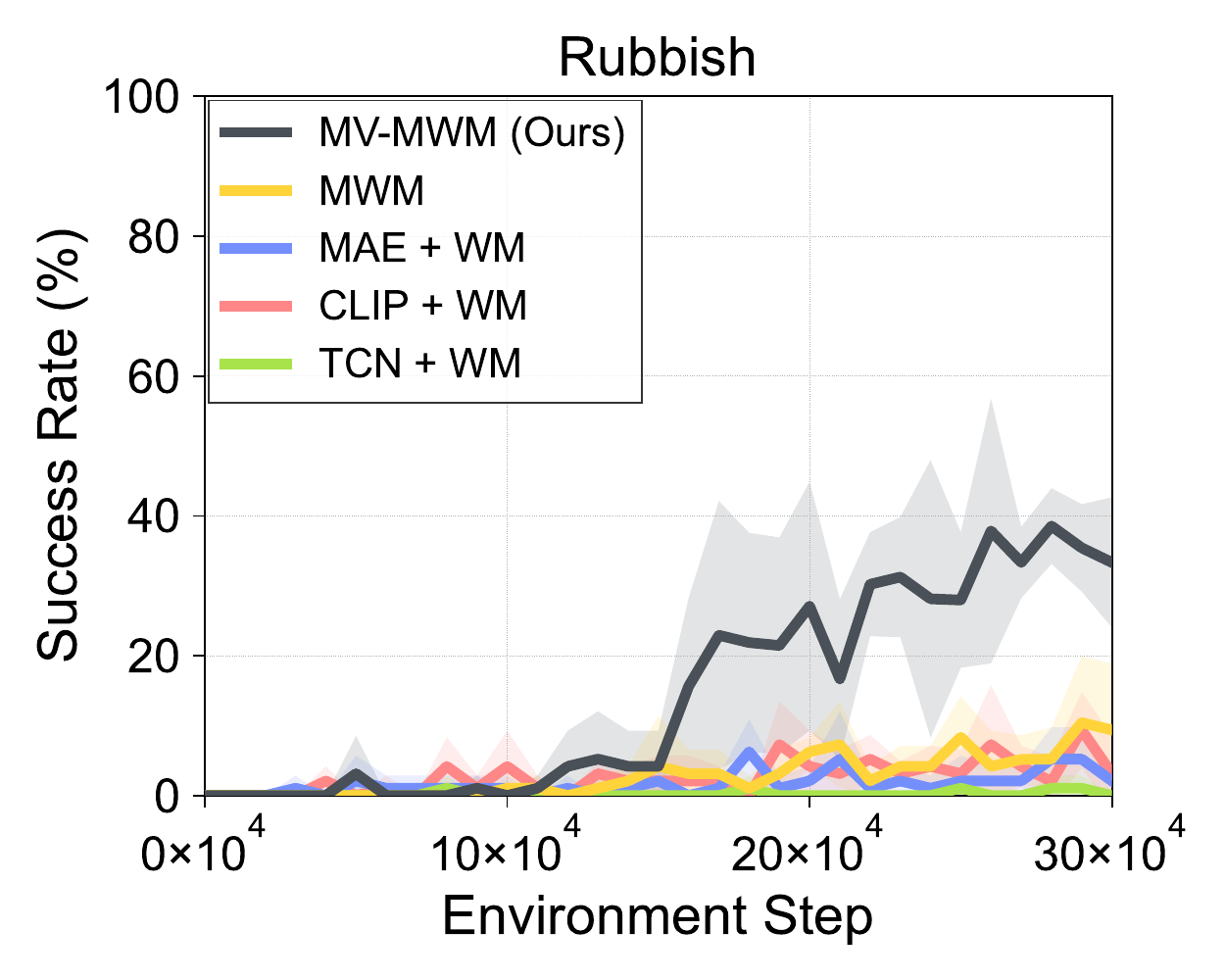}
\label{fig:singleview_rubbish}
}
\subfigure{
\includegraphics[width=0.31\textwidth]{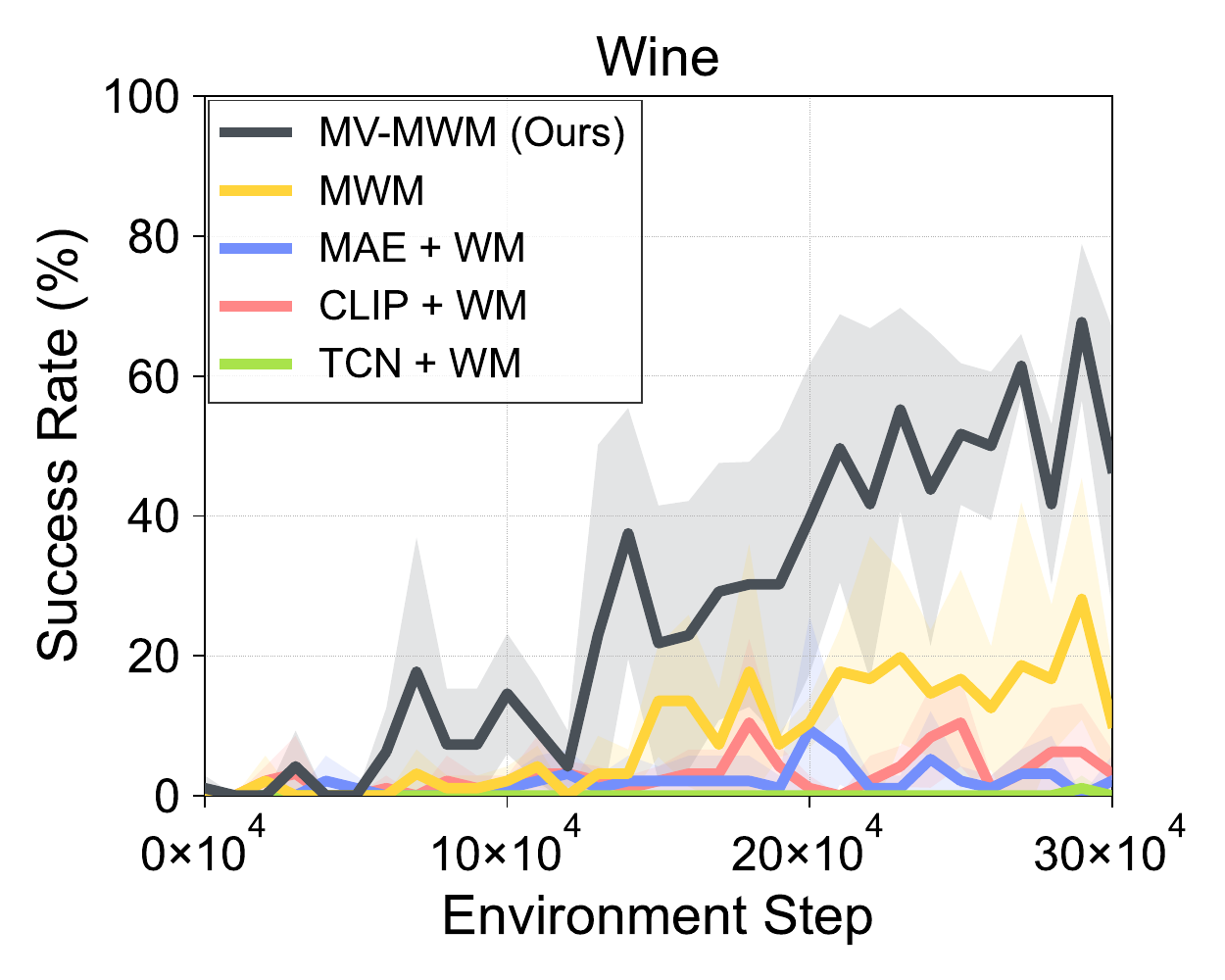}
\label{fig:singleview_wine}
}
\vspace{-.2in}
\caption{Learning curves of RL agents that operate on front camera observation for solving five tasks from RLBench as measured on the success rate. The solid line and shaded regions represent the mean and standard deviations, respectively, across 4 runs.}
\end{figure*}

\clearpage
\subsection{Viewpoint-Robust Control}
\vspace{-.1in}

\begin{figure}[h]
\centering
\subfigure{
\includegraphics[width=0.31\textwidth]{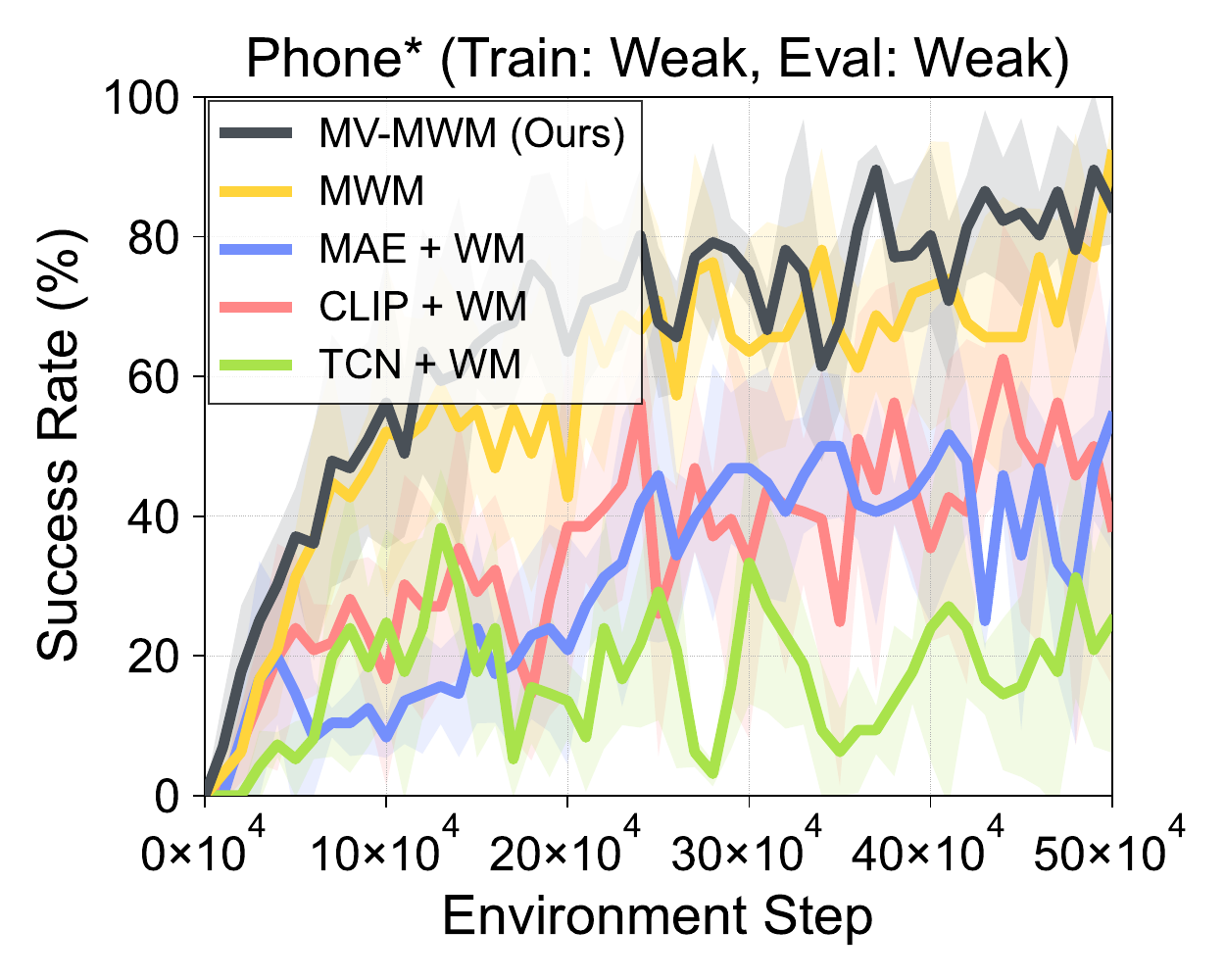}
\label{fig:weak_seen_phone_cano_aug}
}
\subfigure{
\includegraphics[width=0.31\textwidth]{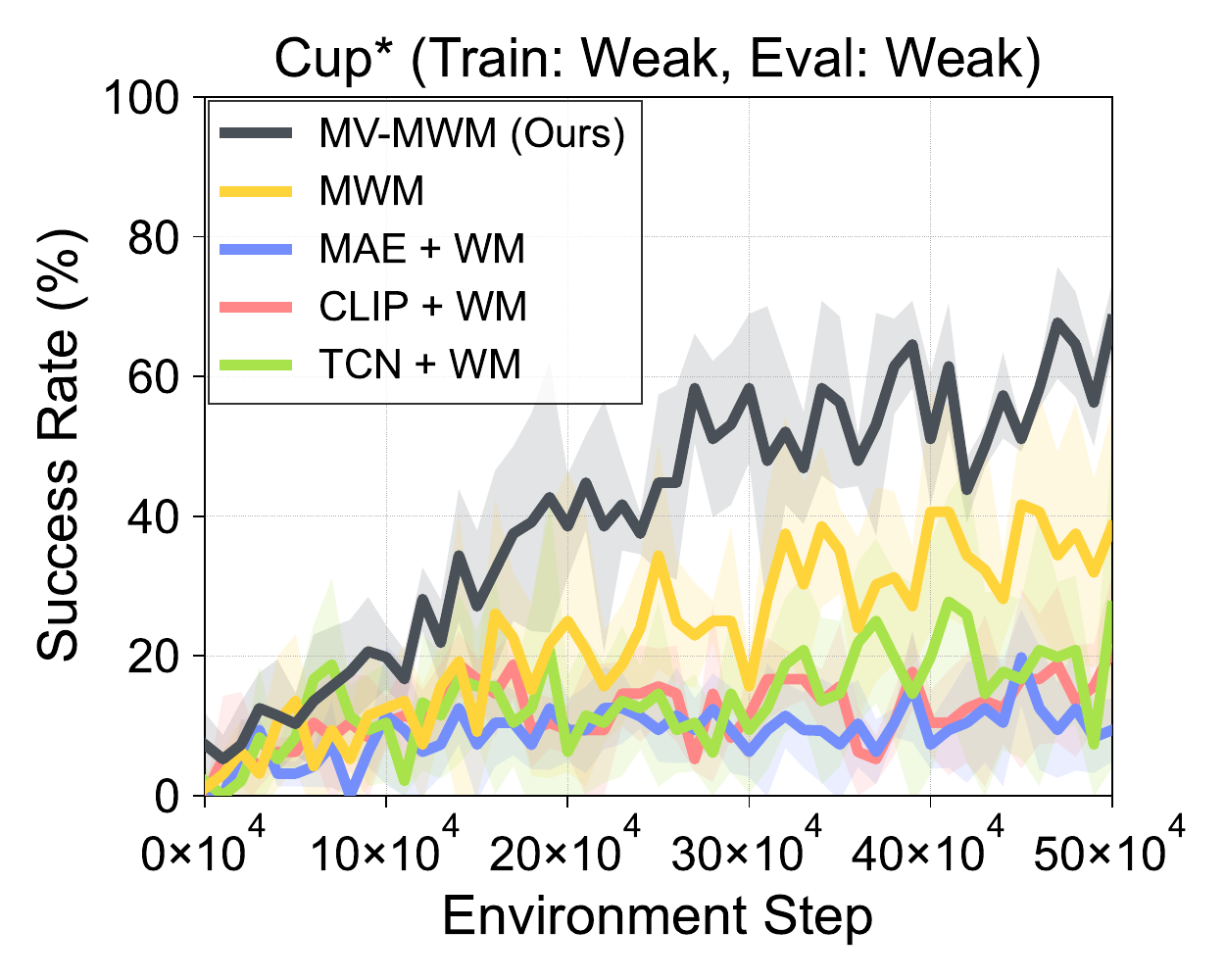}
\label{fig:weak_seen_umbrella_cano_aug}
}\\
\subfigure{
\includegraphics[width=0.31\textwidth]{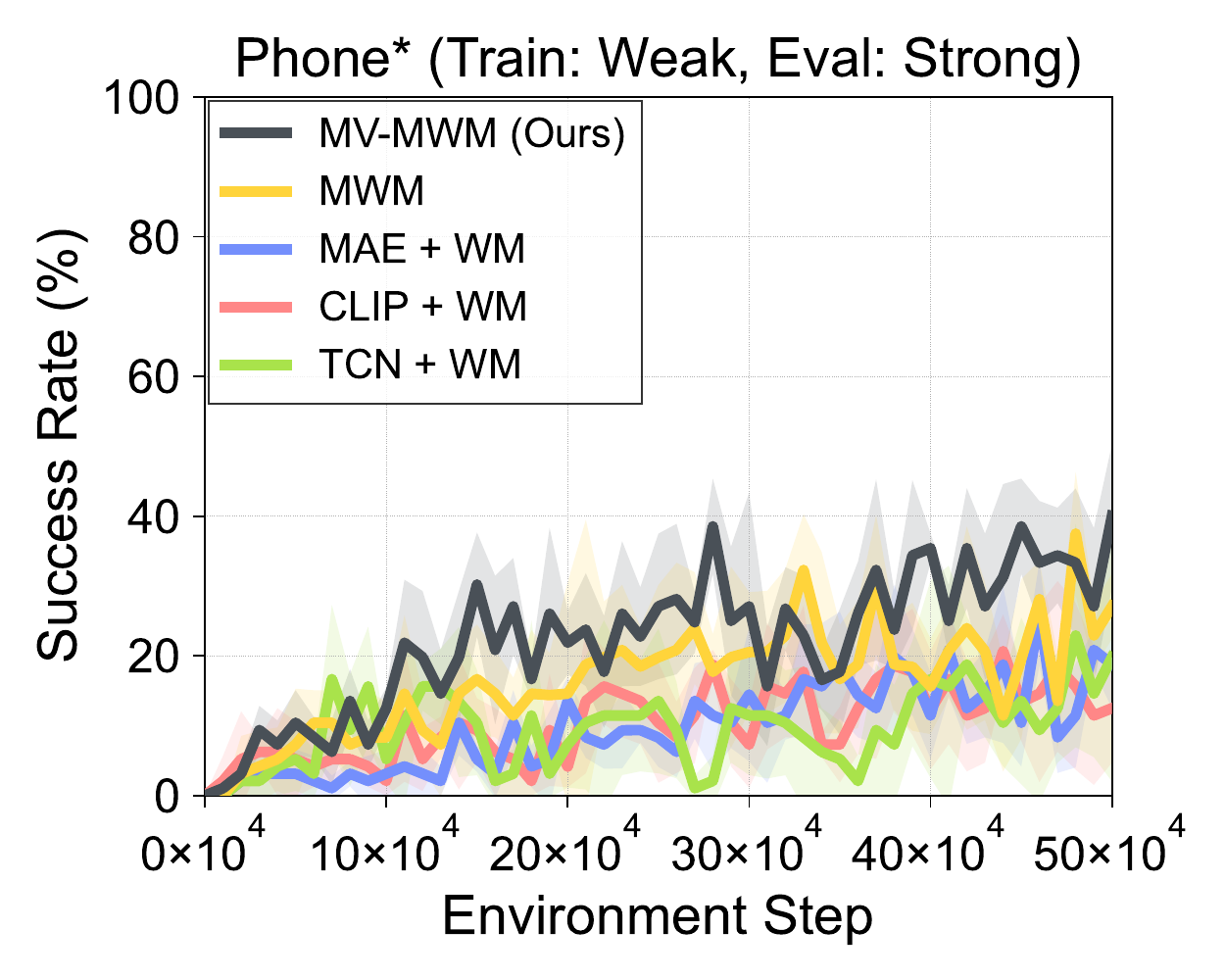}
\label{fig:weak_unseen_phone_cano_aug}
}
\subfigure{
\includegraphics[width=0.31\textwidth]{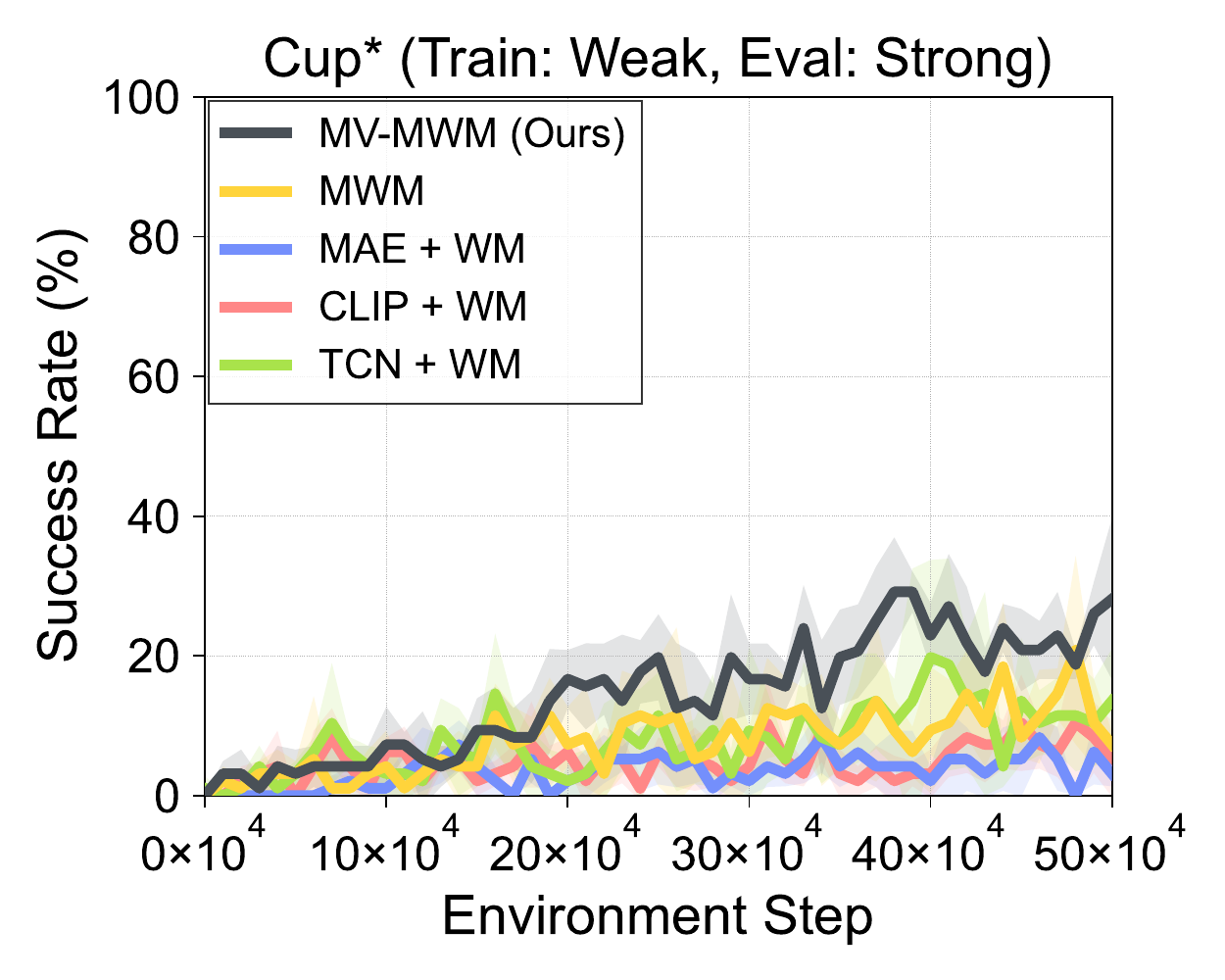}
\label{fig:weak_unseen_umbrella_cano_aug}
}
\caption{Success rate on seen (weak) and unseen (strong) viewpoints by RL agents trained on weak randomization across two tasks from RLBench. The solid line and shaded regions represent the mean and standard deviations, respectively, across 4 runs.}
\end{figure}

\begin{figure*}[b!]
\centering
\subfigure{
\includegraphics[width=0.31\textwidth]{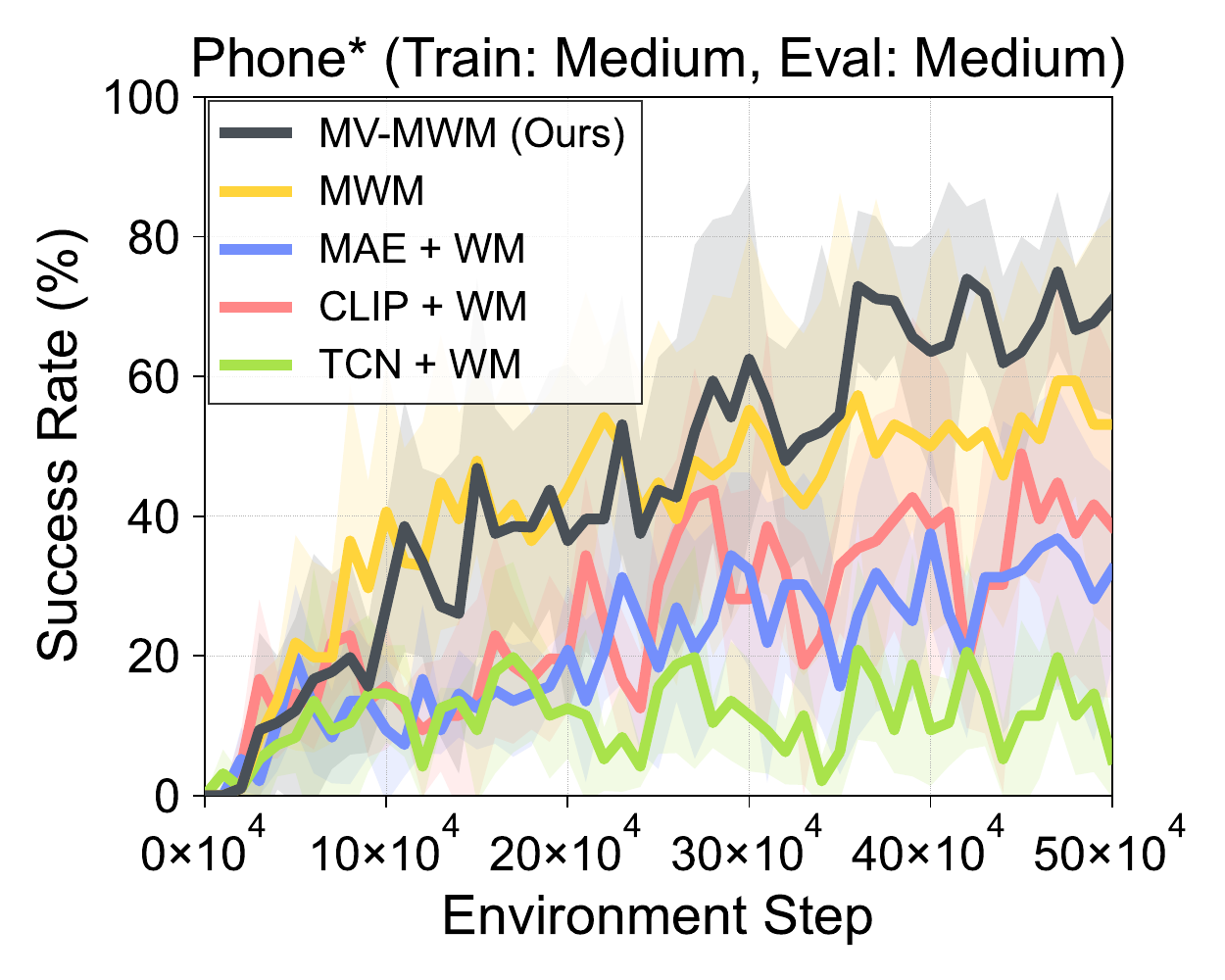}
\label{fig:strong_seen_phone_cano_aug}
}
\subfigure{
\includegraphics[width=0.31\textwidth]{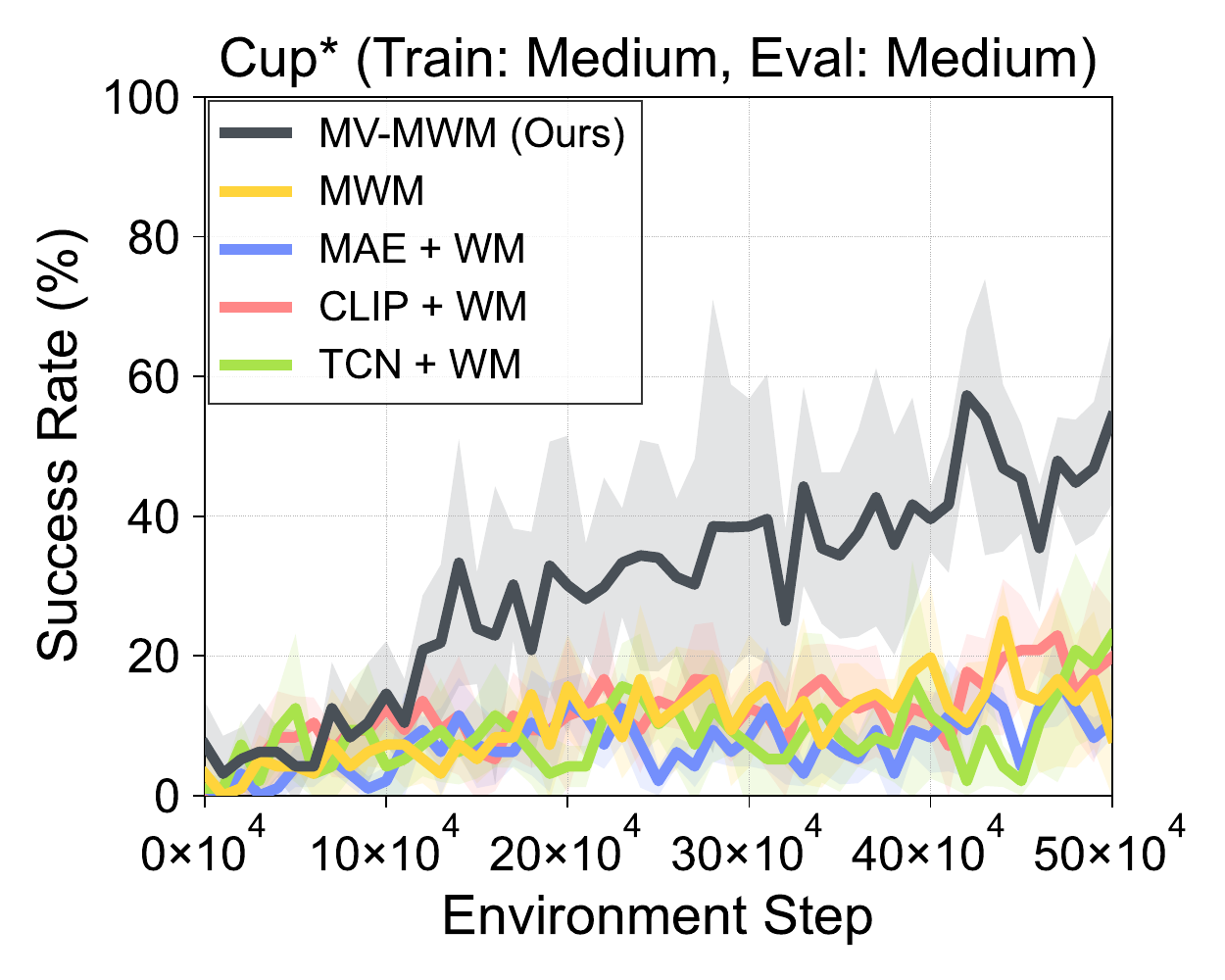}
\label{fig:strong_seen_umbrella_cano_aug}
}\\
\subfigure{
\includegraphics[width=0.31\textwidth]{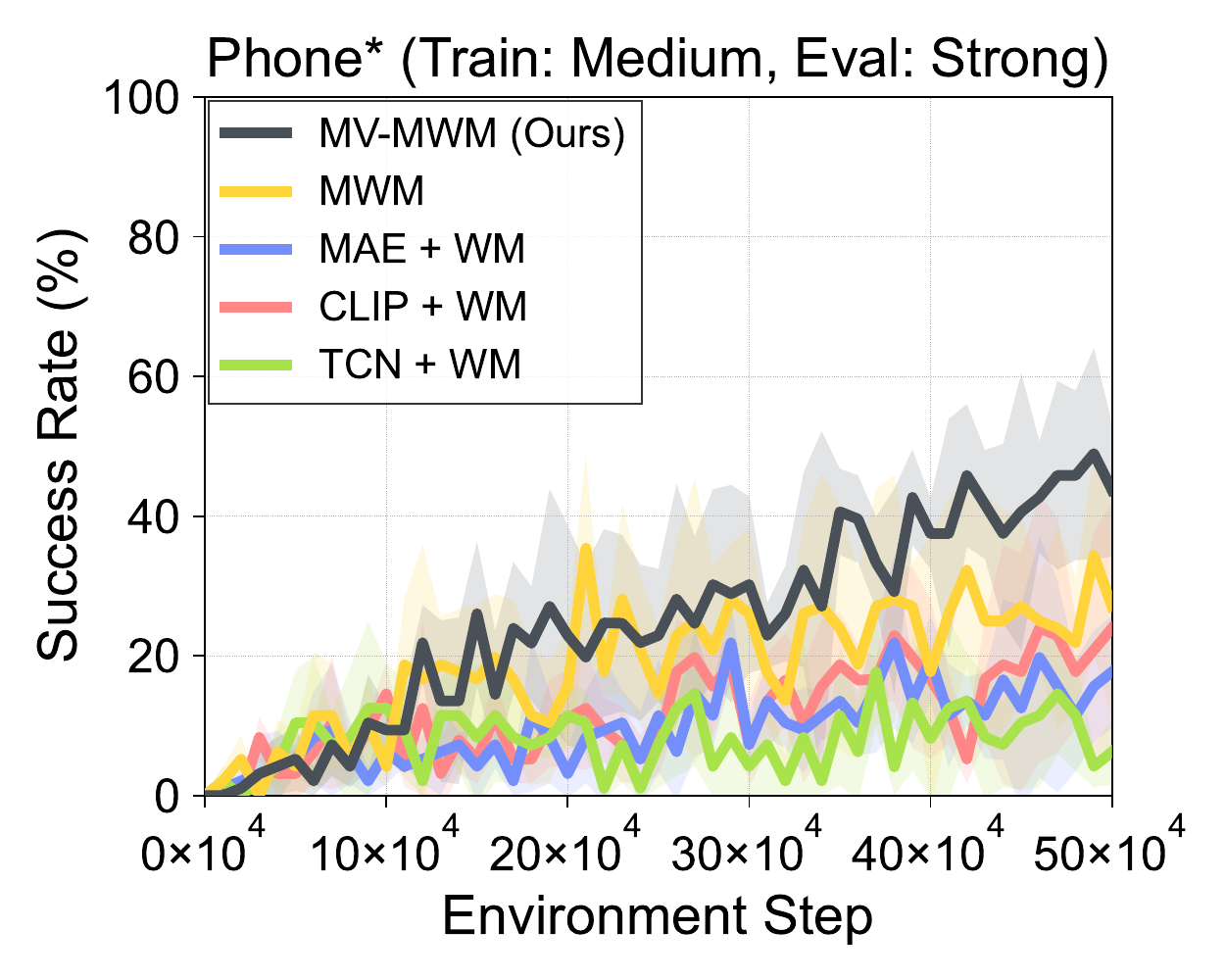}
\label{fig:strong_unseen_phone_cano_aug}
}
\subfigure{
\includegraphics[width=0.31\textwidth]{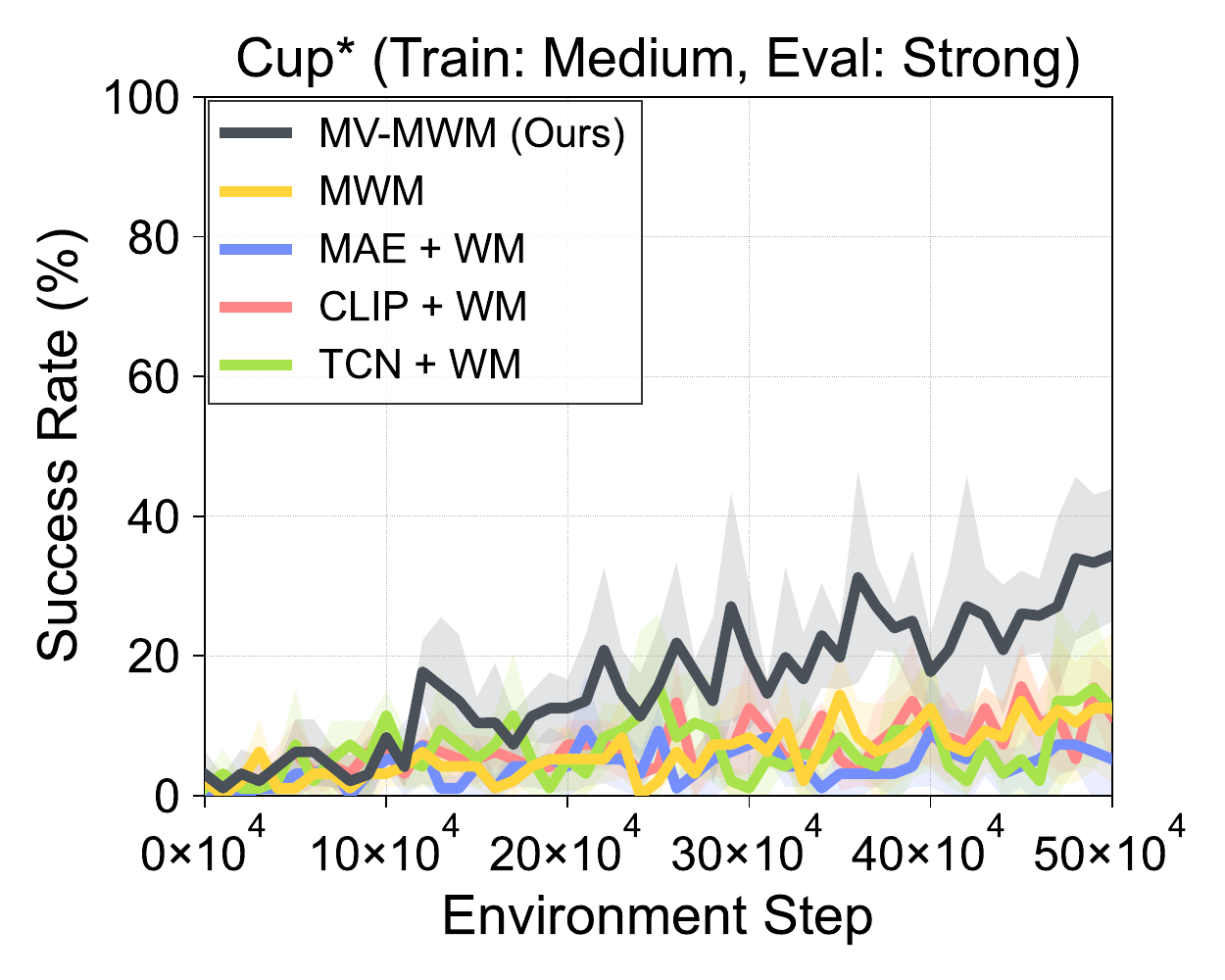}
\label{fig:strong_unseen_umbrella_cano_aug}
}
\caption{Success rate on seen (medium) and unseen (strong) viewpoints by agents trained on medium randomization across two tasks from RLBench. The solid line and shaded regions represent the mean and standard deviations, respectively, across 4 runs.}
\end{figure*}

\clearpage

\subsection{Ablation Study and Analysis}

\begin{figure*}[h]
\centering
\subfigure{
\includegraphics[width=0.31\textwidth]{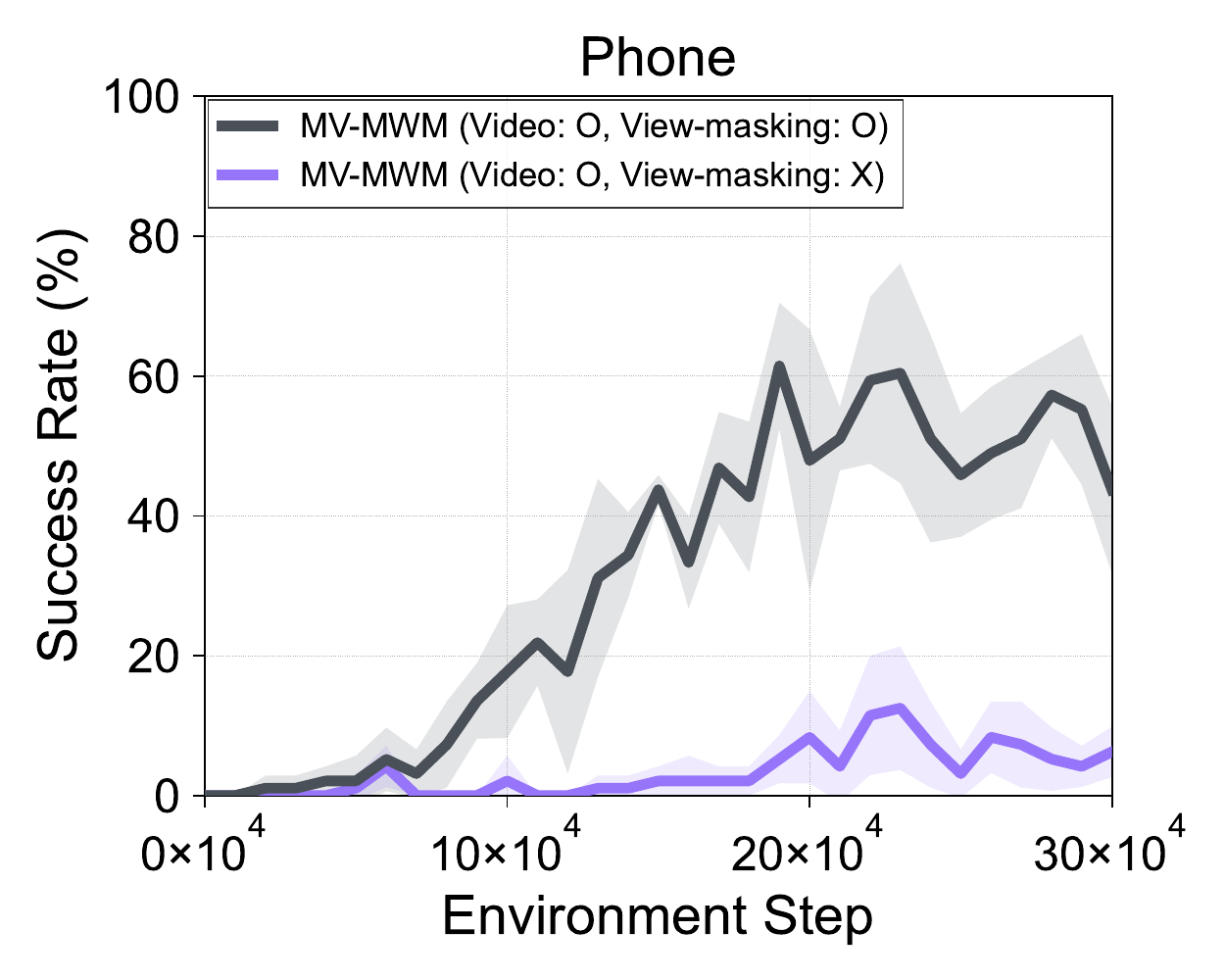}
\label{fig:vm_phone}
}
\subfigure{
\includegraphics[width=0.31\textwidth]{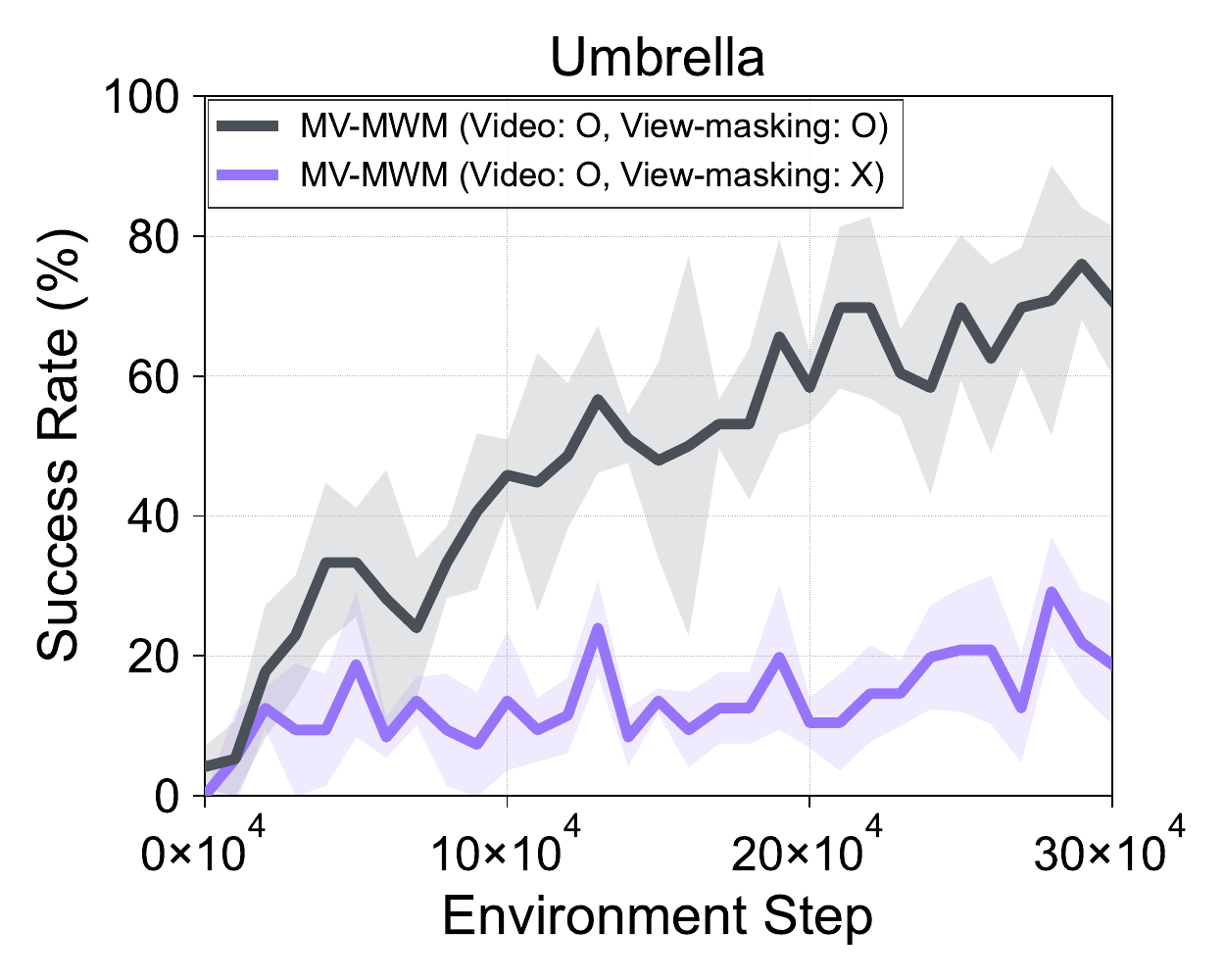}
\label{fig:vm_umbrella}
}
\subfigure{
\includegraphics[width=0.31\textwidth]{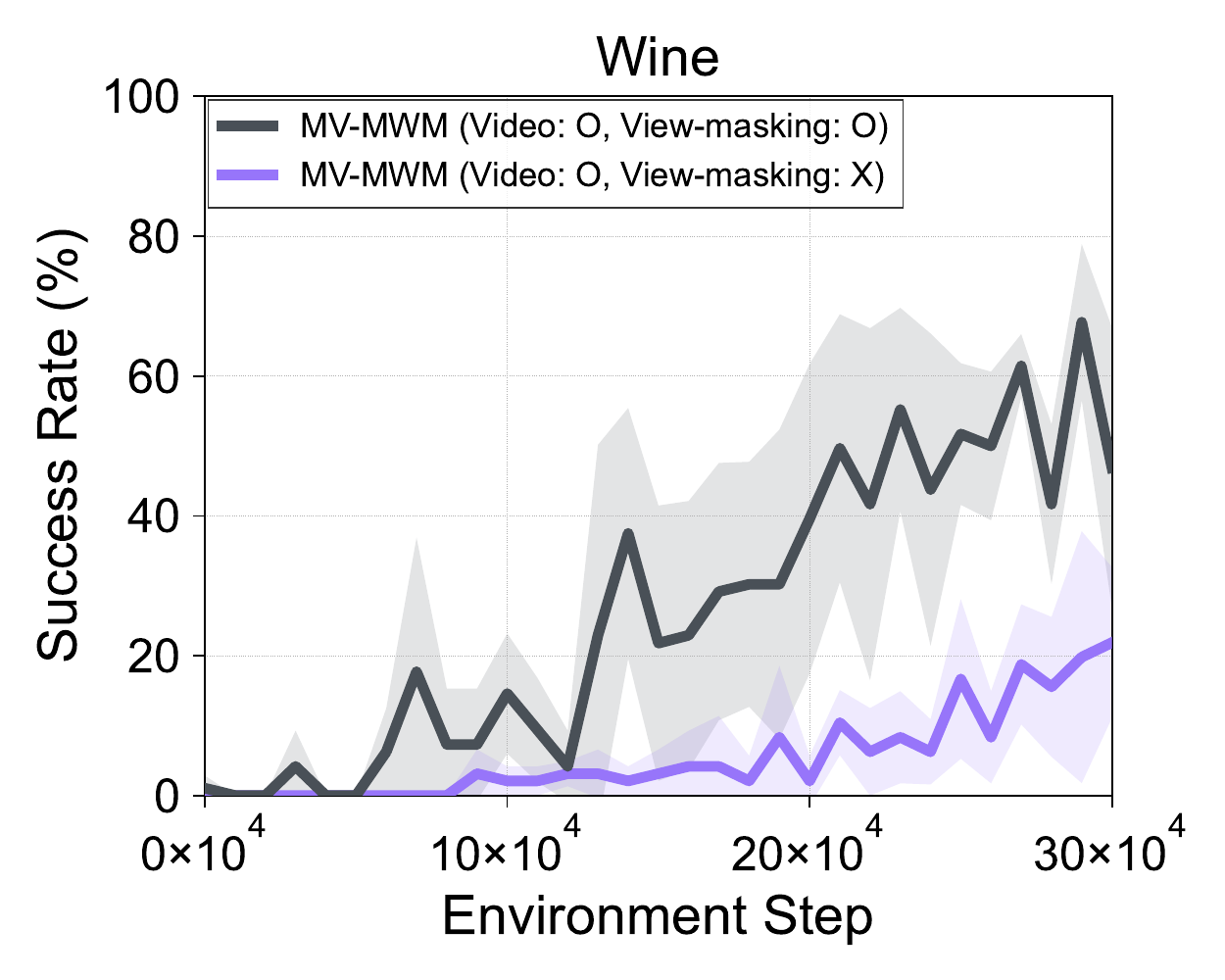}
\label{fig:vm_wine}
}
\caption{Effect of view masking. The solid line and shaded regions represent the mean and standard deviation across 4 runs.}
\end{figure*}

\begin{figure*}[h]
\centering
\subfigure{
\includegraphics[width=0.31\textwidth]{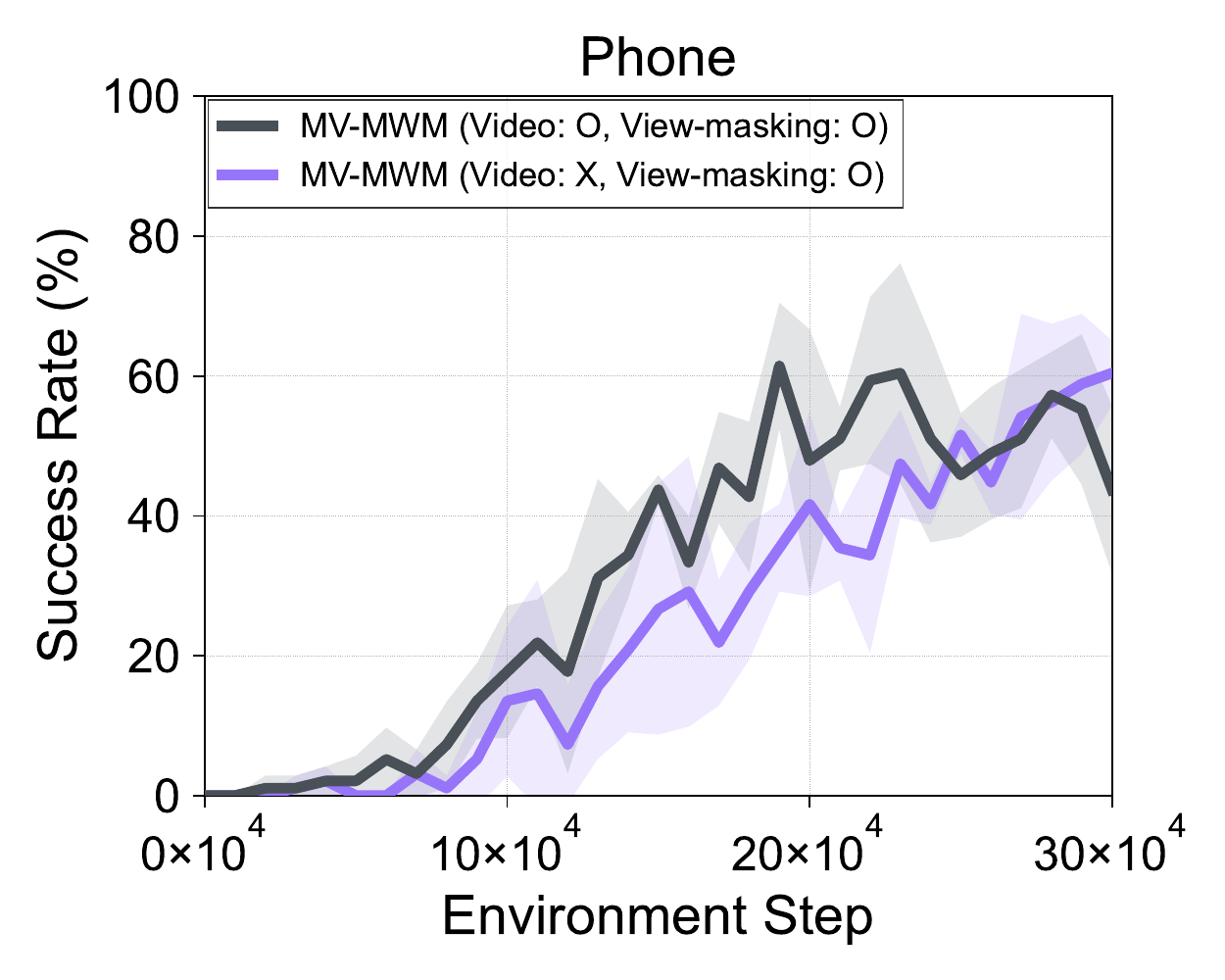}
\label{fig:vae_phone}
}
\subfigure{
\includegraphics[width=0.31\textwidth]{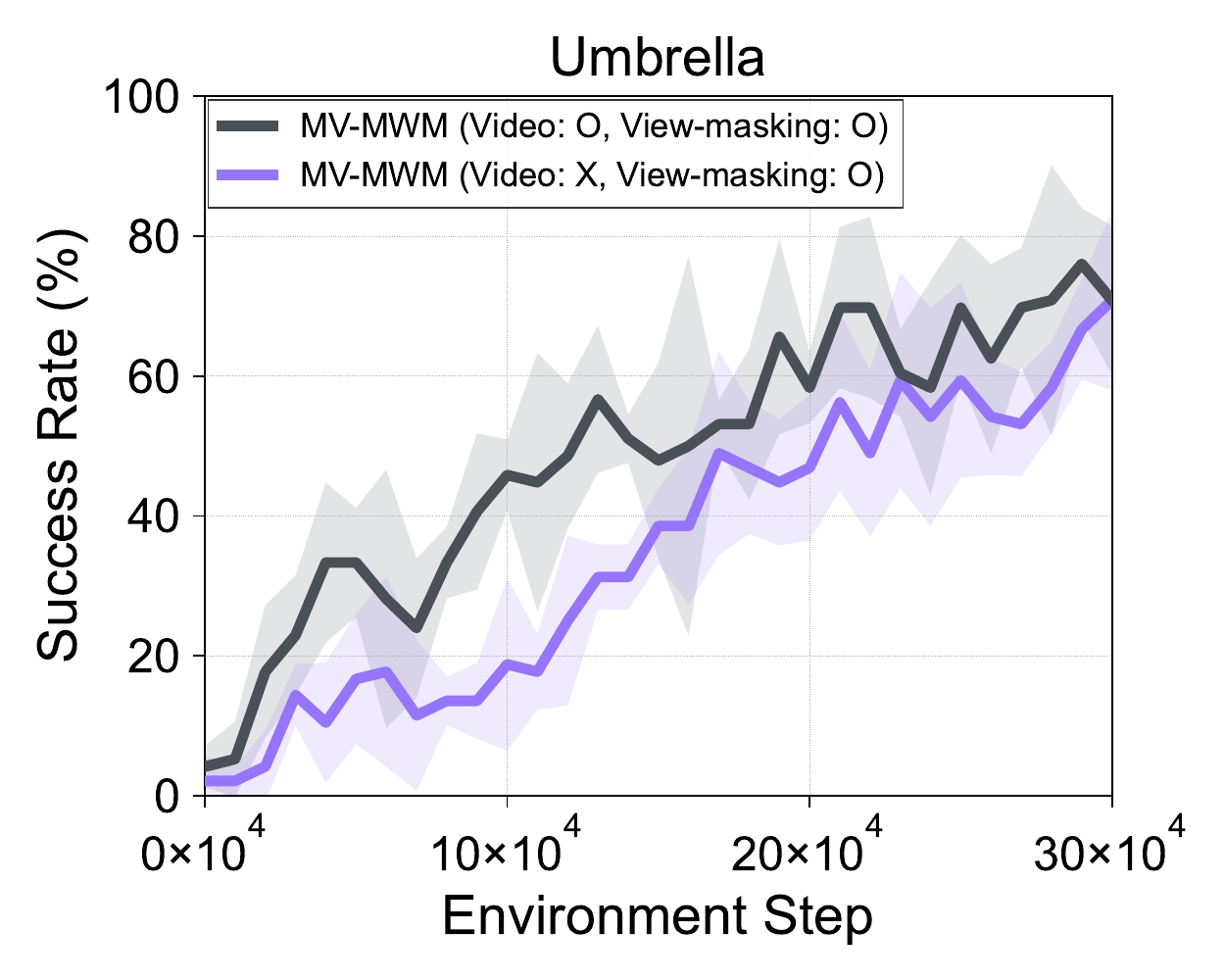}
\label{fig:vae_umbrella}
}
\subfigure{
\includegraphics[width=0.31\textwidth]{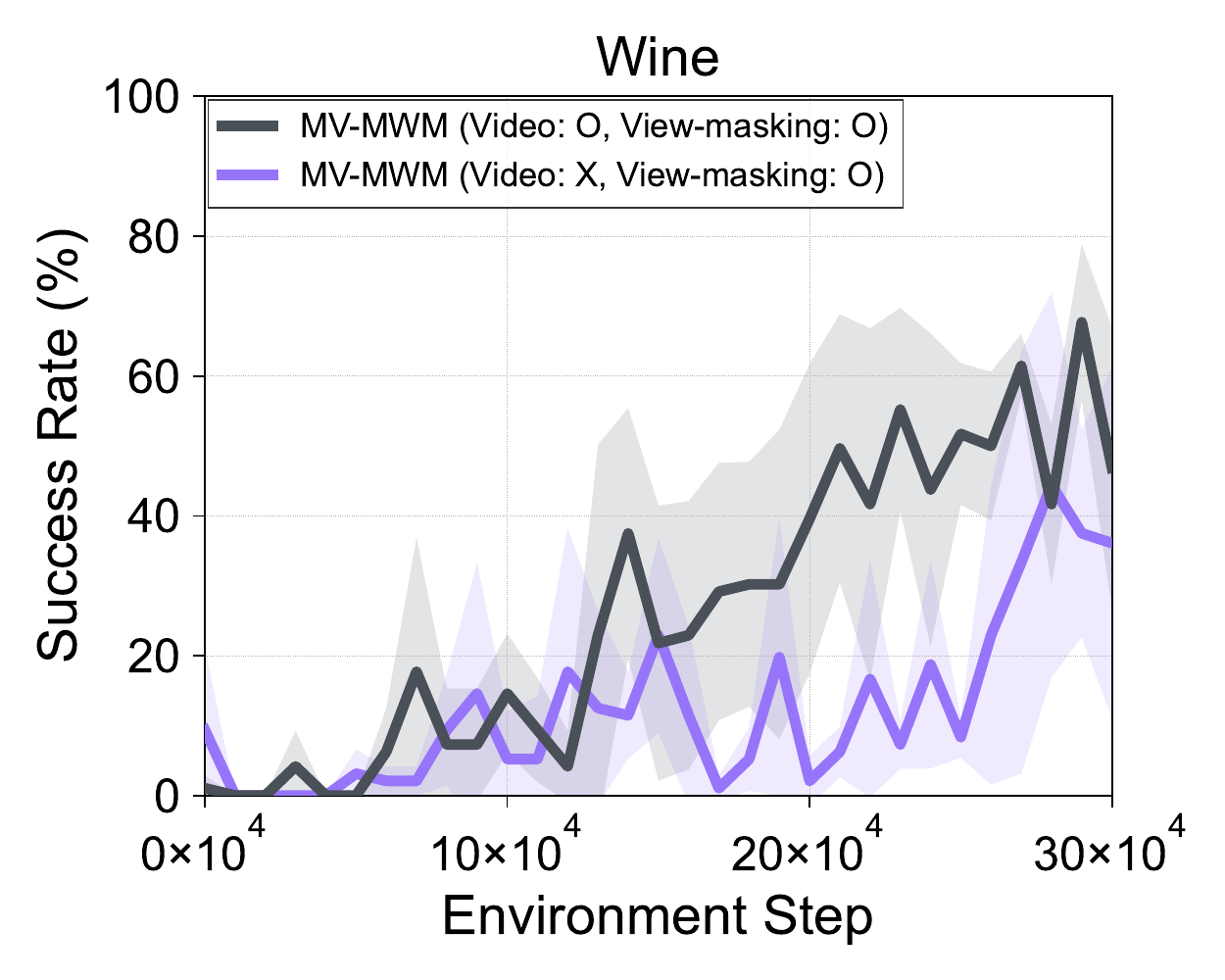}
\label{fig:vae_wine}
}
\caption{Effect of video autoencoding. The solid line and shaded regions represent the mean and standard deviation across 4 runs.}
\end{figure*}

\begin{figure*}[h]
\centering
\subfigure{
\includegraphics[width=0.31\textwidth]{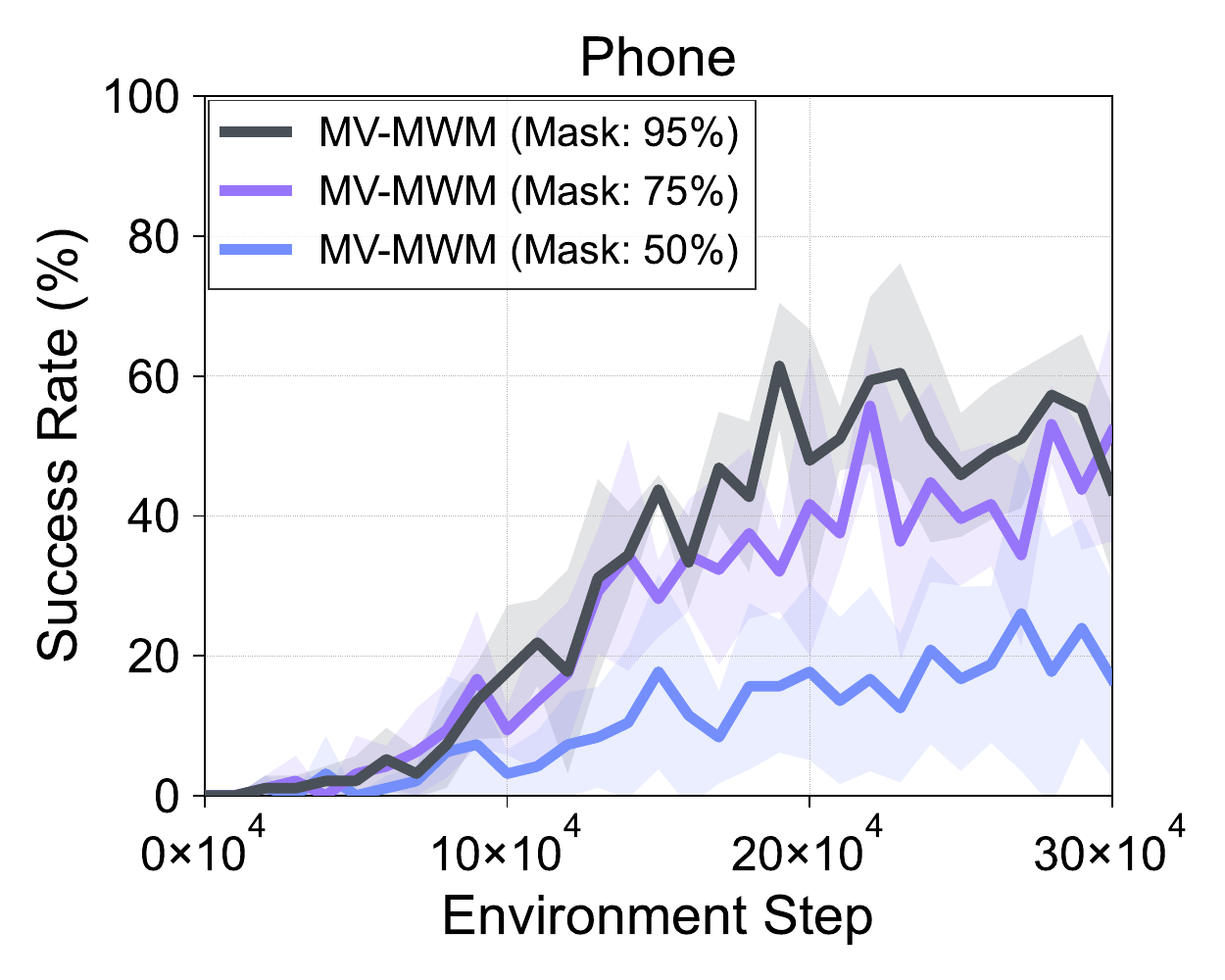}
\label{fig:mr_phone}
}
\subfigure{
\includegraphics[width=0.31\textwidth]{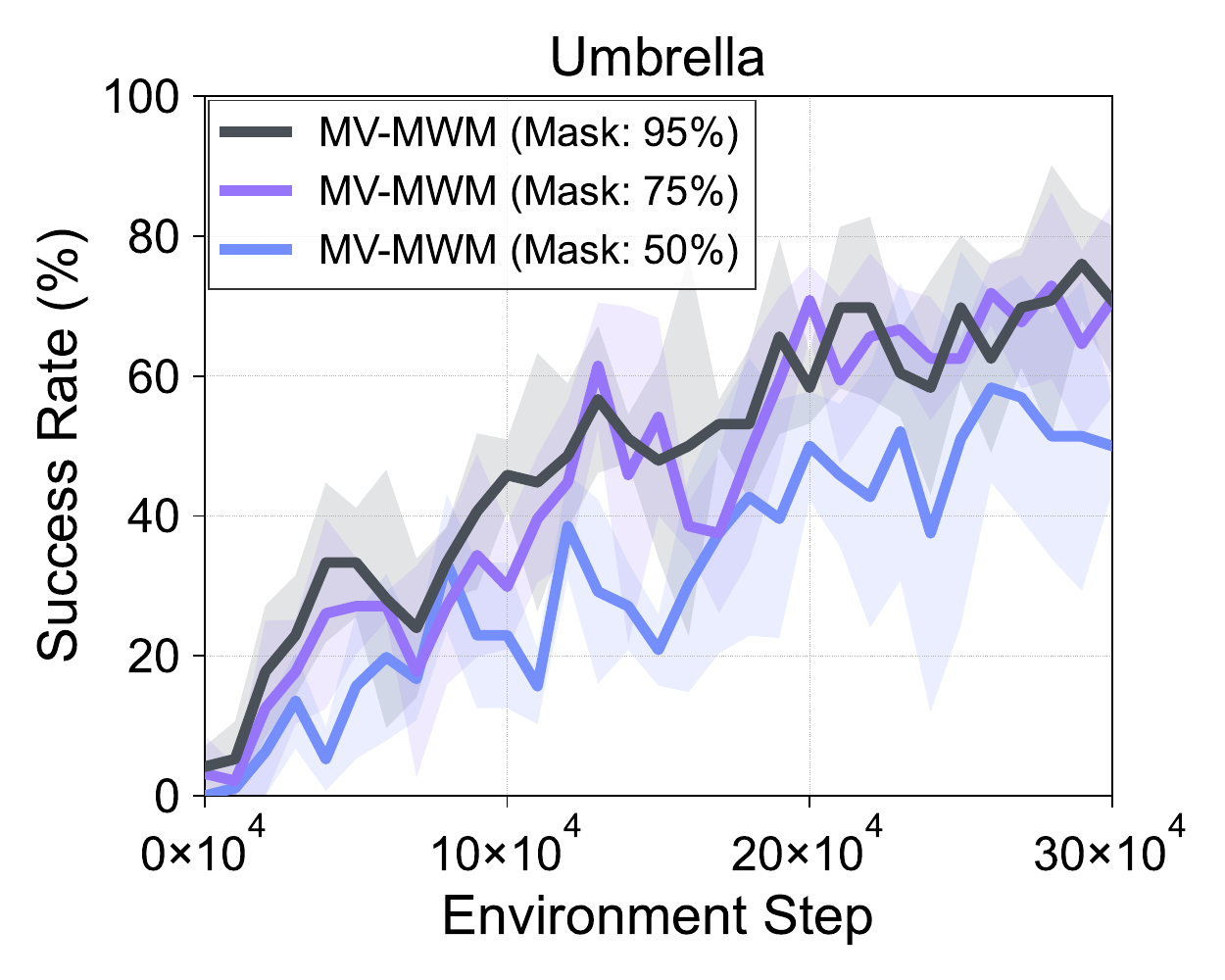}
\label{fig:mr_umbrella}
}
\subfigure{
\includegraphics[width=0.31\textwidth]{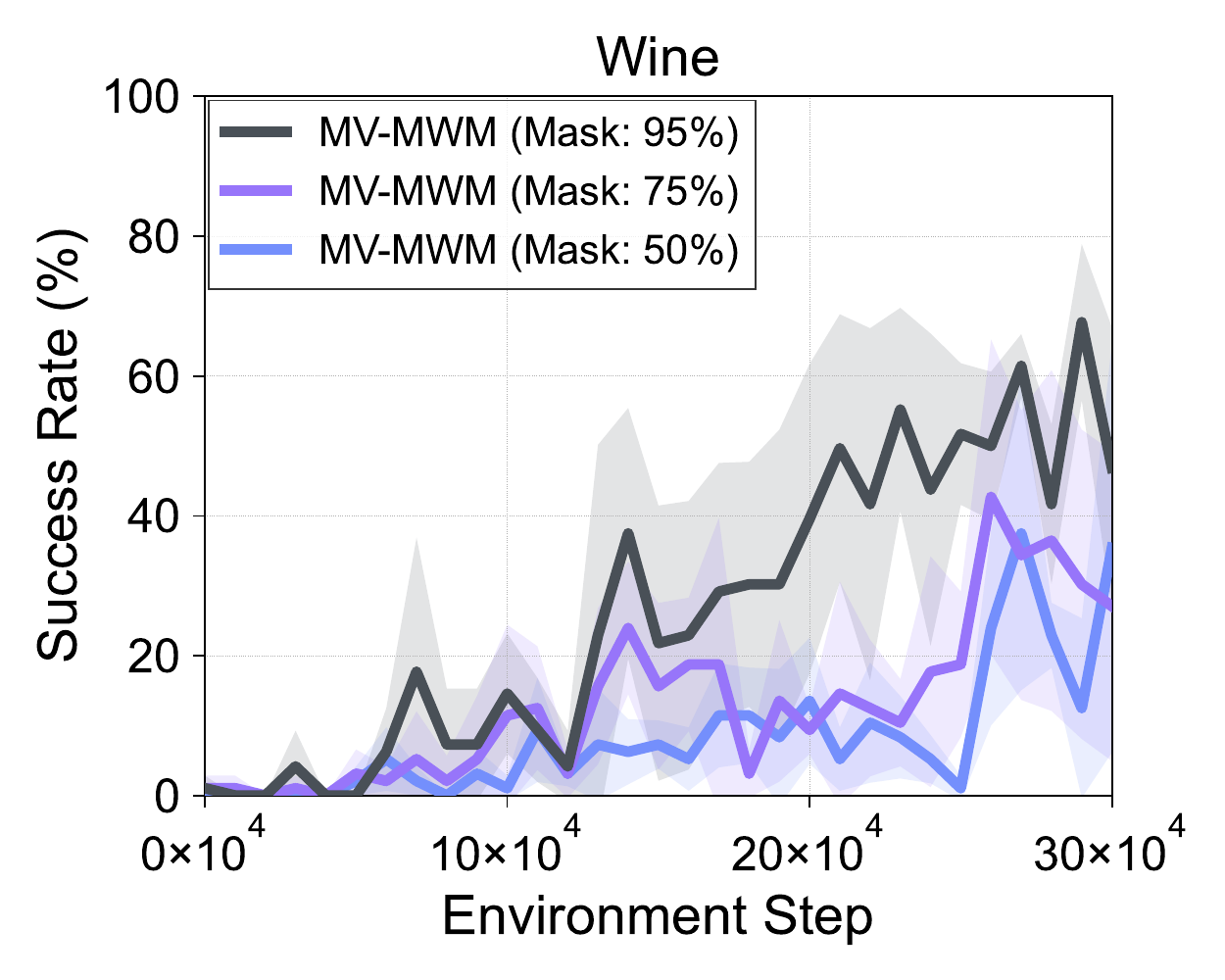}
\label{fig:mr_wine}
}
\caption{Effect of mask ratio. The solid line and shaded regions represent the mean and standard deviation across 4 runs.}
\end{figure*}

\clearpage
\section{Additional Experiments}
\label{appendix:additional_expriments}
\begin{figure*}[h]
\centering
\subfigure[Training ratio]{
\includegraphics[width=0.31\textwidth]{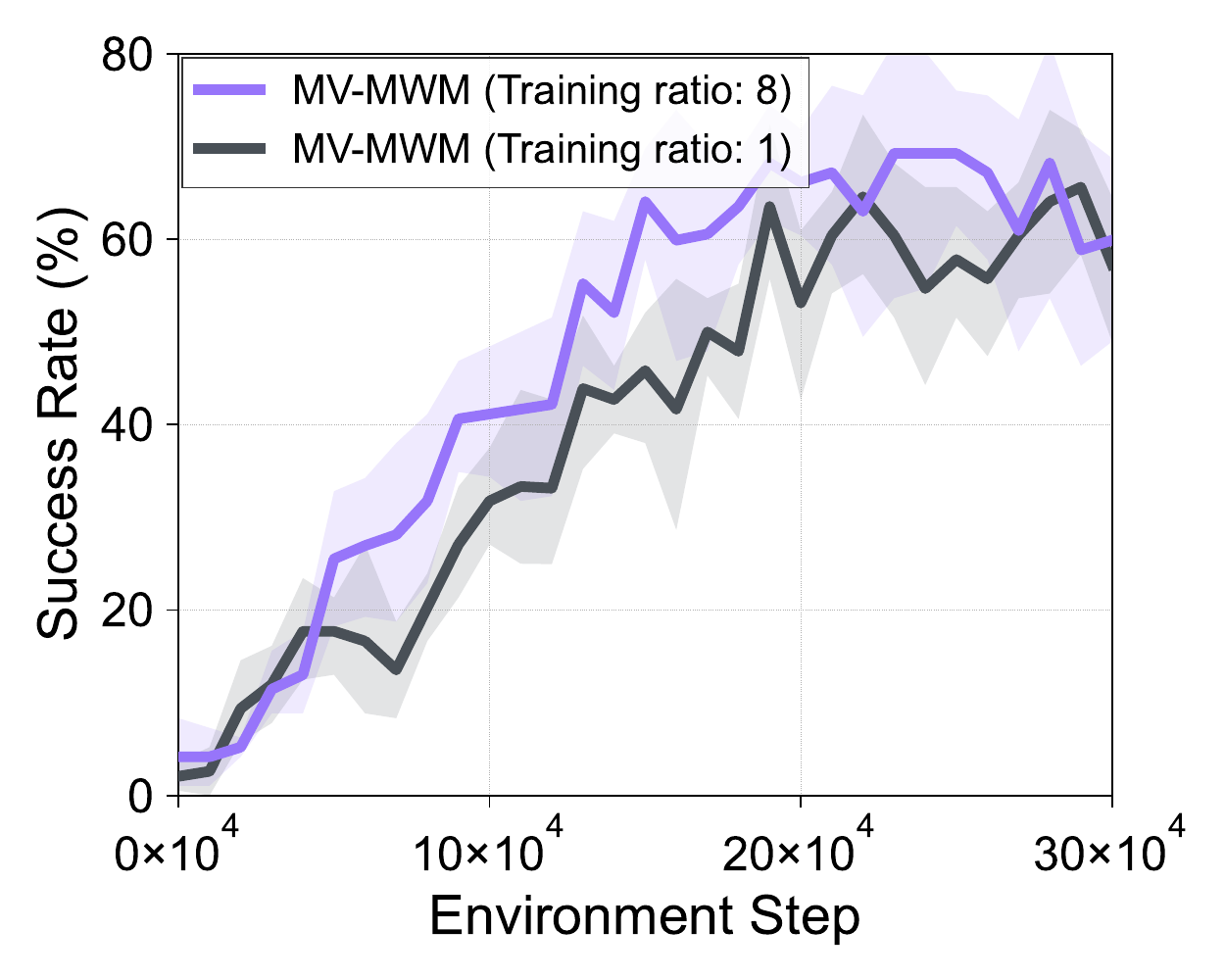}
\label{fig:analysis_training_ratio}
}
\subfigure[Model size]{
\includegraphics[width=0.31\textwidth]{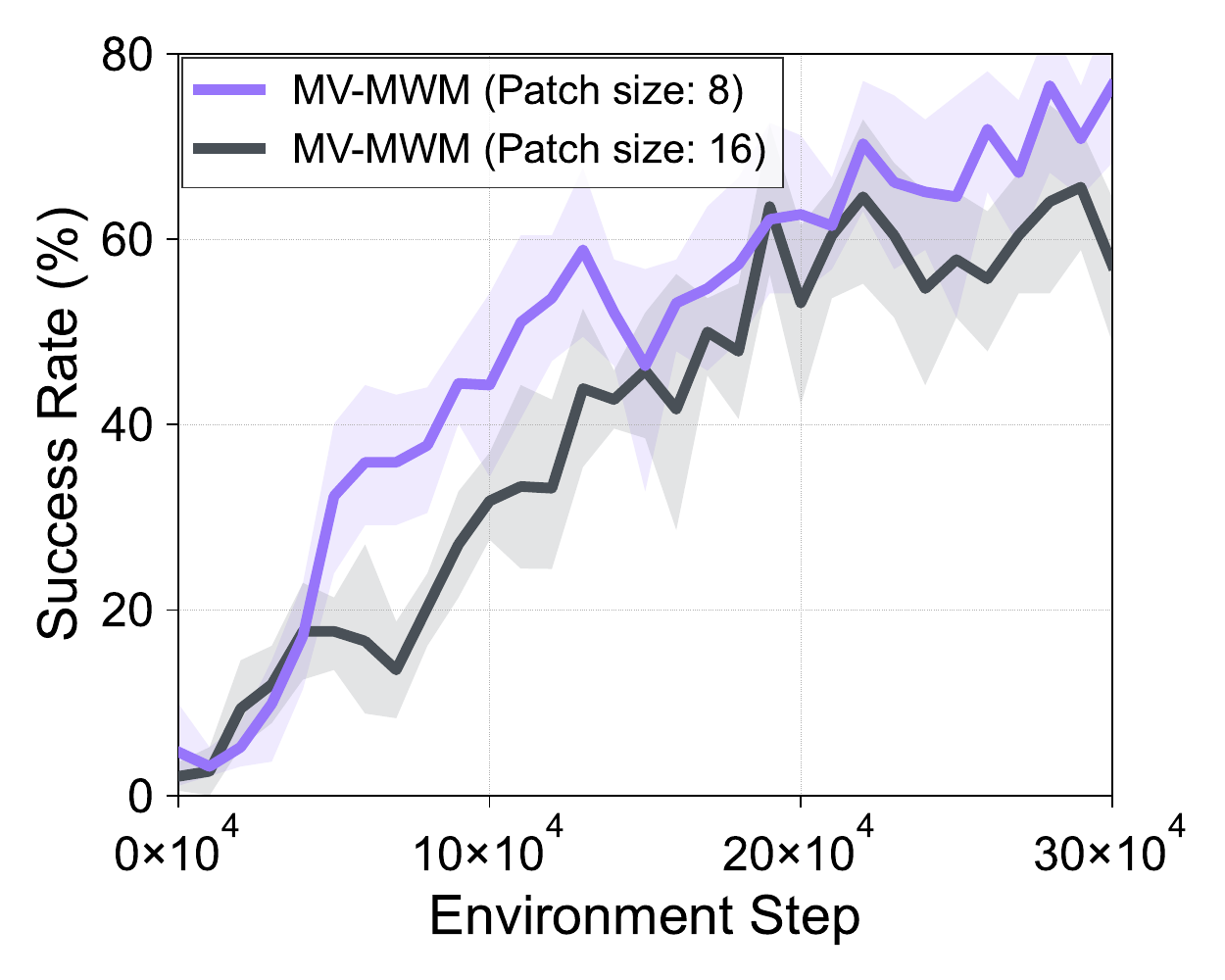}
\label{fig:analysis_model_size}
}
\subfigure[Proprioceptive states]{
\includegraphics[width=0.31\textwidth]{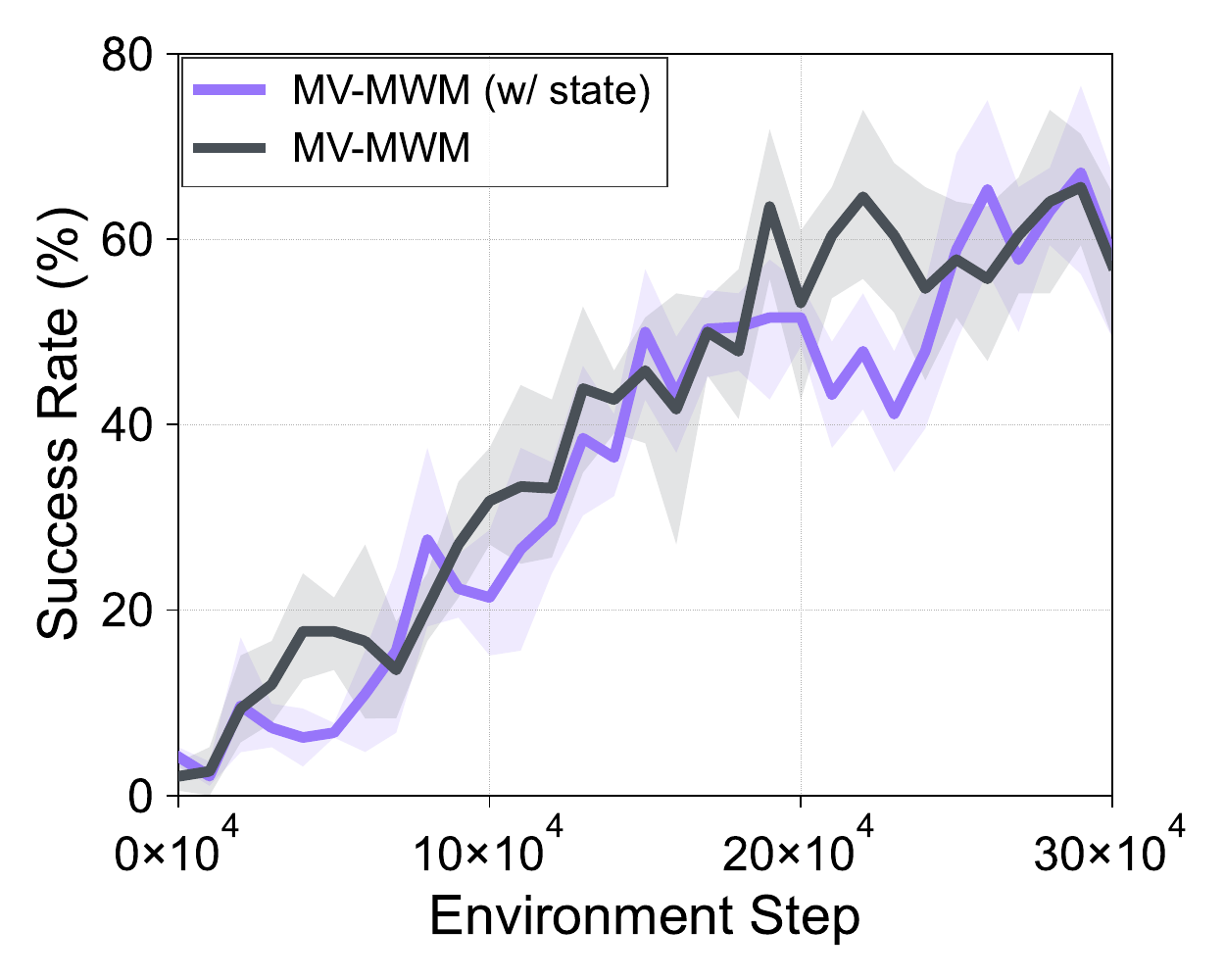}
\label{fig:analysis_state}
}
\\
\subfigure[Finetuning MV-MAE encoder]{
\includegraphics[width=0.31\textwidth]{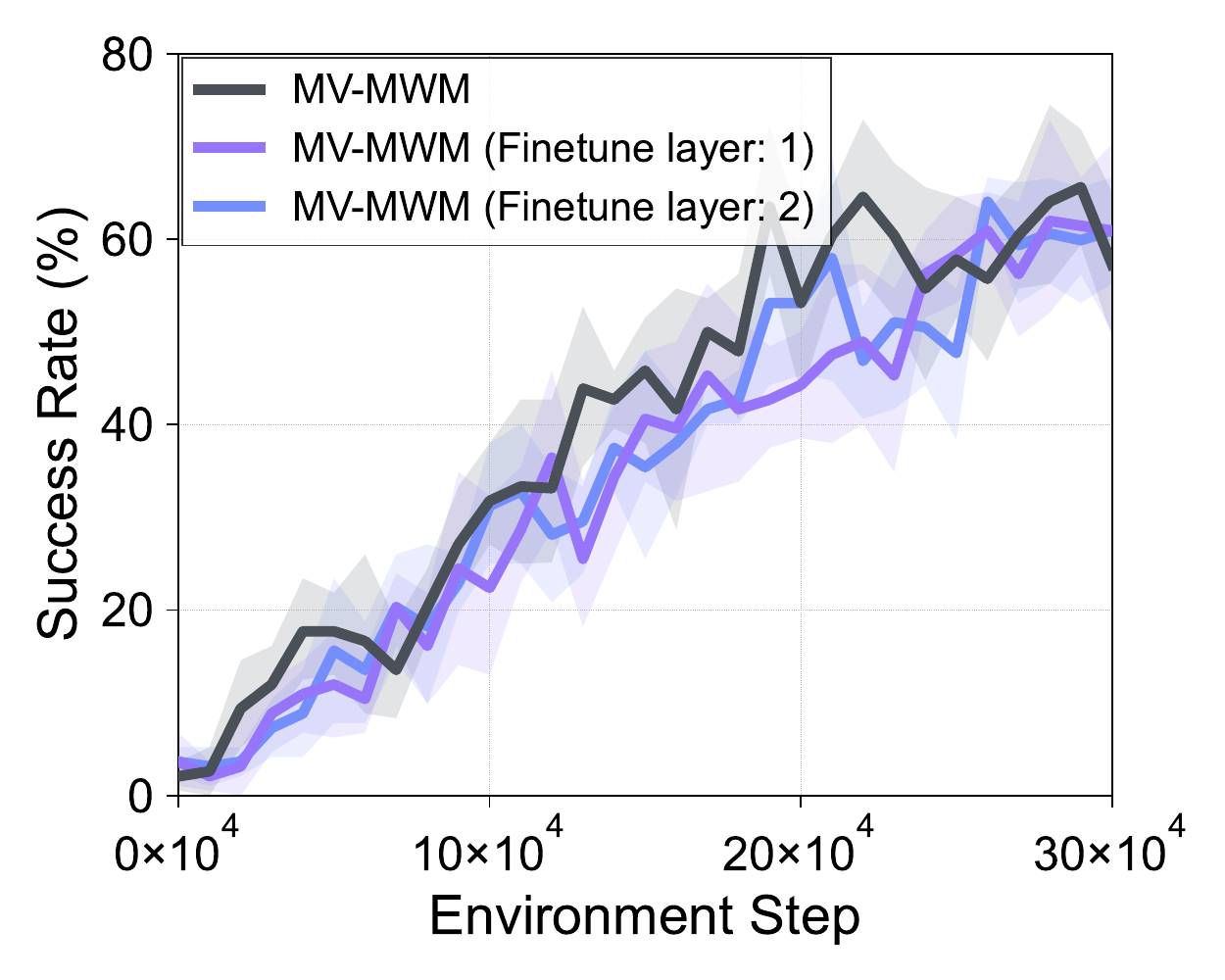}
\label{fig:analysis_finetune}
}
\subfigure[Behavior cloning loss]{
\includegraphics[width=0.31\textwidth]{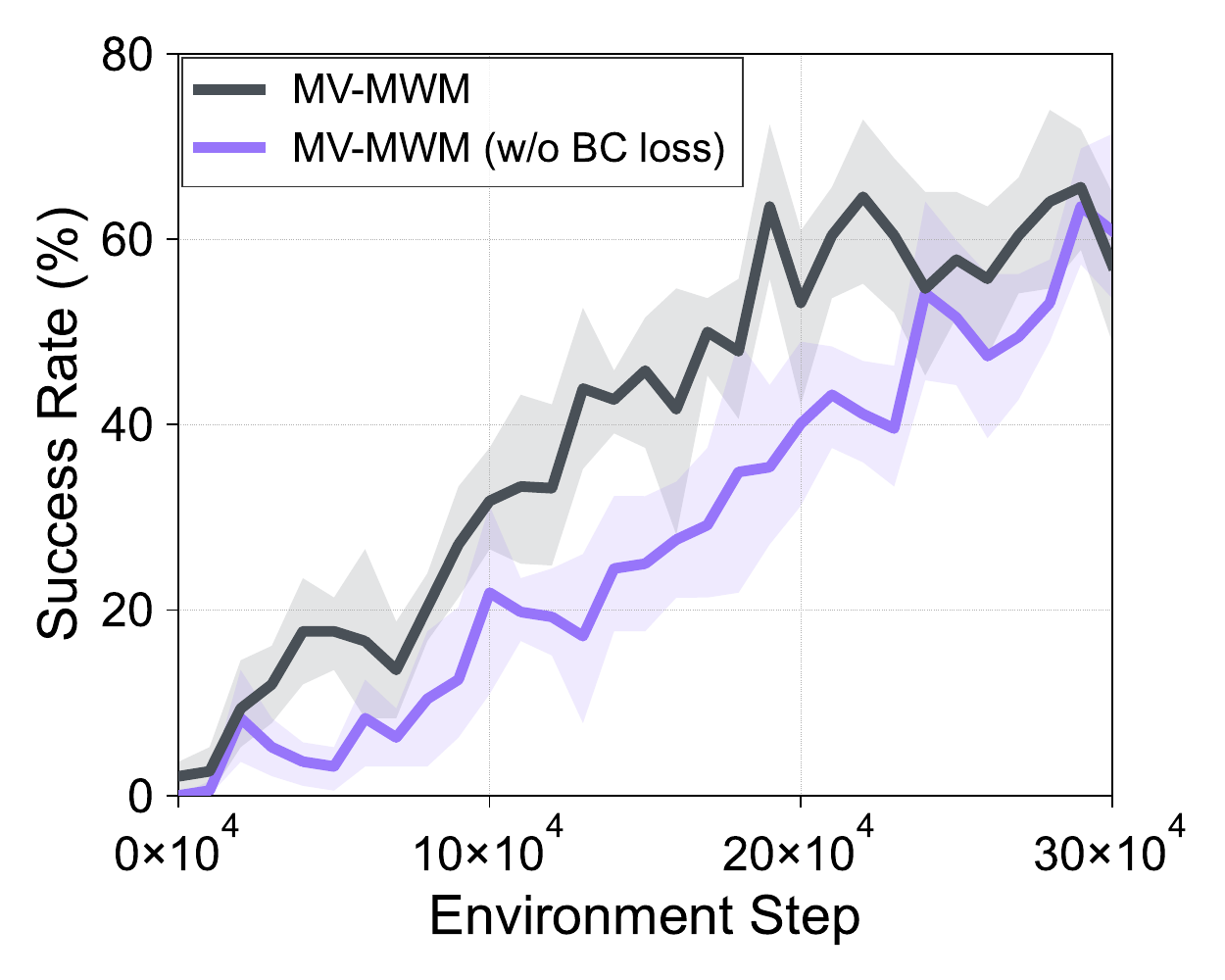}
\label{fig:analysis_bc_loss}
}
\caption{Aggregate learning curves on Phone On Base and Take Umbrella Out of Stand in single-view control to investigate the effect of (a) training ratio, (b) model size, (c) proprioceptive states, (d) fine-tuning MV-MAE encoder, and (e) auxiliary behavior cloning loss with expert demonstration. The solid line and shaded regions represent the mean and standard deviation across 4 runs.}
\label{fig:additional_experiments}
\end{figure*}

\paragraph{Training ratio}
In~\cref{fig:analysis_training_ratio}, we investigate whether MV-MWM can be scaled up for better performance by adjusting the training ratio, which is the number of gradient steps per every 16 environment steps.
We find that sample-efficiency can be further improved with training ratio higher than 1.
This insight would bring practical guidance for future researchers when applying MV-MWM to new tasks.

\paragraph{Model size}
We further investigate the scaling property of MV-MWM by scaling up the model size in~\cref{fig:analysis_model_size}.
We increase the number of patches in ViTs by reducing the patch size of MV-MAE from 16 to 8. We find that larger model achieve higher asymptotic performance as well as higher sample-efficiency.

\paragraph{Proprioceptive states}
In~\cref{fig:analysis_state}, we analyze whether feeding auxiliary proprioceptive states into the world model could improve performance. We observe that additional proprioceptive states does not make performance boost. We hypothesize that this is because our method implicitly learns 3D information by learning cross-view information from multi-view data, thus including proprioceptive state does not make large information gain.

\paragraph{Fine-tuning MV-MAE encoder}
We investigate whether fine-tuning MV-MAE encoder when training world model could improve performance; we fine-tune the last one or two layers of the MV-MAE encoder and freeze all the other layers of the MV-MAE encoder. We observe that fine-tuning does not make gains in~\cref{fig:analysis_finetune}, which shows that our visual representation learning scheme enables the autoencoder to effectively capture information required for solving the task.

\paragraph{Behavior cloning loss with expert demonstration}
\cref{fig:analysis_bc_loss} shows that how behavior cloning loss using expert demonstrations affect performance. We find that using behavior loss improves sample-efficiency, yet the asymptotic performance converges to that without the behavior cloning loss. The behavior cloning loss could implicitly reduce an exploration space into area nearby expert demonstrations, which accelerates training especially in the early phase. 

\clearpage

\begin{figure*}[hb]
\centering
\subfigure[Data augmentation]{
\includegraphics[width=0.31\textwidth]{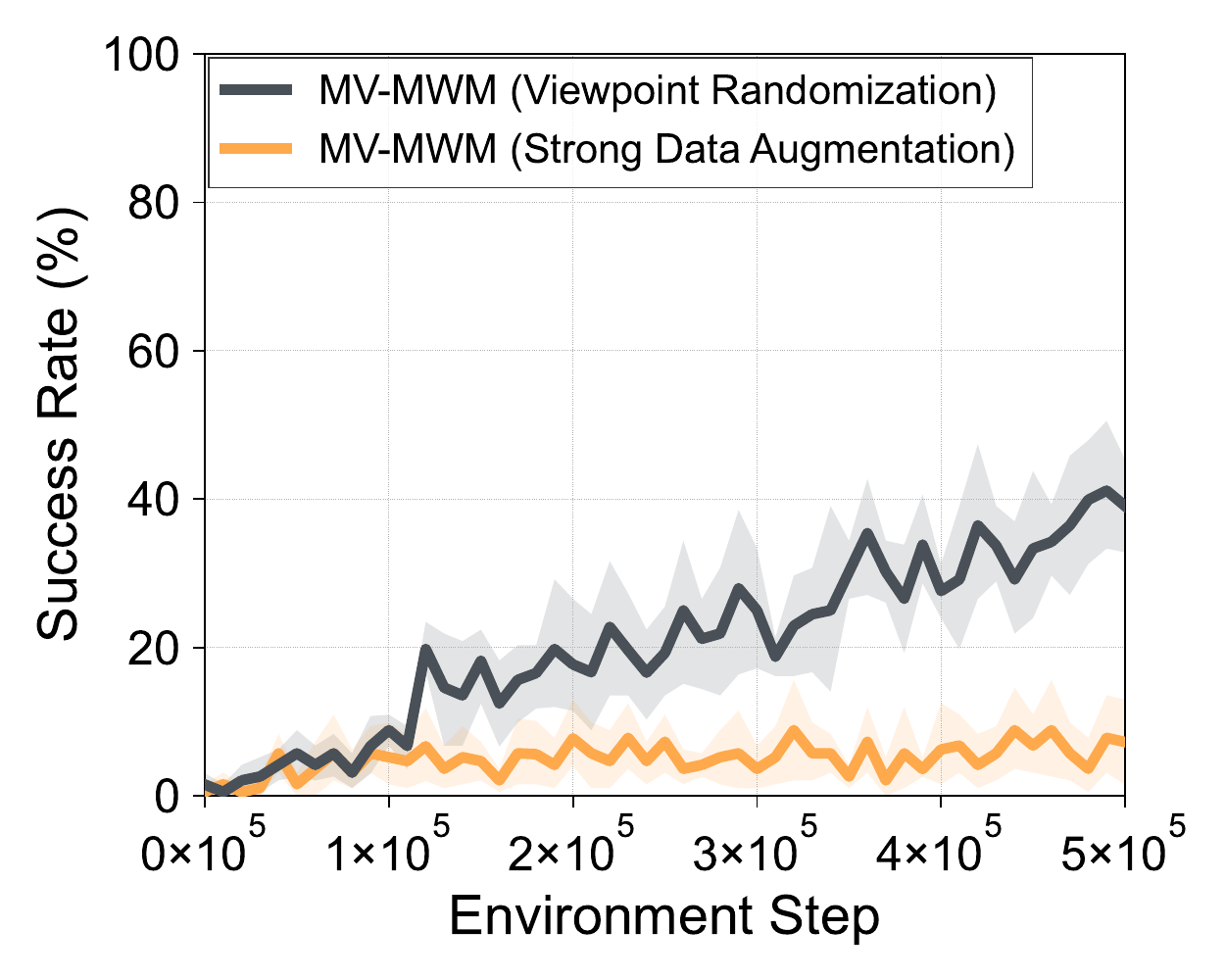}
\label{fig:data_aug}
}
\subfigure[Three cameras]{
\includegraphics[width=0.31\textwidth]{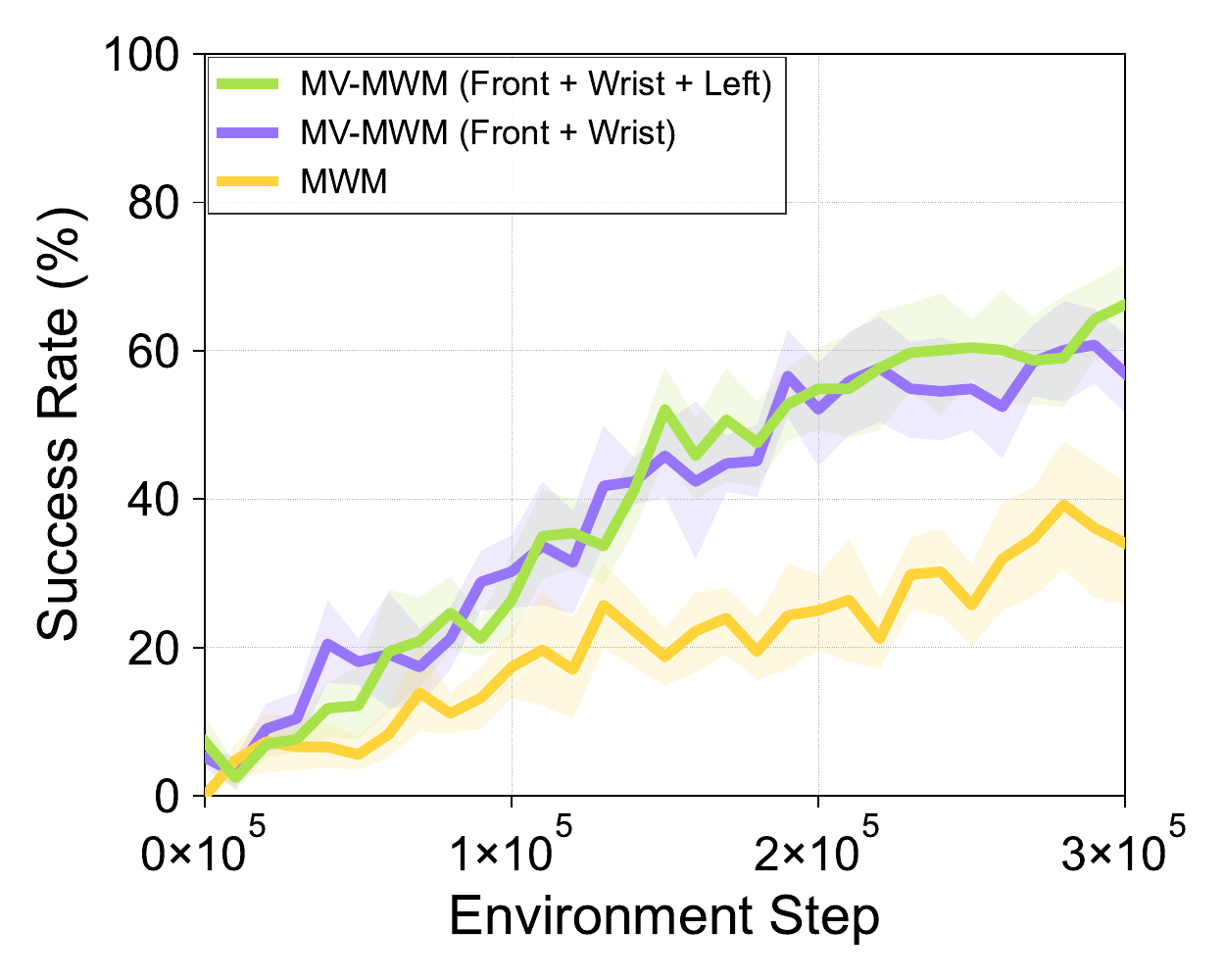}
\label{fig:three_cameras}
}
\vspace{-0.15in}
\caption{(a) Aggregate learning curves on Phone On Base* and Pick Up Cup* in viewpoint-robust control to investigate the effect of using randomized cameras. (b) Aggregate learning curves of multi-view visual control agents for solving three manipulation tasks.}
\vspace{-0.2in}
\end{figure*}

\paragraph{Data augmentation for viewpoint-robust control}
To investigate the importance of using visual observations from randomized cameras, we consider a baseline that uses images perturbed with data augmentation (\textit{i.e.,} rotation, translation, brightness, and contrast) for multi-view representation learning.
As shown in \cref{fig:data_aug}, we observe that using the images from physically perturbed images largely outperforms the baseline based on data augmentation.
This is because such a randomized camera can provide images from different perspectives that contain additional information which is not available from the single, fixed camera viewpoint.
On the other hand, data augmentation fails to give such information useful for implicitly capturing 3D information of a robot workspace.
Given this result, investigation into how to set up a real-world robot learning setup with a randomized camera would be an interesting and important future research direction.

\paragraph{Three cameras experiments}
To assess how the number of viewpoints used for representation learning affects the performance, we train MV-MWM with more than two cameras in multi-view representation learning: three cameras of \{Front, Wrist, and Left Shoulder\}.
As shown in~\cref{fig:three_cameras}, we find that including additional Left Shoulder camera does not largely affect the performance compared to using only Front and Wrist cameras.
We hypothesize this is because a widely-used camera configuration with Front and Wrist camera is already sufficient for capturing the information required for completing the considered tasks.
It would be an interesting future direction to investigate the importance of specific camera viewpoints for solving a variety of tasks.

\begin{figure*}[hb]
\centering
\subfigure{
\includegraphics[width=0.31\textwidth]{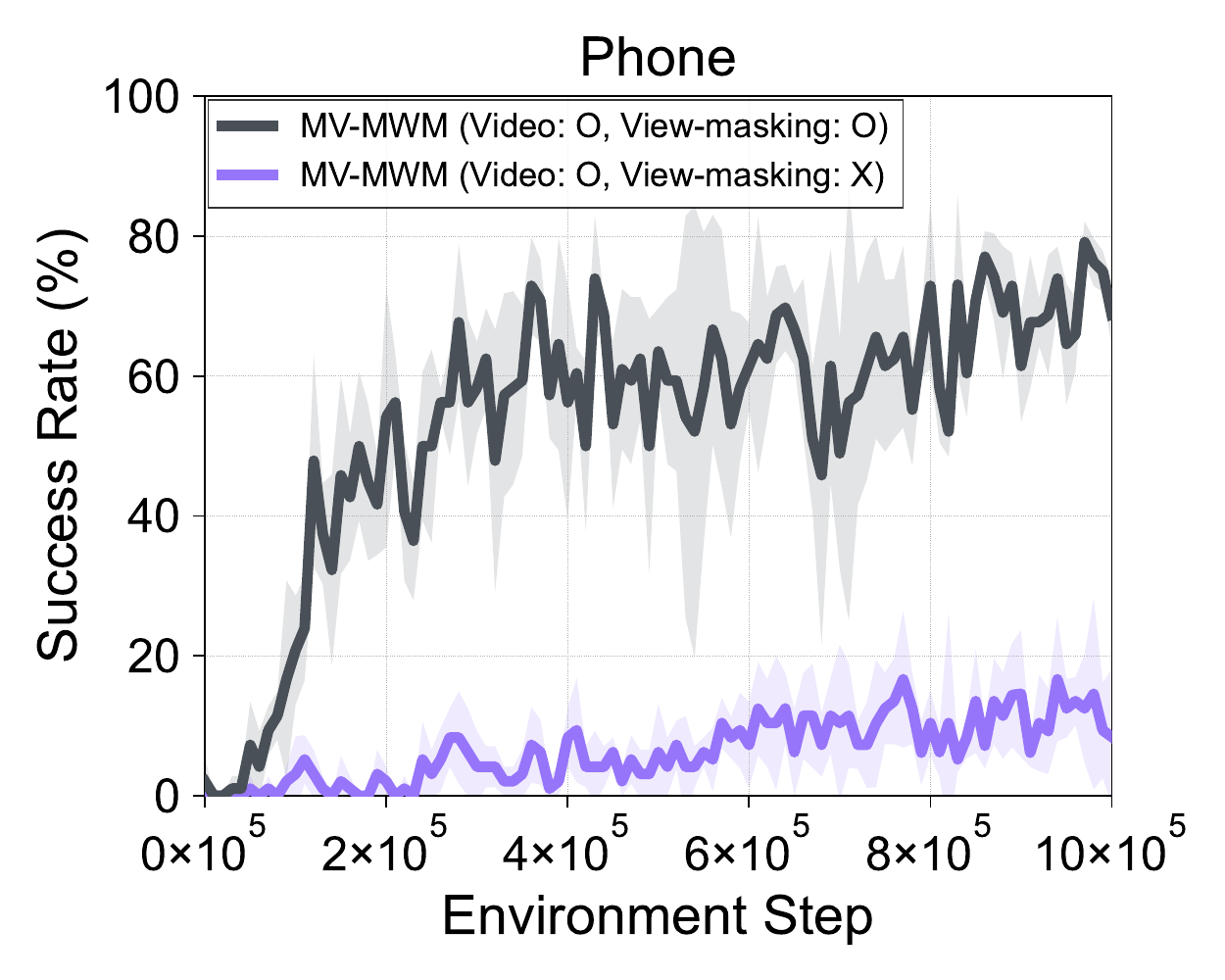}
\label{fig:longer_training_view}
}
\subfigure{
\includegraphics[width=0.31\textwidth]{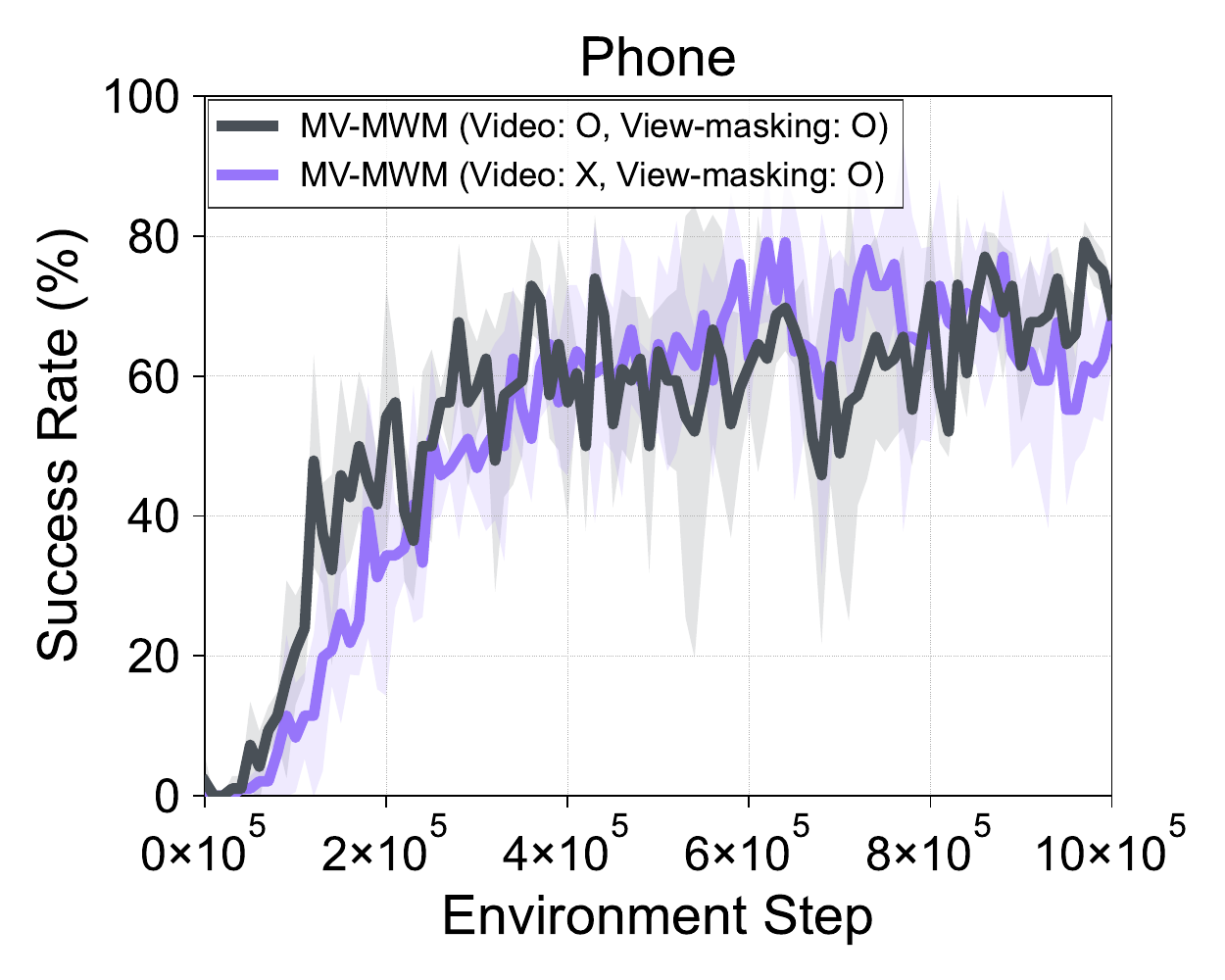}
\label{fig:longer_training_video}
}
\subfigure{
\includegraphics[width=0.31\textwidth]{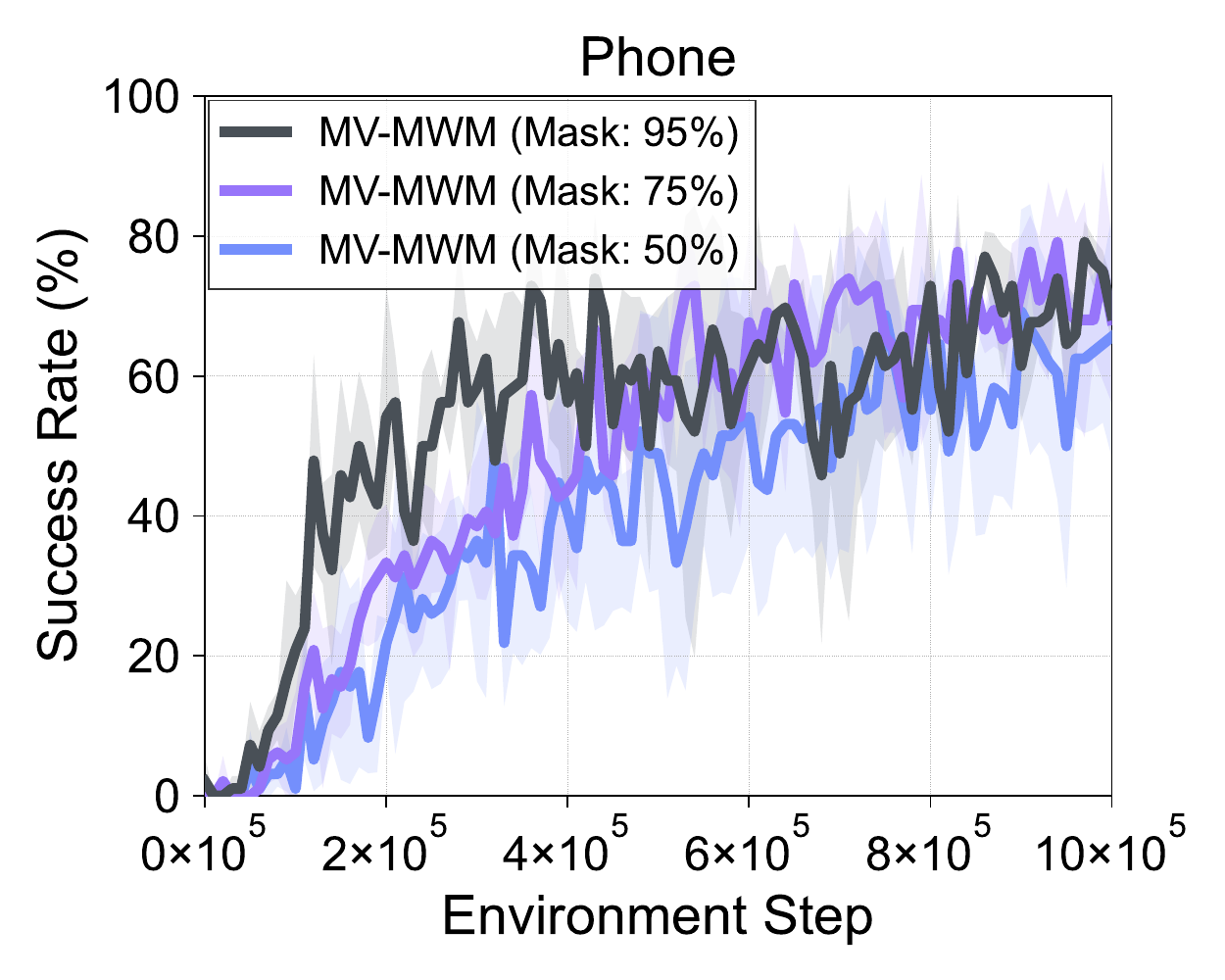}
\label{fig:longer_training_masking}
}
\vspace{-0.15in}
\caption{Learning curves of single-view visual control agents operating on the front camera for solving Phone on Base task from RLBench~\citep{james2020rlbench}, investigating the effect of (a) view masking, (b) video autoencoding and (c) masking ratio. The solid line and shaded regions represent the mean and stratified bootstrap confidence interval across 12 runs.}
\label{fig:longer_training}
\end{figure*}

\paragraph{Longer training step} 
In~\cref{fig:longer_training}, we report the experimental results with larger training steps to provide asymptotic results of our analysis experiments.
We observe that the performance gain from view-masking is significant over longer training horizon, which shows the effectiveness of the proposed masking scheme.
On the other hand, because the video autoencoding is introduced to improve the sample-efficiency at the initial phase of training by addressing the difficulty of training with the view-masking, we find that the benefit of employing the video autoencoding becomes less significant in a more long-term manner.
Moreover, we observe that high masking ratio is crucial for sample-efficiency but asymptotic performance with different masking ratios is similar.
We hypothesize this is because our proposed view-masking scheme can asymptotically encourage the model to learn useful representations even with the low masking ratio as 50\%.

\end{document}